# Artificial Intelligence and Robotics

Javier Andreu Perez, Fani Deligianni, Daniele Ravi and Guang-Zhong Yang

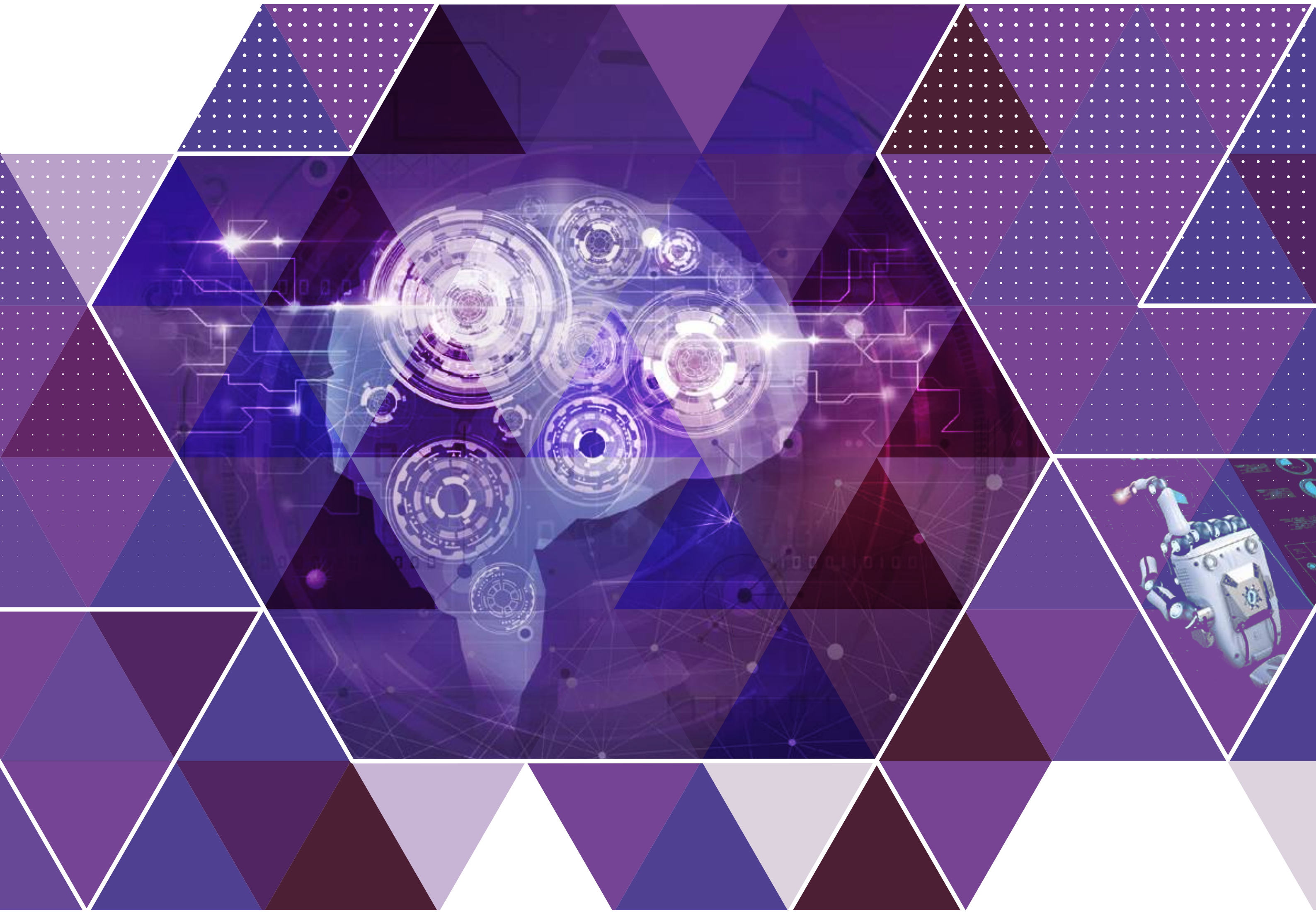

UK-RAS NETWORK
ROBOTICS & AUTONOMOUS SYSTEMS

UKRAS.ORG

# FOREWORD

Welcome to the UK-RAS White Paper Series on Robotics and Autonomous Systems (RAS). This is one of the core activities of UK-RAS Network, funded by the Engineering and Physical Sciences Research Council (EPSRC). By bringing together academic centres of excellence, industry, government, funding bodies and charities, the Network provides academic leadership, expands collaboration with industry while integrating and coordinating activities at EPSRC funded RAS capital facilities, Centres for Doctoral Training and partner universities.

The recent successes of AI have captured the wildest imagination of both the scientific communities and the general public. Robotics and AI amplify human potentials, increase productivity and are moving from simple reasoning towards human-like cognitive abilities. Current AI technologies are used in a set area of applications, ranging from healthcare, manufacturing, transport, energy, to financial services, banking, advertising, management consulting and government agencies. The global AI market is around 260 billion USD in 2016 and it is estimated to exceed 3 trillion by 2024. To understand the impact of AI, it is important to draw lessons from it's past successes and failures and this white paper provides a comprehensive explanation of the evolution of AI, its current status and future directions.

The UK-RAS white papers are intended to serve as a basis for discussing the future technological roadmaps, engaging the wider community and stakeholders, as well as policy makers, in assessing the potential social, economic and ethical/legal impact of RAS. It is our plan to provide annual updates for these white papers so your feedback is essential - whether it is to point out inadvertent omissions of specific areas of development that need to covered, or to suggest major future trends that deserve further debate and in-depth analysis.

Please direct all your feedback to white-paper@ukras.org. We look forward to hearing from you!

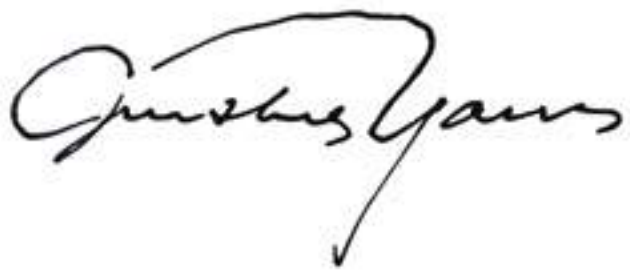

Prof Guang-Zhong Yang, CBE, FREng
Chair, UK-RAS Network

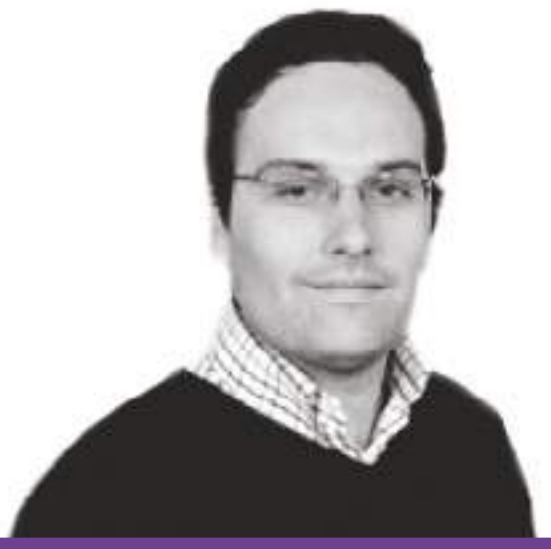

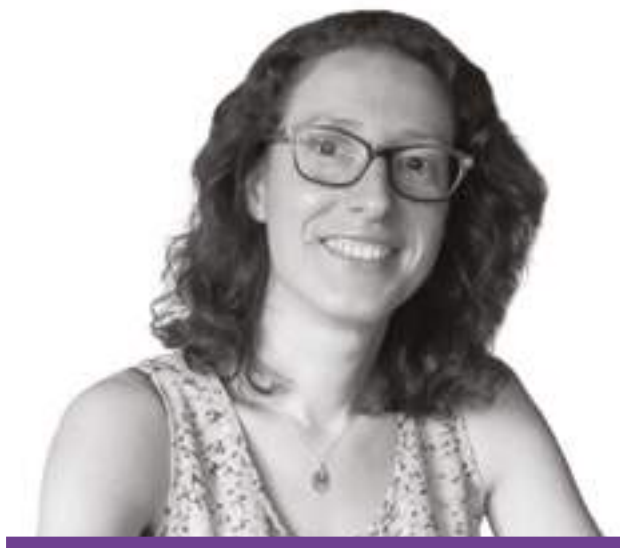

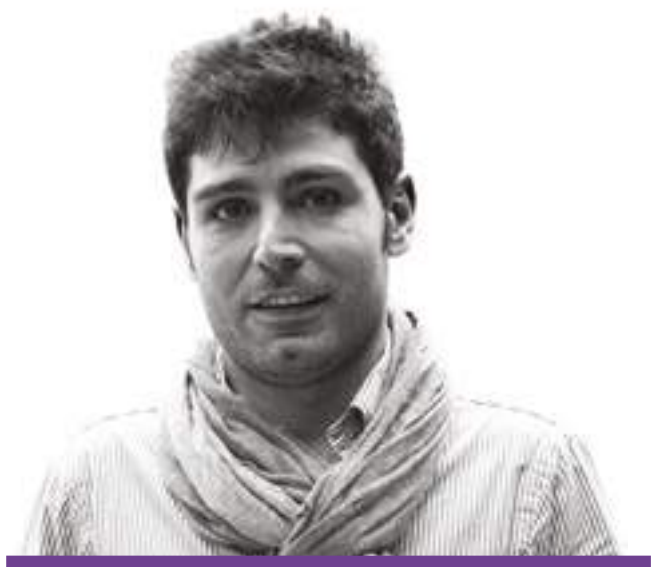

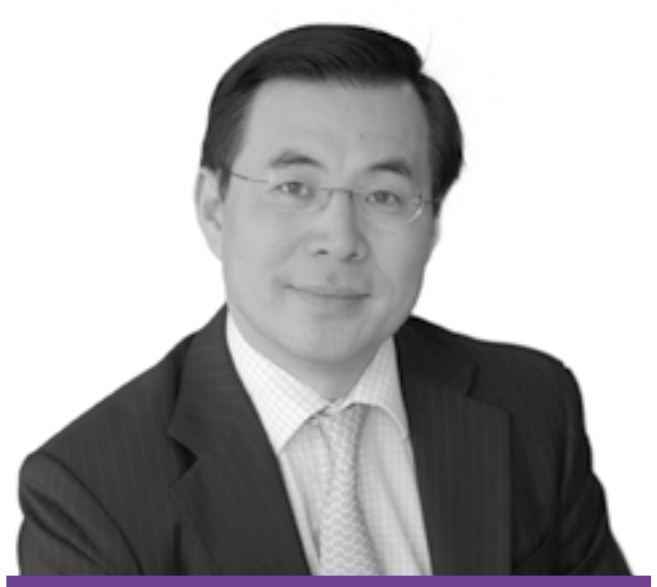

Dr. Javier Andreu Perez,
The Hamlyn Centre,
Imperial College London

Dr. Fani Deligianni,
The Hamlyn Centre,
Imperial College London

Dr. Daniele Ravi,
The Hamlyn Centre,
Imperial College London

Prof. Dr. Guang-Zhong Yang,
The Hamlyn Centre,
Imperial College London

On behalf of the UK-RAS Network, established to provide academic leadership, expand collaboration with industry while integrating and coordinating activities at EPSRC funded RAS capital facilities, Centres for Doctoral Training and partner universities.

# EXECUTIVE SUMMARY

In 2016, Artificial Intelligence (AI) celebrated its 60th anniversary of the Dartmouth Workshop, which marked the beginning of AI being recognised as an academic discipline. One year on, the pace of AI has captured the wildest imagination of both the scientific community and the general public.

The term AI now encompasses the whole conceptualisation of a machine that is intelligent in terms of both operational and social consequences. With the prediction of the AI market to reach 3 trillion by 2024, both industry and government funding bodies are investing heavily in AI and robotics. As the availability of information around us grows, humans will rely more and more on AI systems to live, to work, and to entertain. Given increased accuracy and sophistication of AI systems, they will be used in an increasingly diverse range of sectors including finance, pharmaceuticals, energy, manufacturing, education, transport and public services. It has been predicted that the next stage of AI is the era of augmented intelligence. Ubiquitous sensing systems and wearable technologies are driving towards intelligent embedded systems that will form a natural extension of human beings and our physical abilities. Will AI trigger a transformation leading to super-intelligence that would surpass all human intelligence?

This white paper explains the origin of AI, its evolution in the last 60 years, as well as related subfields including machine learning, computer vision and the rise of deep learning. It provides a rational view of the different seasons of AI and how to learn from these 'boom-and-bust' cycles to ensure the current progresses are sustainable and here to stay. Along with the unprecedented enthusiasm of AI, there are also fears about the impact of the technology on our society. A clear strategy is required to consider the associated ethical and legal challenges to ensure that society as a whole will benefit from the evolution of AI and its potential negative impact is mitigated from early on. To this end, the paper outlines the ethical and legal issues of AI, which encompass privacy, jobs, legal responsibility, civil rights, and wrongful use of AI for military purposes. The paper concludes by providing a set of recommendations to the research community, industry, government agencies and policy makers.

To sustain the current progress of AI, it is important to understand what is science fiction and what is practical reality. A rational and harmonic interaction is required between application specific projects and visionary research ideas. Neither the unrealistic enthusiasm nor the unjustified fears of AI should hinder its progress. They should be used to motivate the development of a systematic framework on which the future of AI will flourish. With sustained funding and responsible investment, AI is set to transform the future of our society - our life, our living environment and our economy.

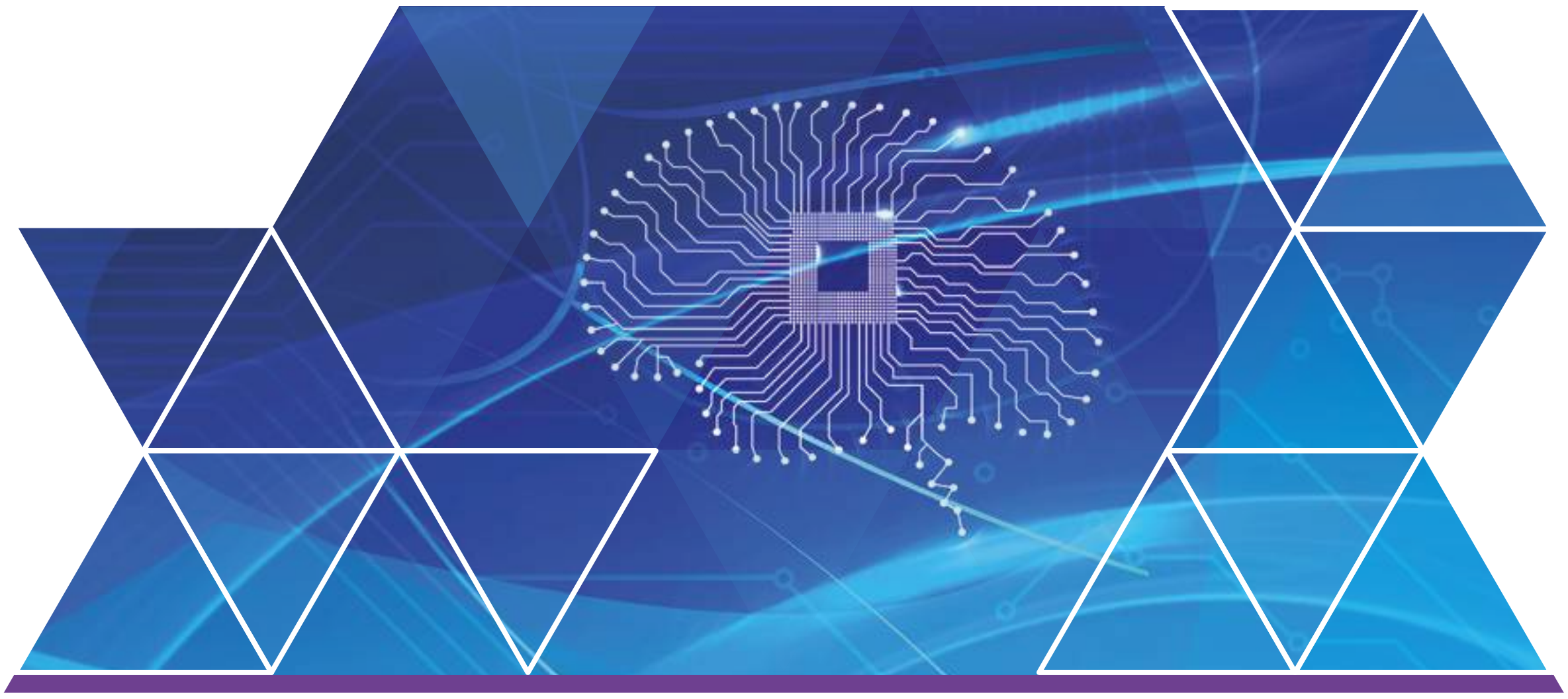

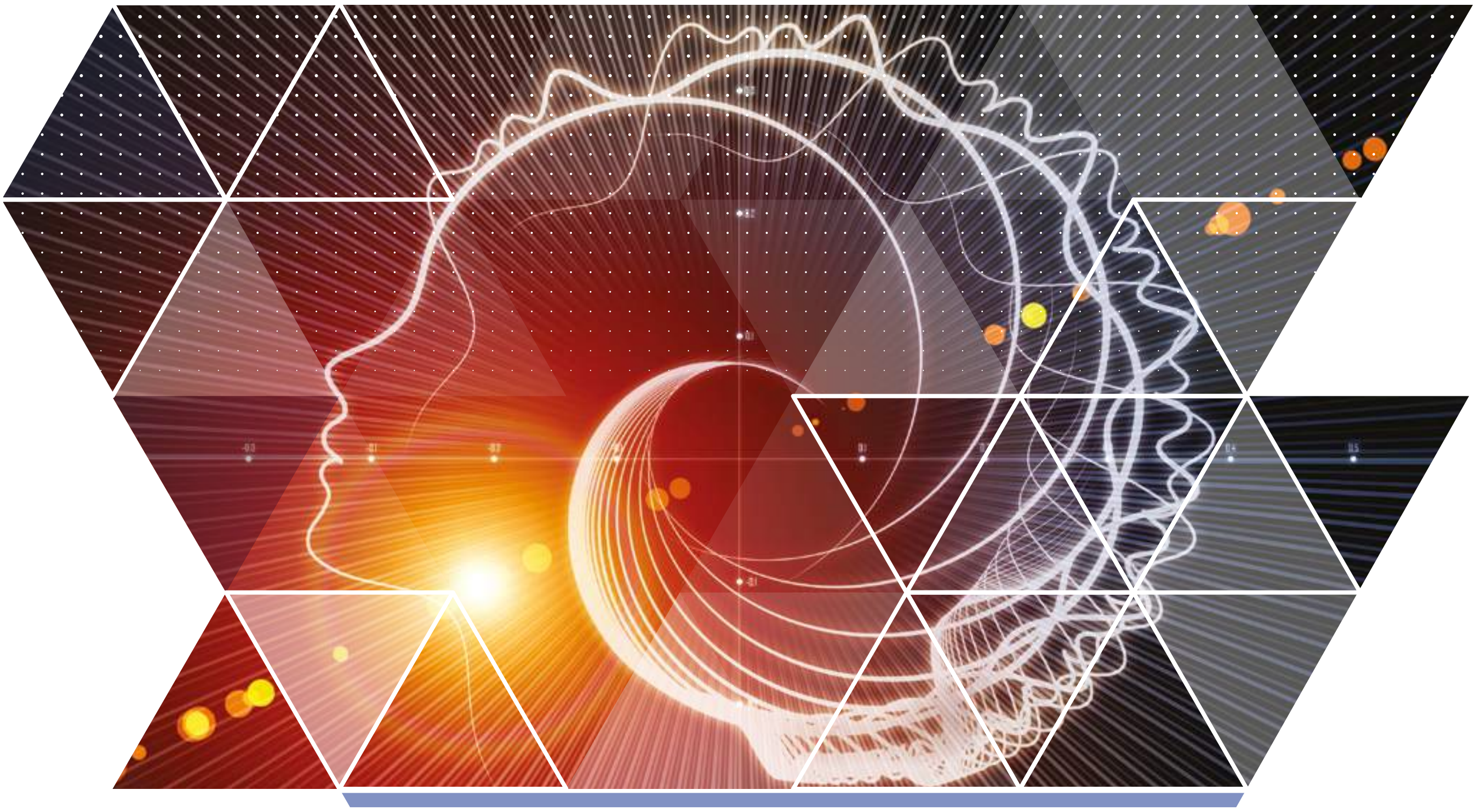

"Robotics and AI augment and amplify human potentials, increase productivity and are moving from simple reasoning towards human-like cognitive abilities. To understand the impact of AI, it is important to draw lessons from the past successes and failures, as well as to anticipate its future directions and potential legal, ethical and socio-economic implications."

# CONTENTS

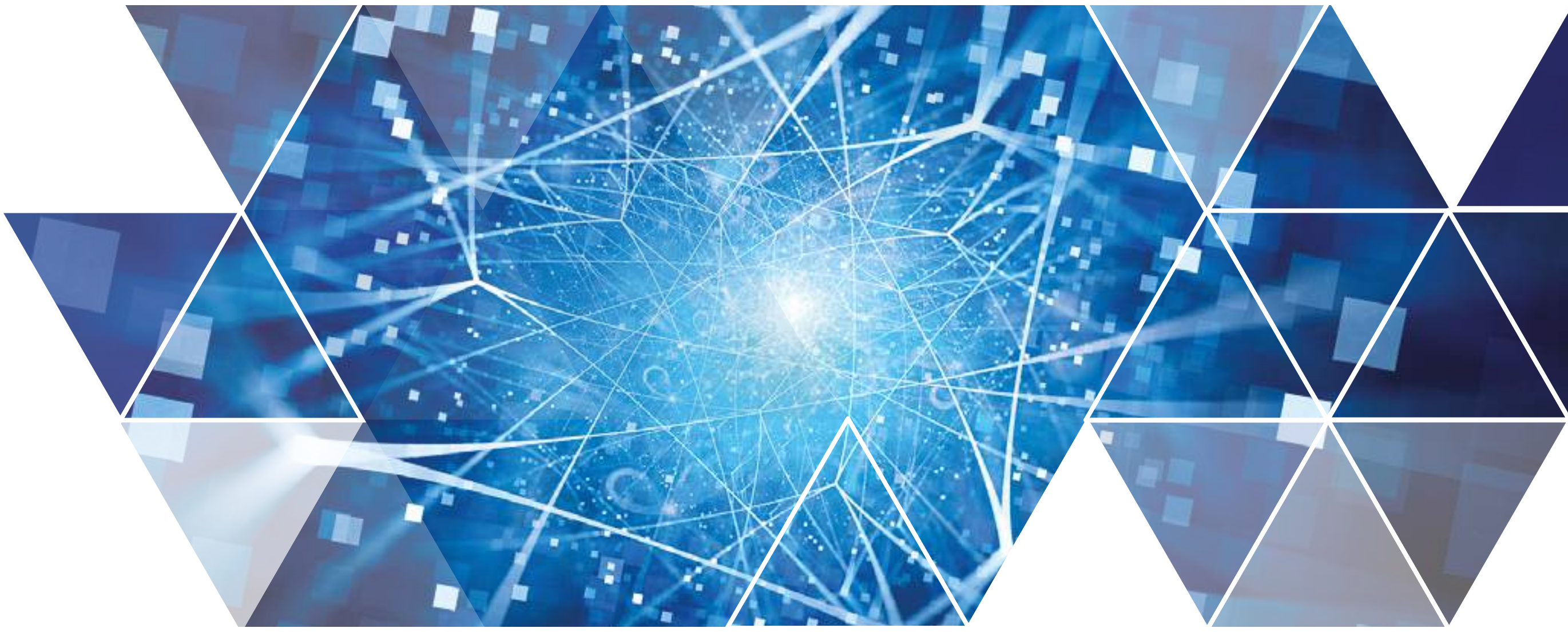

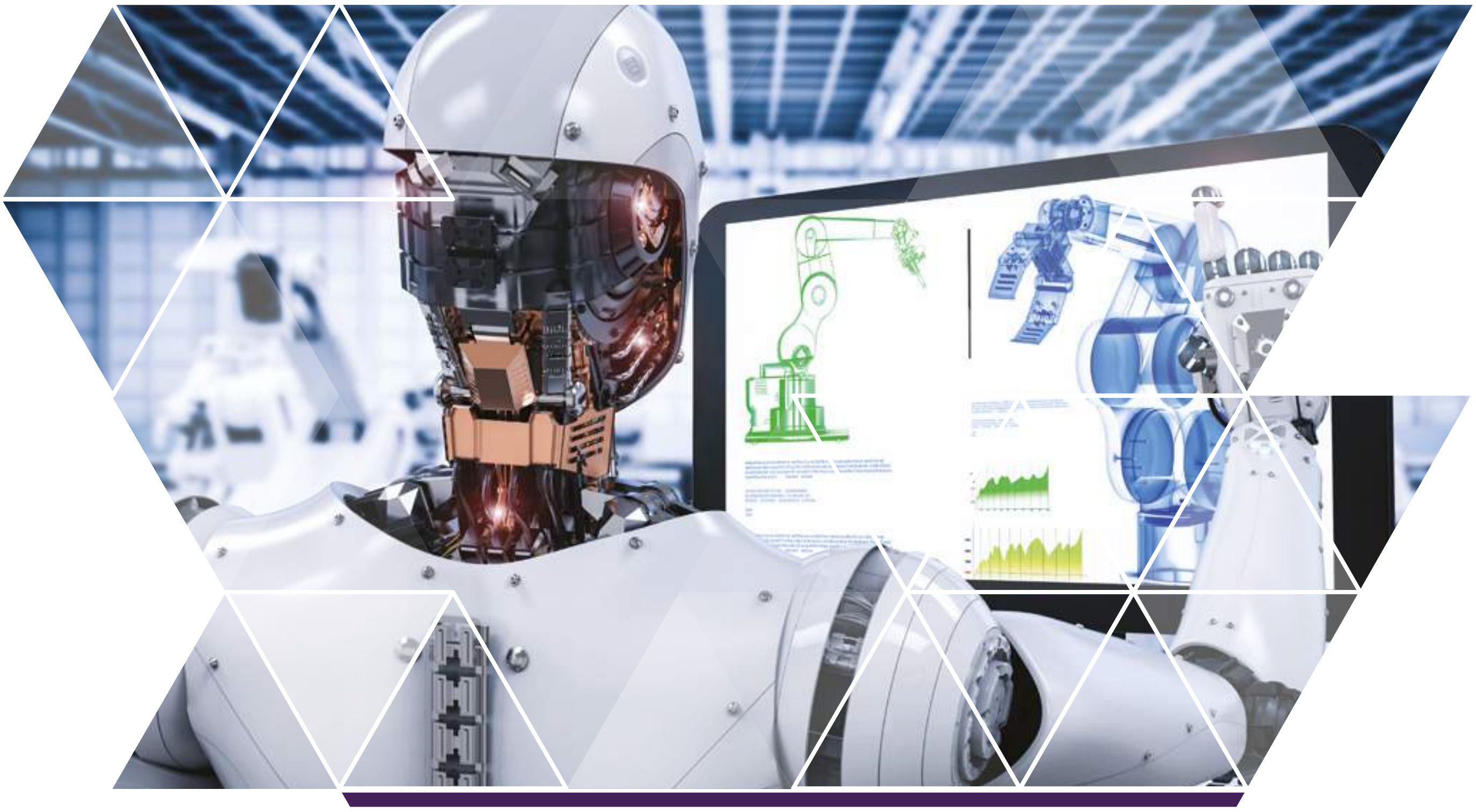

> "With increased capabilities and sophistication of AI systems, they will be used in more diverse ranges of sectors including finance, pharmaceuticals, energy, manufacturing, education, transport and public services. The next stage of AI is the era of augmented intelligence, seamlessly linking human and machine together."

# 1. INTRODUCTION

Artificial Intelligence (AI) is a commonly employed appellation to refer to the field of science aimed at providing machines with the capacity of performing functions such as logic, reasoning, planning, learning, and perception. Despite the reference to "machines" in this definition, the latter could be applied to "any type of living intelligence". Likewise, the meaning of intelligence, as it is found in primates and other exceptional animals for example, it can be extended to include an interleaved set of capacities, including creativity, emotional knowledge, and self-awareness.

The term AI was closely associated with the field of "symbolic AI", which was popular until the end of the 1980s. In order to overcome some of the limitations of symbolic AI, subsymbolic methodologies such as neural networks, fuzzy systems, evolutionary computation and other computational models started gaining popularity, leading to the term "computational intelligence" emerging as a subfield of AI.

Nowadays, the term AI encompasses the whole conceptualisation of a machine that is intelligent in terms of both operational and social consequences. A practical definition used is one proposed by Russell and Norvig: "Artificial Intelligence is the study of human intelligence and actions replicated artificially, such that the resultant bears to its design a reasonable level of rationality" [1]. This definition can be further refined by stipulating that the level of rationality may even supersede humans, for specific and well-defined tasks.

Current AI technologies are used in online advertising, driving, aviation, medicine and personal assistance image recognition. The recent success of AI has captured the imagination of both the scientific community and the public. An example of this is vehicles equipped with an automatic steering system, also known as autonomous cars. Each vehicle is equipped with a series of lidar sensors and cameras which enable recognition of its three-dimensional environment and provides the ability to make intelligent decisions on maneuvers in variable, real-traffic road conditions. Another example is the Alpha-Go, developed by Google Deepmind, to play the board game Go. Last year, Alpha-Go defeated the Korean grandmaster Lee Sedol, becoming the first machine to beat a professional player and recently it went on to win against the current world number one, Ke Jie, in China. The number of possible games in Go is estimated to be $10^{761}$ and given the extreme complexity of the game, most AI researchers believed it would be years before this could happen. This has led to both the excitement and fear in many that AI will surpass humans in all the fields it marches into.

However, current AI technologies are limited to very specific applications. One limitation of AI, for example, is the lack of "common sense"; the ability to judge information beyond its acquired knowledge. A recent example is that of the AI robot Tay developed by Microsoft and designed for making conversations on social networks. It had to be disconnected shortly after its launch because it was not able to distinguish between positive and negative human interaction. AI is also limited in terms of emotional intelligence. AI can only detect basic human emotional states such as anger, joy, sadness, fear, pain, stress and neutrality. Emotional intelligence is one of the next frontiers of higher levels of personalisation.

True and complete AI does not yet exist. At this level, AI will mimic human cognition to a point that it will enable the ability to dream, think, feel emotions and have own goals. Although there is no evidence yet this kind of true AI could exist before 2050, nevertheless the computer science principles driving AI forward, are rapidly advancing and it is important to assess its impact, not only from a technological standpoint, but also from a social, ethical and legal perspective.

# 2. THE BIRTH AND BOOM OF AI

The birth of the computer took place when the first calculator machines were developed, from the mechanical calculator of Babbage, to the electromechanical calculator of Torres-Quevedo. The dawn of automata theory can be traced back to World War II with what was known as the “codebreakers”. The amount of operations required to decode the German trigrams of the Enigma machine, without knowing the rotor’s position, proved to be too challenging to be solved manually. The inclusion of automata theory in computing conceived the first logical machines to account for operations such as generating, codifying, storing and using information. Indeed, these four tasks are the basic operations of information processing performed by humans. The pioneering work by Ramón y Cajal marked the birth of neuroscience, although many neurological structures and stimulus responses were already known and studied before him. For the first time in history the concept of “neuron” was proposed. McClulloch and Pitts further developed a connection between automata theory and neuroscience, proposing the first artificial neuron which, years later, gave rise to the first computational intelligence algorithm, namely “the perceptron”. This idea generated great interest among prominent scientists of the time, such as Von Neumann, who was the pioneer of modern computers and set the foundation for the connectionism movement.

**1956 –**
**when the term AI was first coined.**

The Dartmouth Conference of 1956 was organized by Marvin Minsky, John McCarthy and two senior scientists, Claude Shannon and Nathan Rochester of IBM. At this conference, the expression “Artificial Intelligence” was first coined as the title of the field. The Dartmouth conference triggered a new era of discovery and unrestrained conquests of new knowledge. The computer programmes developed at the time are considered by most as simply "extraordinary"; computers solve algebraic problems, demonstrate theorems in geometry and learnt to speak English. At that time, many didn’t believe that such "intelligent" behavior was possible in machines. Researchers displayed a great deal of optimism both in private and in scientific publications. They predicted that a completely intelligent machine would be built in the next 20 years. Government agencies, such as the US Defence and Research Project Agency (DARPA), were investing heavily in this new area. It is worth mentioning, that some of the aforementioned scientists, as well as major laboratories of the time, such as Los Alamos (Nuevo Mexico, USA), had strong connections with the army and this link had a prominent influence, as the work at Bletchley Park (Milton Keynes, UK) had, over the course of WWII, as did political conflicts like the Cold War in AI innovation.

> “Since 2000, a third renaissance of the connectionism paradigm arrived with the dawn of Big Data, propelled by the rapid adoption of the internet and mobile communication.”

In 1971, DARPA funded a consortium of leading laboratories in the field of speech recognition. The project had the ambitious goal of creating a fully functional speech recognition system with a large vocabulary. In the middle of the 1970s, the field of AI endured fierce criticism and budgetary restrictions, as AI research development did not match the overwhelming expectations of researchers.. When promised results did not materialize, investment in AI eroded. Following disappointing results, DARPA withdrew funding in speech recognition and this, coupled with other events such as the failure of machine translation, the abandonment of connectionism and the Lighthill report, marked the first winter of AI [2]. During this period, connectionism stagnated for the next 10 years following a devastating critique by Marvin Minksy on perceptrons [3].

From 1980 until 1987, AI programmes, called "expert systems", were adopted by companies and knowledge acquisition become the central focus of AI research. At the same time, the Japanese government launched a massive funding program on AI, with its fifth-generation computers initiative. Connectionism was also revived by the work of John Hopfield [4] and David Rumelhart [5].

AI researchers who had experienced the first backlash in 1974, were sceptical about the reignited enthusiasms of expert systems and sadly their fears were well founded. The first sign of a changing tide was with the collapse of the AI computer hardware market in 1987. Apple and IBM desktops had gradually improved their speed and power and in 1987 they were more powerful than the best LISP machines on the market. Overnight however, the industry collapsed and billions of dollars were lost. The difficulty of updating and reprograming the expert systems, in addition to the high maintenance costs, led to the second AI winter. Investment in AI dropped and DARPA stopped its strategic computing initiative, claiming AI was no longer the "latest mode". Japan also stopped funding its fifth-generation computer program as the proposed goals were not achieved.

In the 1990s, the new concept of "intelligent agent" emerged [6]. An agent is a system that perceives its environment and undertakes actions that maximize its chances of being successful. The concept of agents conveys, for the first time, the idea of intelligent units working collaboratively with a common objective. This new paradigm was intended to mimic how humans work collectively in groups, organizations and/or societies. Intelligent agents proved to be a more polyvalent concept of intelligence. In the late 1990s, fields such as statistical learning from several perspectives including probabilistic, frequentist and possibilistic (fuzzy logic) approaches, were linked to AI to deal with the uncertainty of decisions. This brought a new wave of successful applications for AI, beyond what expert systems had achieved during the 1980s. These new ways of reasoning were more suited to cope with the uncertainty of intelligent agent states and perceptions and had its major impact in the field of control. During this time, high-speed trains controlled by fuzzy logic, were developed [7] as were many other industrial applications (e.g. factory valves, gas and petrol tanks surveillance, automatic gear transmission systems and reactor control in power plants) as well as household appliances with advanced levels of intelligence (e.g. air-conditioners, heating systems, cookers and vacuum-cleaners). These were different to the expert systems in 1980s; the modelling of the inference system for the task, achieved through learning, gave rise to the field of Machine Learning. Nevertheless, although machine reasoning exhibited good performance, there was still an engineering requirement to digest the input space into a new source, so that intelligence could reason more effectively. Since 2000, a third renaissance of the connectionism paradigm arrived with the dawn of Big Data, propelled by the rapid adoption of the Internet and mobile communication. Neural networks were once more considered, particularly in the role they played in enhancing perceptual intelligence and eliminating the necessity of feature engineering. Great advances were also made in computer vision, improving visual perception, increasing the capabilities of intelligent agents and robots in performing more complex tasks, combined with visual pattern recognition. All these paved the way to new AI challenges such as, speech recognition, natural language processing, and self-driving cars. A timeline of key highlights in the history of AI is shown in Figure 1.

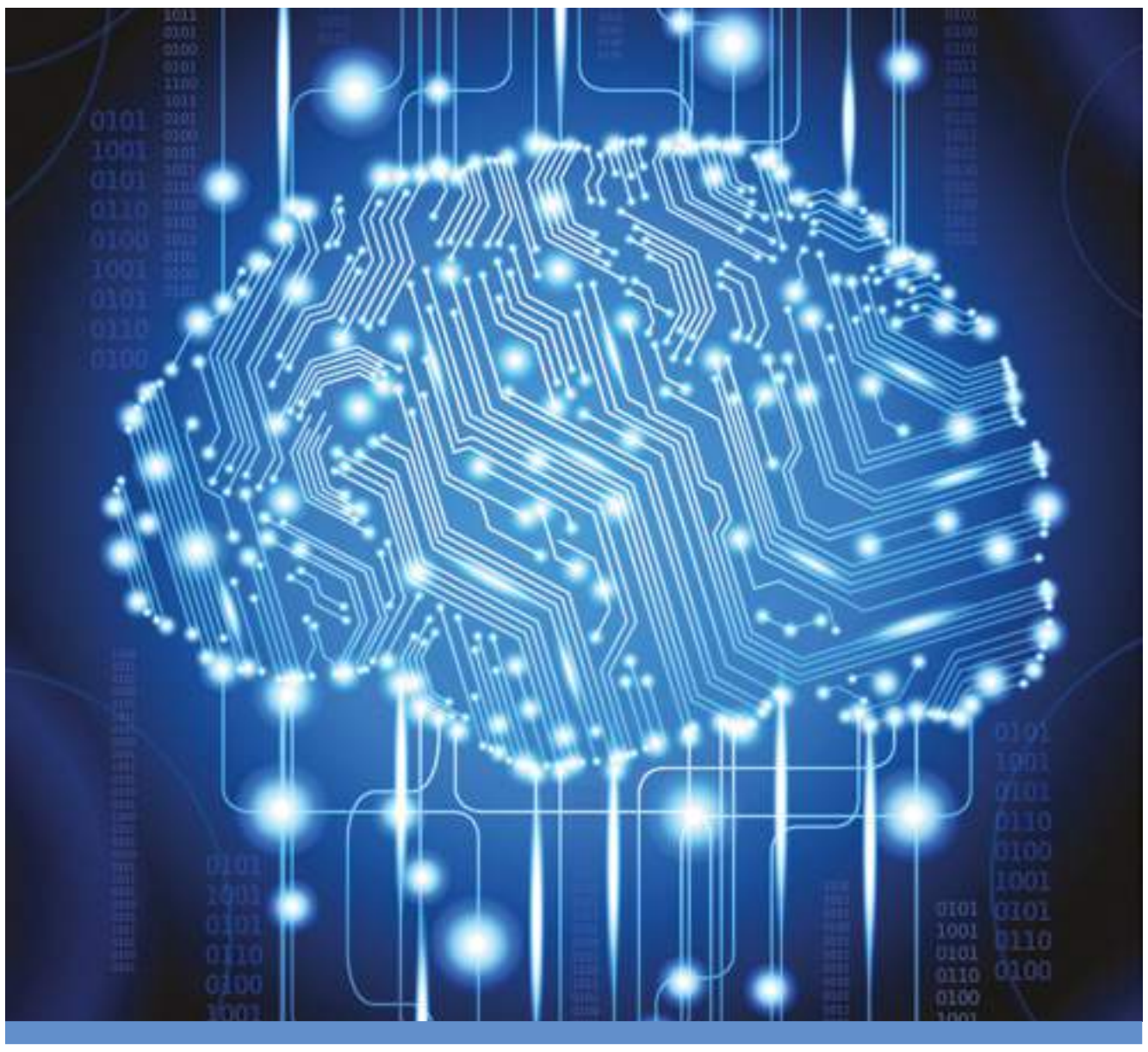

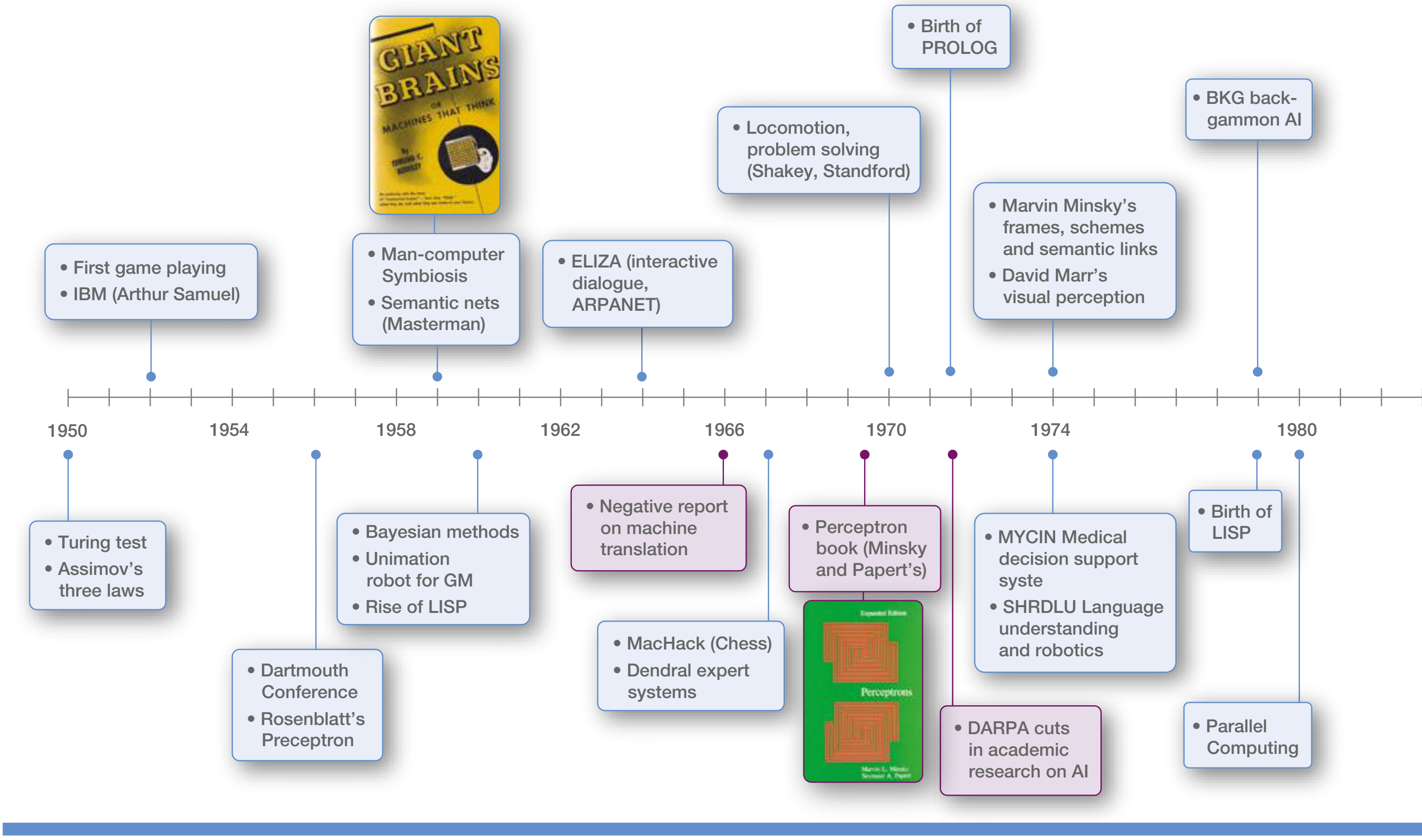

**Figure 1.**
A timeline highlighting some of the most relevant events of AI since 1950. The blue boxes represent events that have had a positive impact on the development of AI. In contrast, those with a negative impact are shown in red and reflect the low points in the evolution of the field, i.e. the so-called "winters" of AI.

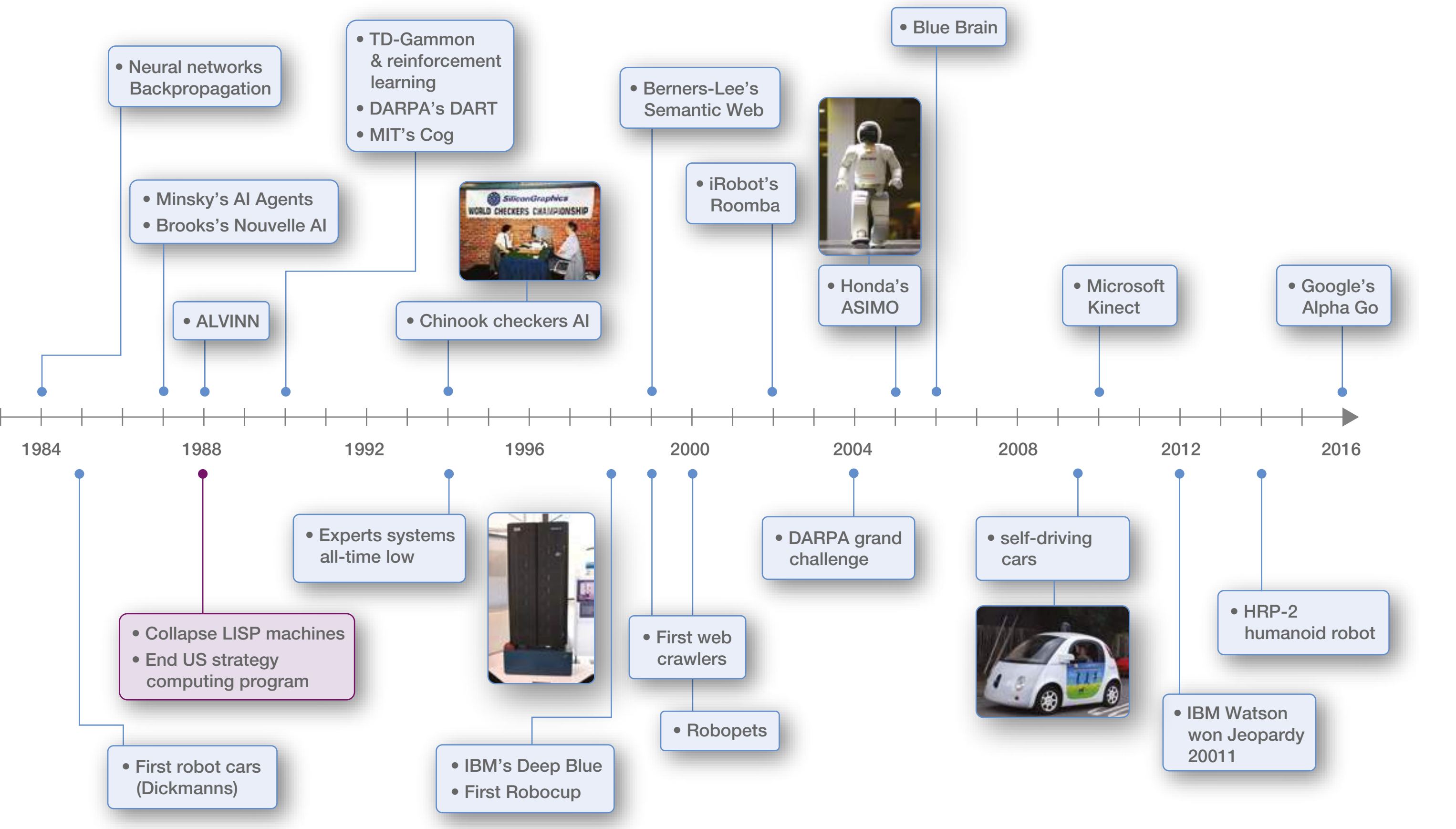

## Weak and Strong AI

When defining the capacity of AI, this is frequently categorised in terms of weak or strong AI.
Weak AI (narrow AI) is one intended to reproduce an observed behaviour as accurately as possible. It can carry out a task for which they have been precision-trained. Such AI systems can become extremely efficient in their own field but lack generalisational ability. Most existing intelligent systems that use machine learning, pattern recognition, data mining or natural language processing are examples of weak AI. Intelligent systems, powered with weak AI include recommender systems, spam filters, self-driving cars, and industrial robots.
Strong AI is usually described as an intelligent system endowed with real consciousness and is able to think and reason in the same way as a human being. A strong AI can, not only assimilate information like a weak AI, but also modify its own functioning, i.e. is able to autonomously reprogram the AI to perform general intelligent tasks. These processes are regulated by human-like cognitive abilities including consciousness, sentience, sapience and self-awareness. Efforts intending to generate a strong AI have focused on whole brain simulations, however this approach has received criticism, as intelligence cannot be simply explained as a biological process emanating from a single organ but is a complex coalescence of effects and interactions between the intelligent being and its environment, encompassing a series of diverse ways via interlinked biological process.

# 3. QUESTIONING THE IMPACT OF AI

Given the exponential rise of interest in AI, experts have called for major studies on the impact of AI on our society, not only in technological but also in legal, ethical and socio-economic areas. This response also includes the speculation that autonomous super artificial intelligence may one day supersede the cognitive capabilities of humans. This future scenario is usually known in AI forums as the “AI singularity” [8]. This is commonly defined as the ability of machines to build better machines by themselves. This futuristic scenario has been questioned and is received with scepticism by many experts. Today’s AI researchers are more focused on developing systems that are very good at tasks in a narrow range of applications. This focus is at odds with the idea of the pursuit of a super generic AI system that could mimic all different cognitive abilities related to human intelligence such as self-awareness and emotional knowledge. In addition to this debate, about AI development and the status of our hegemony as the most intelligent species on the planet, further societal concerns have been raised. For example, the AI100 (One Hundred Year Study on Artificial Intelligence) a committee led by Stanford University, defined 18 topics of importance for AI [9]. Although these are not exhaustive nor definitive, it sets forth the range of topics that need to be studied, for the potential impact of AI and stresses that there are a number of concerns to be addressed. Many similar assessments have been performed and they each outline similar concerns related to the wider adoption of AI technology.

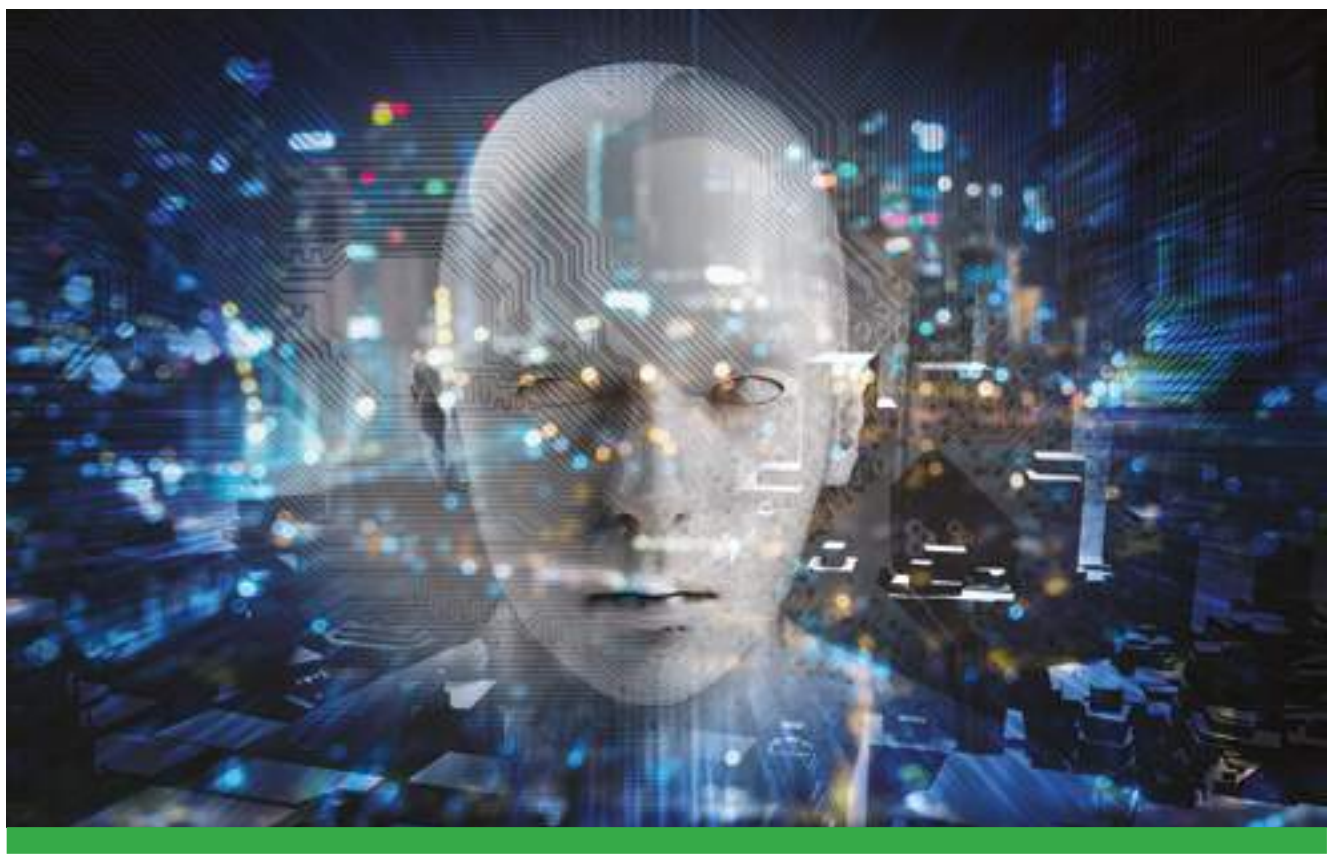

**The 18 topics covered by the AI100**

Technical trends and surprises: This topic aims at forecasting the future advances and competencies of AI technologies in the near future. Observatories of the trend and impact of AI should be created, helping to plan the setting of AI in specific sectors, and preparing the necessary regulation to smooth its introduction.

Key opportunities for AI: How advances in AI can help to transform the quality of societal services such as health, education, management and government, covering not just the economic benefits but also the social advantages and impact.

Delays with translating AI advances into real-world values: The pace of translating AI into real world applications is currently driven by potential economic prospects [10]. It is necessary to take measures to foster a rapid translation of those potential applications of AI that can improve or solve a critical need of our society, such as those that can save lives or greatly improve the organisation of social services, even though their economic exploitation is not yet assured.

Privacy and machine intelligence: Personal data and privacy is a major issue to consider and it is important to envisage and prepare the regulatory, legal and policy frameworks related to the sharing of personal data in developing AI systems.

Democracy and freedom: In addition to privacy, ethical questions with respect to the stealth use of AI for unscrupulous applications must be considered. The use of AI should not be at the expense of limiting or influencing the democracy and the freedom of people.

Law: This considers the implications of relevant laws and regulations. First, to identify which aspects of AI require legal assessment and what actions should be undertaken to ensure law enforcement for AI services. It should also provide frameworks and guidelines about how to adhere to the approved laws and policies.

Ethics: By the time AI is deployed into real world applications there are ethical concerns referring to their interaction with the world. What uses of AI should be considered unethical? How should this be disclosed?

**Economics:** The economic implications of AI on jobs should be monitored and forecasted such that policies can be implemented to direct our future generation into jobs that will not be soon overtaken by machines. The use of sophisticated AI in the financial markets could potentially cause volatilities and it is necessary to assess the influence AI systems may have on financial markets.

**AI and warfare:** AI has been employed for military applications for more than a decade. Robot snipers and turrets have been developed for military purposes [11]. Intelligent weapons have increasing levels of autonomy and there is a need for developing new conventions and international agreements to define a set of secure boundaries of the use of AI in weaponry and warfare.

**Criminal uses of AI:** Implementations of AI into malware are becoming more sophisticated thus the chances of stealing personal information from infected devices are getting higher. Malware can be more difficult to detect as evasion techniques by computer viruses and worms may leverage highly sophisticated AI techniques [12, 13]. Another example is the use of drones and their potential to fall into the hands of terrorists the consequence of which would be devastating.

**Collaboration with machines:** Humans and robots need to work together and it is pertinent to envisage in which scenarios collaboration is critical and how to perform this collaboration safely. Accidents by robots working side by side with people had happened before [14] and robotic and autonomous systems development should focus on not only enhanced task precision but in also being able to understand the environment and human intention.

**AI and human cognition:** AI has the potential for enhancing human cognitive abilities. Some relevant research disciplines with this objective are sensor informatics and human-computer interfaces. Apart from applications to rehabilitation and assisted living, they are also used in surgery [15] and air traffic control [16]. Cortical implants are increasingly used for controlling prosthesis, our memory and reasoning are increasingly relying on machines and the associated health, safety and ethical impacts must be addressed.

**Safety and Autonomy:** For the safe operation of intelligent, autonomous systems, formal verification tools should be developed to assess their safety operation. Validation can be focused on the reasoning process and verifying whether the knowledge base of an intelligent system is correct [17] and also making sure that the formulation of the intelligent behaviour will be within safety boundaries [18].

**Loss of control of AI systems:** The potential of AI being independent from human control is a major concern. Studies should be promoted to address this concern both from the technological standpoint and the relevant framework for governing the responsible development of AI.

**Psychology of people and smart machines:** More research should be undertaken to obtain detailed knowledge about the opinions and concerns people have, in the wider usage of smart machines in societies. Additionally, in the design of intelligent systems, understanding people's preferences is important for improving their acceptability [19, 20].

**Communication, understanding and outreach:** Communication and educational strategies must be developed to embrace AI technologies in our society. These strategies must be formulated in ways that are understandable and accessible by non-experts and the general public.

**Neuroscience and AI:** Neuroscience and AI can develop together. Neuroscience plays an important role for guiding research in AI and with new advances in high performance computing, there are also new opportunities to study the brain through computational models and simulations in order to investigate new hypotheses [21].

**AI and philosophy of mind:** When AI can experience a level of consciousness and self-awareness, there will be a need to understand the inner world of the psychology of machines and their subjectivity of consciousness.

# 4. A CLOSER LOOK AT THE EVOLUTION OF AI

## 4.1 SEASONS OF AI

The evolution of AI to date, has endured several cycles of optimism (springs) and pessimism or negativism (winters):

- Birth of AI (1952-1956): Before the term AI was coined, there were already advances in cybernetics and neural networks, which started to attract the attention of both the scientific communities and the public. The Dartmouth Conference (1956) was the result of this increasing interest and gave rise to the following golden years of AI with high levels of optimism in the field.
- First spring (1956-1974): Computers of the time could solve algebra and geometric problems, as well as speak English. Advances were qualified as "impressive" and there was a general atmosphere of optimism in the field. Researchers in the area estimated that a fully intelligent machine would be built in the following 20 years.
- First winter (1974-1980): The winter started when the public and media questioned the promises of AI. Researchers were caught in a spiral of exaggerated claims and forecasts but the limitations the technology posed at the time were inviolable. An abrupt ending of funding by major agencies such as DARPA, the National Research Council and the British Government, led to the first winter of AI.
- Second Spring (1980-1987): Expert systems were developed to solve problems of a specific domain by using logical rules derived from experts. There was also a revival of connectionism and neural networks for character or speech recognition. This period is known as the second spring of AI.
- Second winter (1987-1993): Specialised machines for running expert systems were displaced by new desktop computers. Consequently some companies, that produced expert systems, went into bankruptcy. This led to a new wave of pessimism ending the funding programs initiated during the previous spring.
- In the background (1997-2000): From 1997 to 2000, the field of AI was progressing behind the scenes, as no further multi-million programs were announced. Despite the lack of major funding the area continued to progress, as increased computer power and resources were developed. New applications in specific areas were developed and the concept of "machine learning" started to become the cornerstone of AI.
- Third spring (2000-Present): Since 2000, with the success of the Internet and web, the Big Data revolution started to take off along with newly emerged areas such as Deep Learning. This new period is known as the third spring of AI and for time being, it looks like it is here to stay. Some have even started to predict the imminent arrival of singularity - an intelligence explosion resulting in a powerful super-intelligence that will eventually surpass human intelligence. Is this possible?

## 4.2 INFLUENCE OF FUNDING

Government organisations and the public sector are investing millions to boost artificial intelligence research. For example, the National Research Foundation of Singapore is investing $150 million into a new national programme in AI. In the UK alone, £270 million is being invested from 2017 to 2018 to boost science, research and innovation, via the Government's new industrial strategy and a further funding of £4.7 billion is planned by 2021 [22]. This timely investment will put UK in the technological lead among the best in the world and ensure that UK technological innovations can compete. Recent AI developments have triggered major investment across all sectors including financial services, banking, marketing and advertising, in hospitals and government administration. In fact software and information technology services have more than a 30% share in all AI investments worldwide as of 2016, whereas Internet and telecommunication companies follow with 9% and 4%, respectively [23].

It is also important to note that the funding in AI safety, ethics and strategy/policy has almost doubled in the last three years [24]. Apart from non-profit organisations, such as the Future of Life Institute (FLI) and the Machine Intelligence Research Institute (MIRI), other centres, such as the Centre for Human-Compatible AI and Centre for the Future of Intelligence, have emerged and they, along with key technological firms, invested a total of $6.6 million in 2016.

## 4.3 PUBLICATION VERSUS PATENTING

In terms of international output in publications and patents, there has been a shift of predominant countries influencing the field of AI. In 1995 USA and Japan were the two leading countries in the field of AI patents but this has now shifted to Asia, with China becoming a major player. Since 2010 China and USA have led the way in both scientific publications and in the filing of patents. Other emerging economies, such as India and Brazil, are also rapidly rising.

Recently, there has been a shift relocation of many academic professionals in AI to the industrial sector. The world's largest technology companies have hired several entire research teams previously in universities. These corporations have opted for the publication of pre-prints (ArXiv[1] or viXra[2]) and other non-citable documents, instead of using conventional academic methods of peer-review. This has become an increasing trend. The reason for this is that it allows the prioritisation of claim imprinting without the delay of a peer-review process.

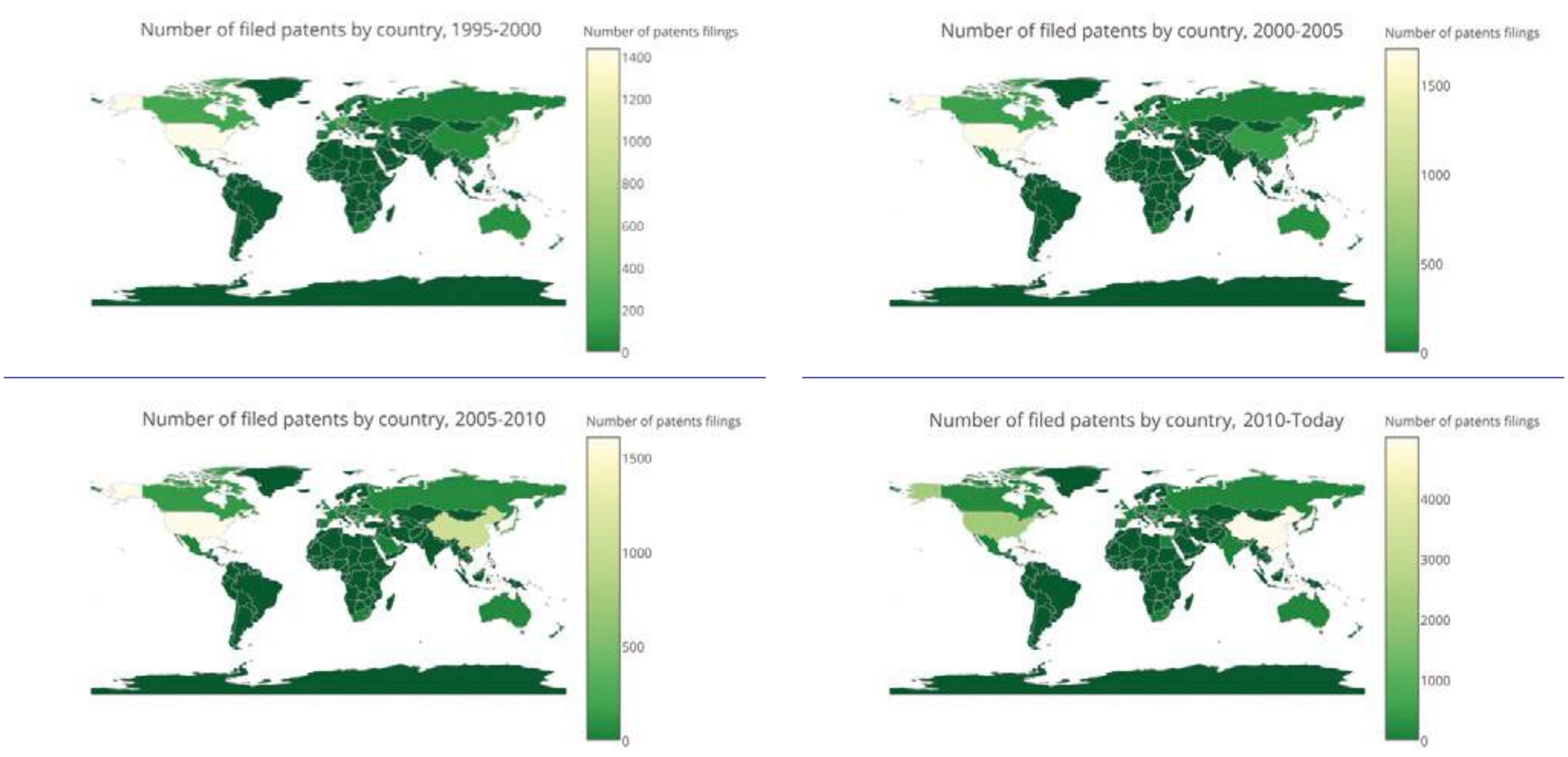

**Figure 2**
Evolution of filed patents in AI across countries for five-year periods since 1995 to present day.

[1] https://arxiv.org/
[2] vixra.org

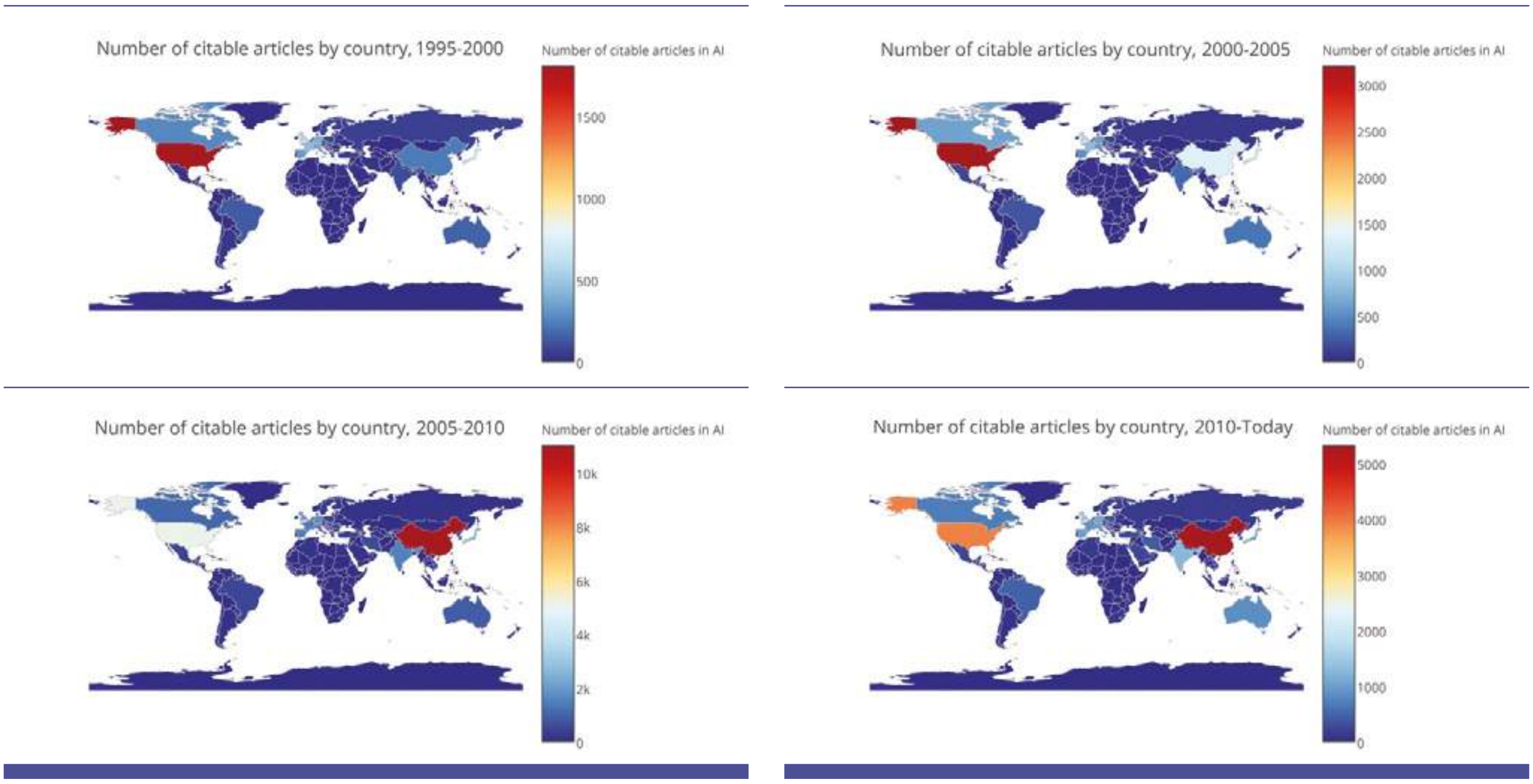

**Figure 3**
Evolution of citable articles in AI across countries for five-year periods since 1995 to present day.

| Time series (yearly) | Database | Publisher | Query |
|---|---|---|---|
| Number of publications of AI by computational intelligence methods | Web of Science | Thomson Reuters [25] | {("artificial neural networks") OR ("evolutionary computation") OR ("evolutionary algorithms") OR ("evolutionary programing") OR ("fuzzy systems") OR ("fuzzy logic") OR ("fuzzy set") OR ("neuro-fuzzy") OR ("bayesian inference") OR ("statistical inference") OR ("graphical model") OR ("markov model") OR ("bayesian network") OR ("gaussian process") OR ("logic programing") OR ("inductive programming") OR ("deep learning")} AND { ("artificial intelligence") OR ("machine learning")} |
| Number of publications of AI by countries | Scimago SJR | Scimago Lab Scopus [26] | |
| Patents in AI | DOCDB database from the EPO (European Patent Office) 102 Countries | Patent inspiration [27] | ANY in ("artificial intelligence") OR ("machine learning") OR ("deep learning") |

**Table 1.**
Sources of information used for Figures 2 and 3.

# 5. FINANCIAL IMPACT OF AI

It has been well recognised that AI amplifies human potential as well as productivity and this is reflected in the rapid increase of investment across many companies and organisations. These include sectors in healthcare, manufacturing, transport, energy, banking, financial services, management consulting, government administration and marketing/advertising. The revenues of the AI market worldwide, were around 260 billion US dollars in 2016 and this is estimated to exceed $3,060 billion by 2024 [23]. This has had a direct effect on robotic applications, including exoskeletons, rehabilitation, surgical robots and personal care-bots. The economic impact of the next 10 years is estimated to be between $1.49 and $2.95 trillion. These estimates are based on benchmarks that take into account similar technological achievements such as broadband, mobile phones and industrial robots [28]. The investment from the private sector and venture capital is a measure of the market potential of the underlying technology. In 2016, a third of the shares from software and information technology have been invested in AI, whereas in 2015, 1.16 billion US dollars were invested in start-up companies worldwide, a 10-fold increase since 2009.

Major technological firms are investing into applications for speech recognition, natural language processing and computer vision. A significant leap in the performance of machine learning algorithms resulting from deep learning, exploited the improved hardware and sensor technology to train artificial networks with large amounts of information derived from ‘big data’ [31, 32]. Current state-of-the-art AI allows for the automation of various processes and new applications are emerging with the potential to change the entire workings of the business world. As a result, there is huge potential for economic growth, which is demonstrated in the fact that between 2014 and 2015 alone, Google, Microsoft, Apple, Amazon, IBM, Yahoo, Facebook, and Twitter, made at least 26 acquisitions of start-ups and companies developing AI technology, totalling over $5 billion in cost.

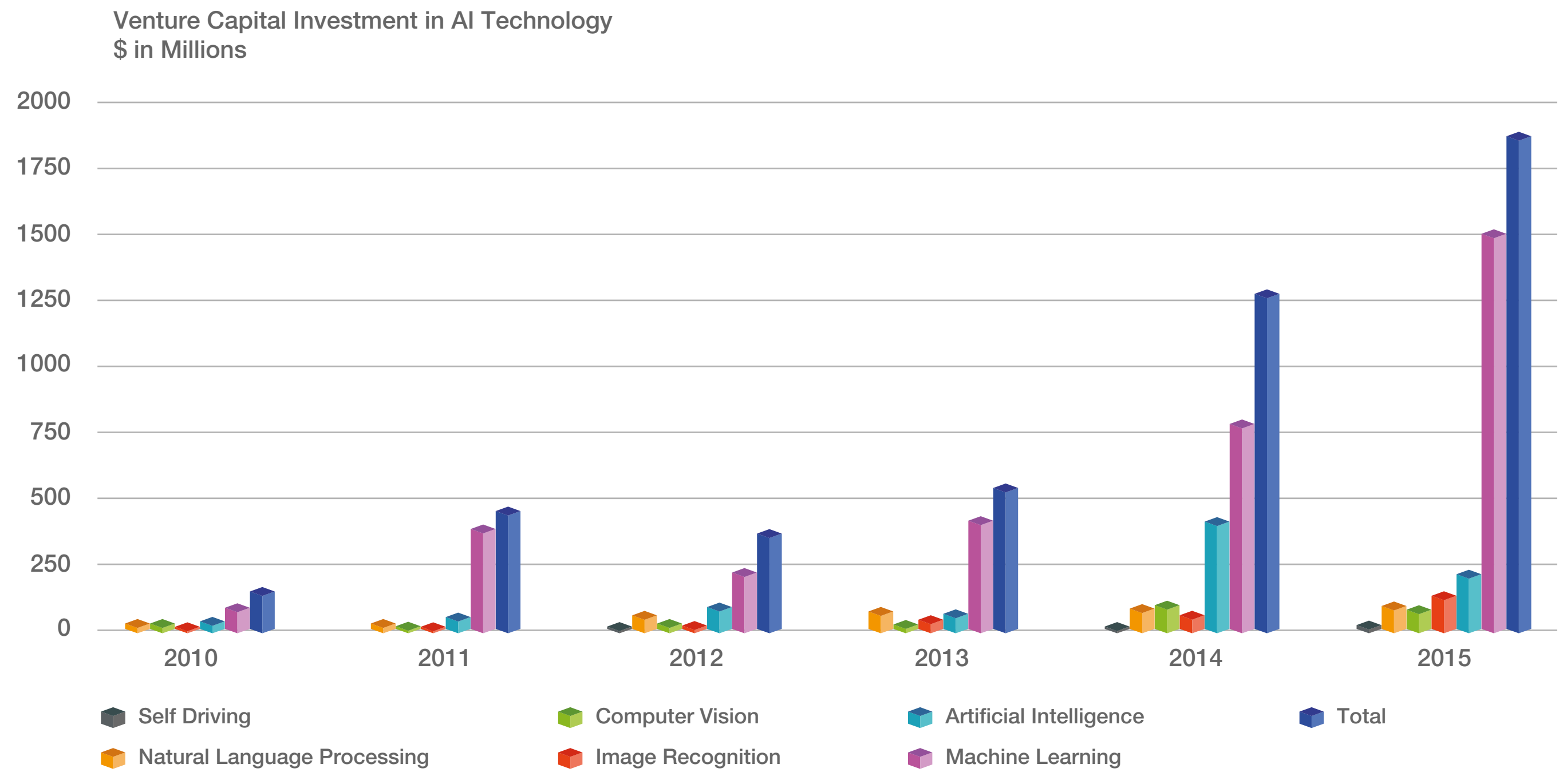

**Figure 4**
A conservative estimate of venture capital investment in AI technology worldwide according to data presented in [28].

In 2014, Google acquired DeepMind, a London-based start-up company specialising in deep learning, for more than $500M and set a record of company investment of AI research to academic standard. In fact, DeepMind has produced over 140 journal and conference papers and has had four articles published in Nature since 2012. One of the achievements of DeepMind was in developing AI technology able to create general-purpose software agents that adjust their actions based only on a cumulative reward. This reinforcement learning approach exceeds human level performance in many aspects and has been demonstrated with the defeat of the world Go game champion; marking a historical landmark in AI progress.

IBM has developed a supercomputer platform, Watson, which has the capability to perform text mining and extract complex analytics from large volumes of unstructured data. To demonstrate its abilities, IBM Watson, in 2011, beat two top players on 'Jeopardy!', a popular quiz show, that requires participants to guess questions from specific answers. Although, information retrieval is trivial for computer systems, comprehension of natural language is still a challenge. This achievement has had a significant impact on the performance of web searches and the overall ability of AI systems to interact with humans. In 2015, IBM bought AlchemyAPI to incorporate its text and image analysis capabilities in the cognitive computing platform of the IBM Watson. The system has already been used to process legal documents and provide support to legal duties. Experts believe that these capabilities can transform current health care systems and medical research.

Research in top AI firms is centred on the development of systems that are able to reliably interact with people. Interaction takes more natural forms through real-time speech recognition and translation capabilities. Robo-advisor applications are at the top of the AI market with a globally estimated 255 billion in US dollars by 2020 [23]. There are already several virtual assistants offered by major companies. For example, Apple offers Siri and Amazon Alexa, Microsoft offers Cortana, and Google has the Google Assistant. In 2016, Apple Inc. purchased Emotient Inc., a start-up using artificial-intelligence technology to read people's emotions by analyzing facial expressions. DeepMind created WaveNet, which is a generative model that mimics human voices. According to the company's website, this sounds more natural than the best existing Text-to-Speech systems. Facebook is also considering machine-human interaction capabilities as a prerequisite to generalised AI.

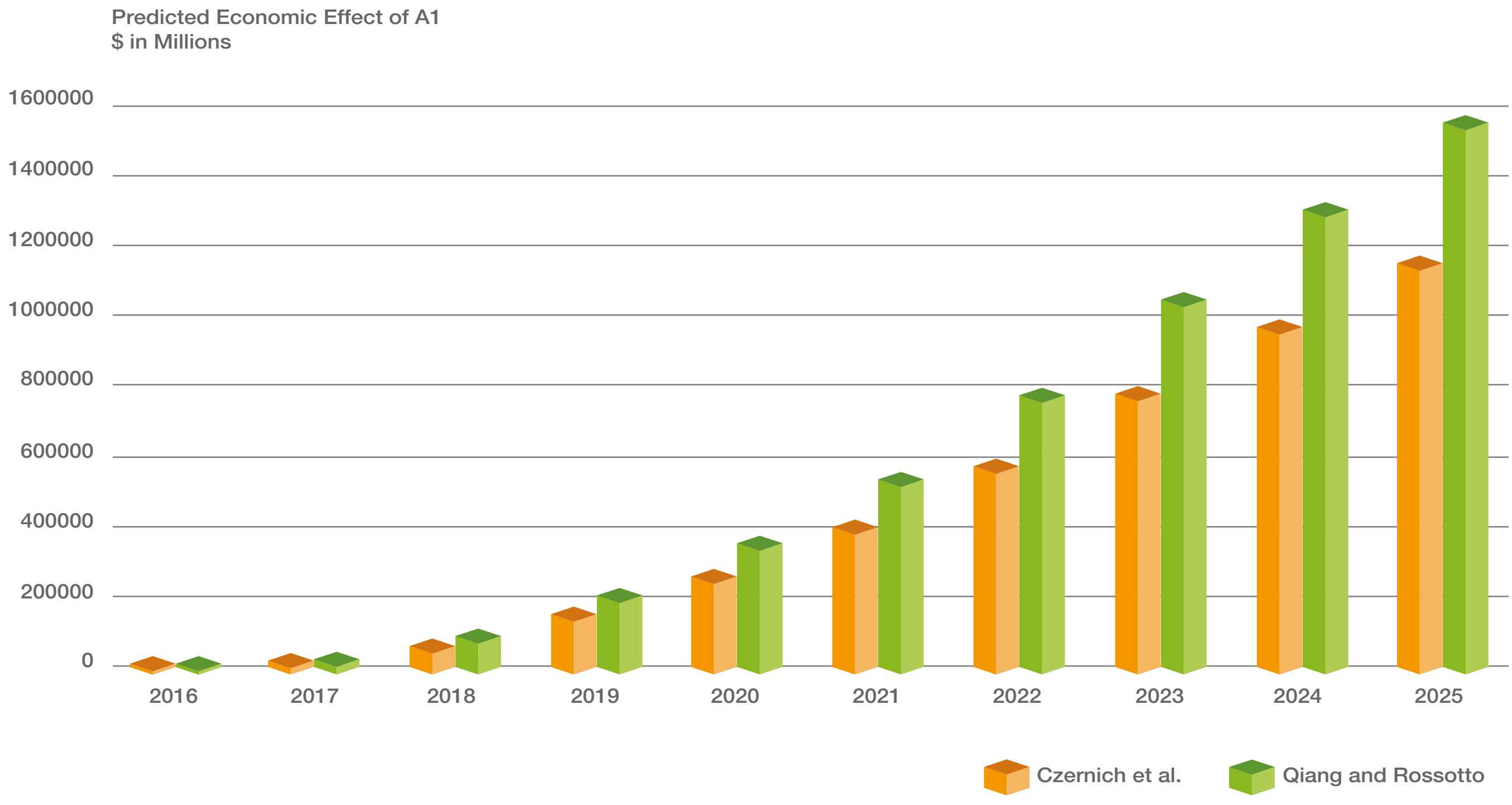

**Figure 5**
Predicted economic effect of AI worldwide estimated based on the GDP of mature economies and benchmark data from broadband Internet economic growth [29, 30].

Recently, OpenAI, a non-profit organisation, has been funded as part of a strategic plan to mitigate the risks of monopolising strong AI. OpenAI has re-designed evolutional algorithms that can work together with deep neural networks to offer state-of-the-art performance. It is considered to rival DeepMind since it offers similar open-source machine learning libraries to TensorFlow, a deep learning library distributed by Google DeepMind. Nevertheless, the big difference between the technology developed at OpenAI and the other private tech companies, is that the created Intellectual Property is accessible by everyone.

Although several companies and organisations, including DeepMind and OpenAI, envision the solution to the creation of intelligence and the so-called Strong AI, developing machines with self-sustained long-term goals is well beyond current technology. Furthermore, there is vigorous debate on whether or not we are going through an AI bubble, which encompasses the paradox that productivity growth in USA, during the last decade, has declined regardless of an explosion of technological progress and innovation. It is difficult to understand whether this reflects a statistical shortcoming or that current innovations are not transformative enough. This decline can be also attributed to the lack of consistent policy frameworks and security standards that can enable the application of AI in projects of significant impact.

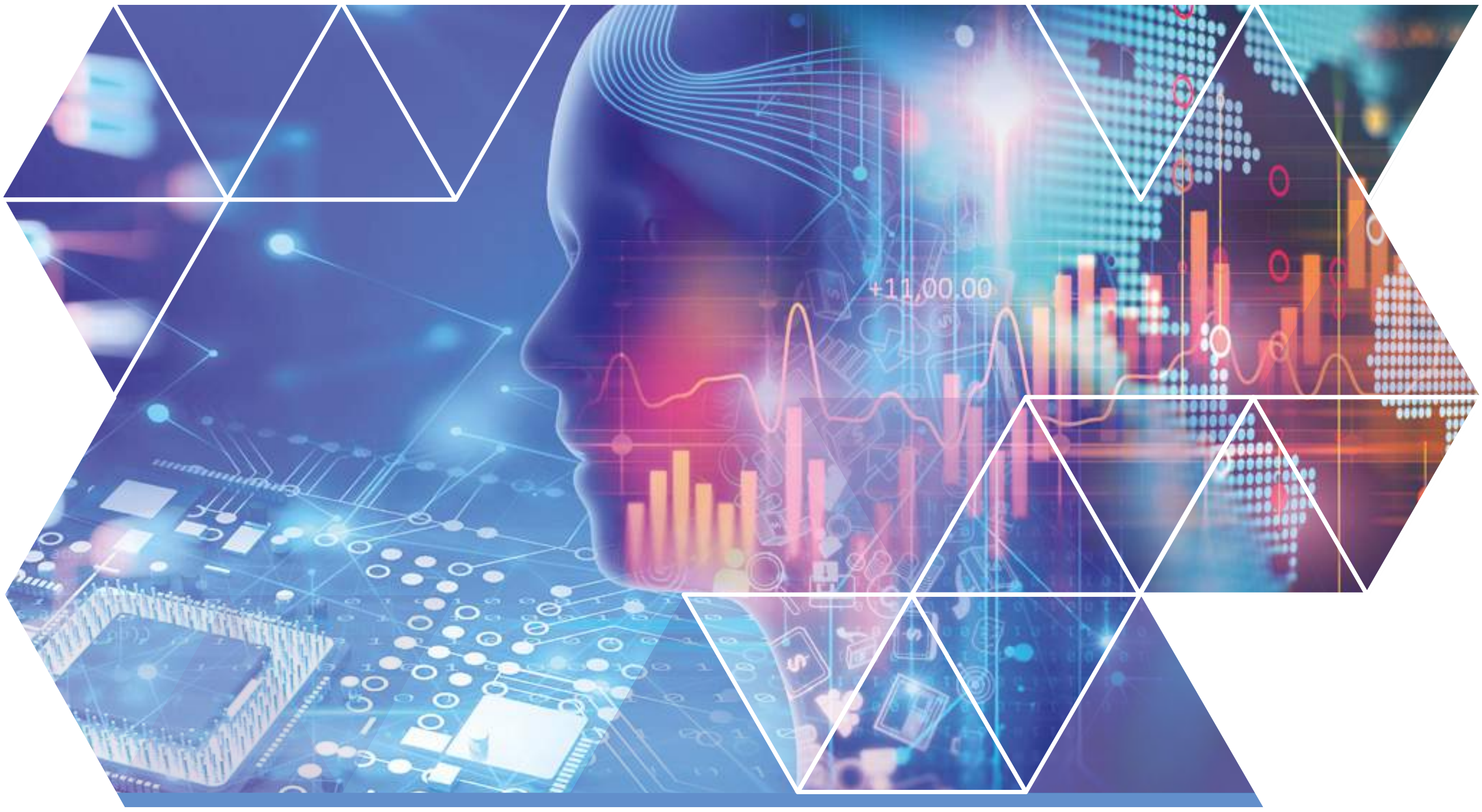

**Table 2.**
Major companies in AI

| | Technology/Platforms | AI Applications of significant impact | Open-Source |
|---|---|---|---|
| Google DeepMind | Search engine, Maps, Ads, Gmail, Android, Chrome, and YouTube | **Self-driving cars:** Technology that allows a car to navigate in normal traffic without any human control. | **TensorFlow:** Construction of Deep Neural Networks |
| | Deep Q-network: Deep Neural Networks with Reinforcement Learning at scale. | **AlphaGo:** The first computer program to beat professional players of Go.<br>**DQN:** Better than human-level control of Atari games through Deep Reinforcement Learning.<br>**Wavenet:** Raw audio form impersonating any human voice | **DeepMind Lab:** 3D game-like platform for agent-based AI research<br>**Sonnet:** Constructing Deep Neural Networks based on TensorFlow |
| OpenAI | Non-profit organisation<br>Evolutionary Algorithms<br>Deep Neural Networks | **Evolutionary Algorithms** tuned to work with Deep Neural Networks<br>**Testbeds for AI:** Benchmarking tools and performance measures for AI algorithms. | **Gym:** Toolkit for developing and comparing reinforcement learning algorithms.<br>**Universe:** Measure an AI's general Intelligence |
| IBM | Manufacturer of computer hardware and software<br>Hosting and consulting services<br>Cognitive Computing | **Deep Blue:** First computer program to defit world champion chess player<br>**Watson:** Won top players on 'Jeopardy!', a popular quiz show. | **Apache SystemML:** Distribution of large-scale machine learning computations on Apache Hadoop and Spark.<br>**Apache UIMA:** Unstructured Information Management |
| facebook | Social Networking Service | **Applied Machine Learning:** Spot suicidal users<br>**Human Computer Interaction:** Image Descriptions for Blind Users | **CommAI-env:** A Communication-based platform for training and evaluating AI systems.<br>**fbcunn:** Deep learning modules for GPUs |
| Apple Inc. | Computer hardware and software<br>Consumer electronics<br>Online services | **Siri:** AI Virtual Assistant<br>**Self-driving car:** AI technology that could drive a car without human interaction. | |
| amazon | Cloud Computing<br>Online retail services<br>Electronics | **Alexa:** AI virtual assistant<br>**Amazon AI platform:** Cloud software and hardware AI tools | **DSSTNE:** Deep Scalable Sparse Tensor Network Engine |
| Microsoft | Developing, manufacturing and licensing computer hardware and software<br>Consumer electronics | **Microsoft Azure:** Cloud services<br>**Cortana:** AI virtual assistant | **CNTK:** Cognitive Toolkit<br>**Microsoft Azure:** Cloud computing platform offered as a service. |

- Natural language processing
- Machine vision
- Robotics
- Machine learning
- Others

## Start-up acquisition

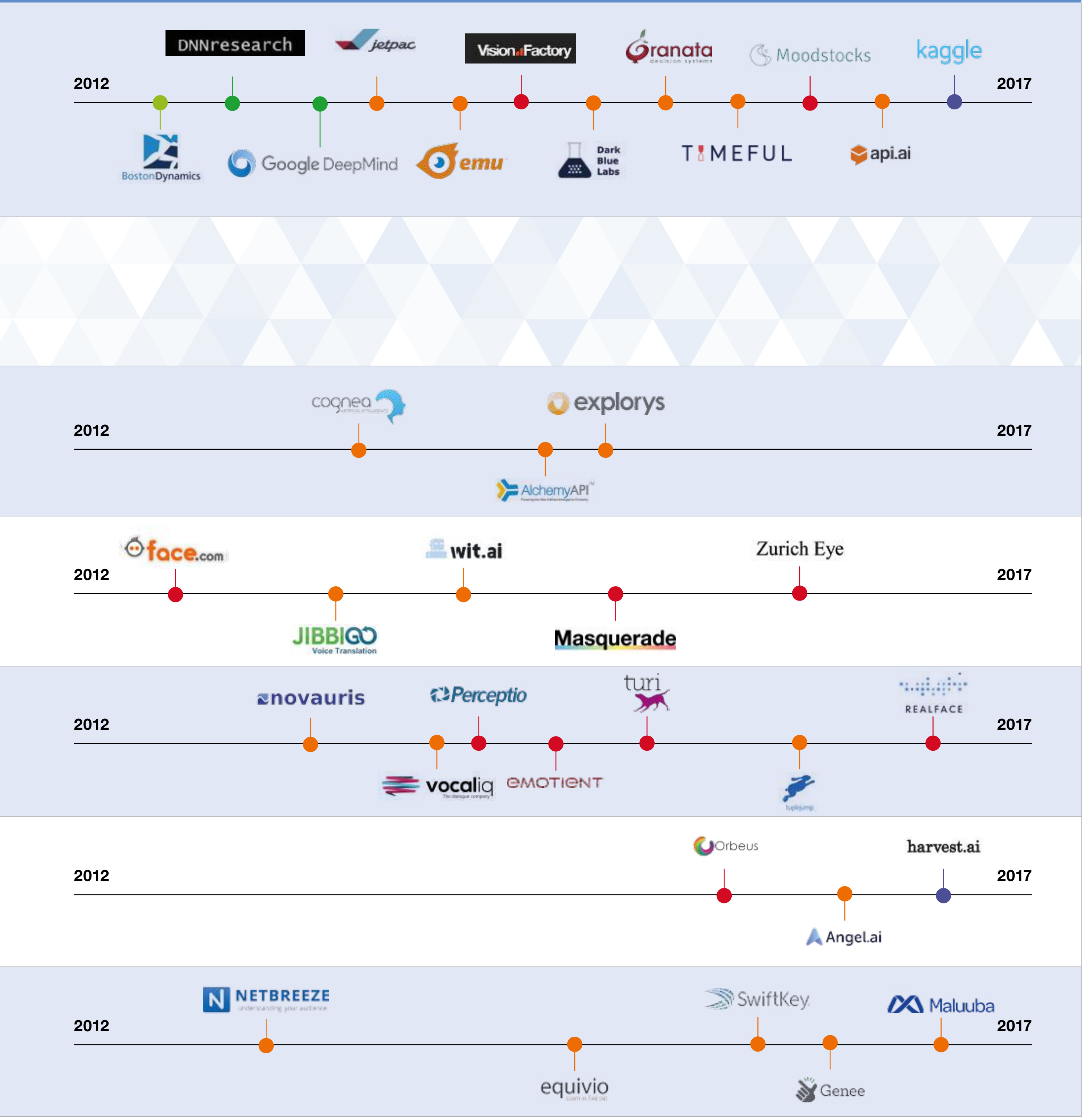

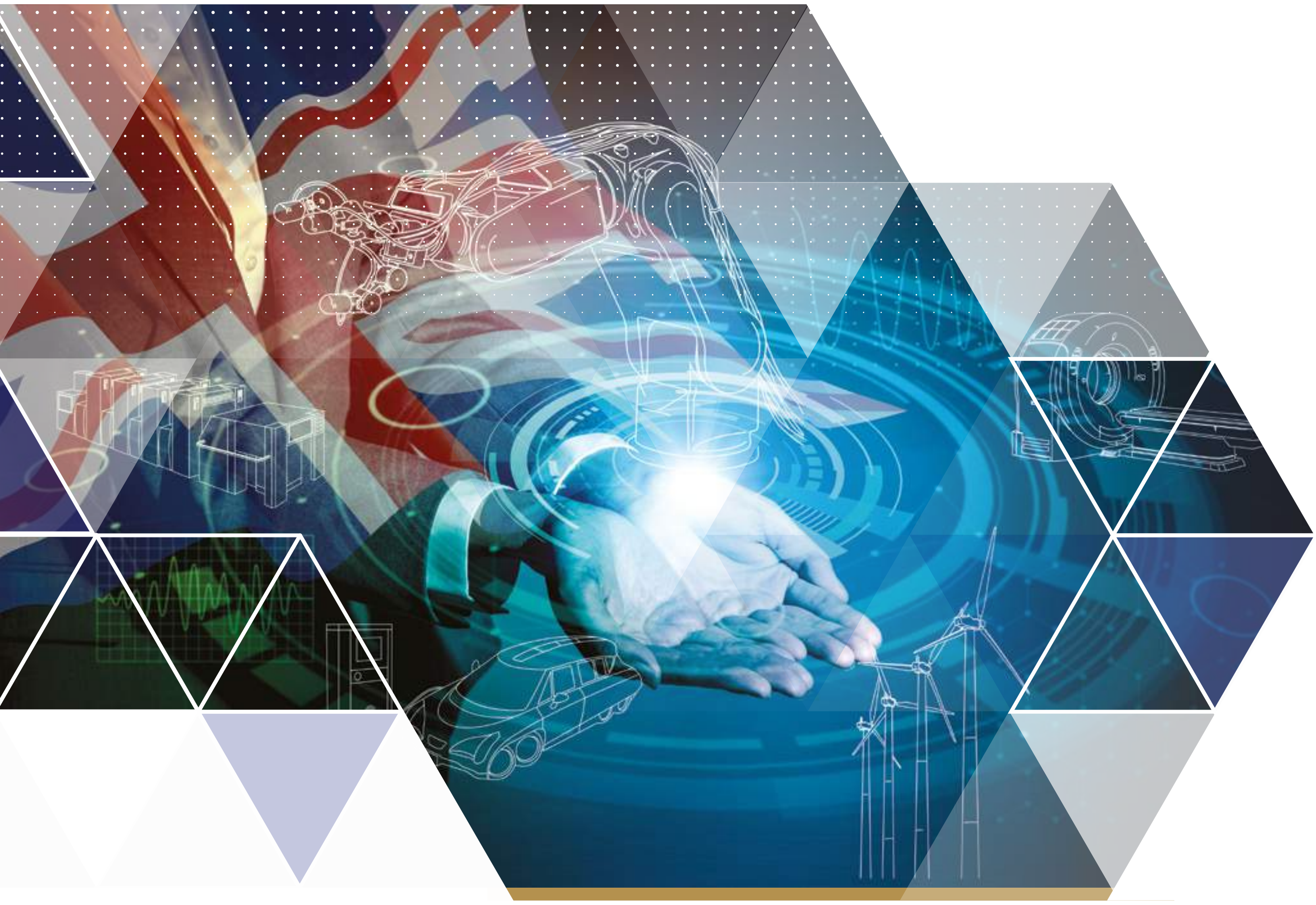

"For the first time, the UK government has singled out robotics and AI in its blueprint for a 'modern' industrial strategy. It brings some certainty in this uncertain time, demonstrating the UK's drive to kick-start disruptive technologies that could transform our economy, with a clear vision for positioning the UK in the international landscape."

# 6. SUBFIELDS AND TECHNOLOGIES THAT UNDERPINNINGS ARTIFICIAL INTELLIGENCE

AI is a diverse field of research and the following subfields are essential to its development. These include neural networks, fuzzy logic, evolutionary computation, and probabilistic methods.

*Neural networks* build on the area of connectionism with the main purpose of mimicking the way the nervous system processes information. Artificial Neural Networks (ANN) and variants have allowed significant progress of AI to perform tasks relative to “perception”. When combined with the current multicore parallel computing hardware platforms, many neural layers can be stacked to provide a higher level of perceptual abstraction in learning its own set of features, thus removing the need for handcrafted features; a process known as deep learning [33]. Limitations of using deep layered ANN include 1) low interpretability of the resultant learned model, 2) large volumes of training data and considerable computational power are often required for the effective application of these neural models.

*Deep learning* is part of machine learning and is usually linked to deep neural networks that consist of a multi-level learning of detail or representations of data. Through these different layers, information passes from low-level parameters to higher-level parameters. These different levels correspond to different levels of data abstraction, leading to learning and recognition. A number of deep learning architectures, such as deep neural networks, deep convolutional neural networks and deep belief networks, have been applied to fields such as computer vision, automatic speech recognition, and audio and music signal recognition and these have been shown to produce cutting-edge results in various tasks.

*Fuzzy logic* focuses on the manipulation of information that is often imprecise. Most computational intelligence principles account for the fact that, whilst observations are always exact, our knowledge of the context, can often be incomplete or inaccurate as it is in many real-world situations. Fuzzy logic provides a framework in which to operate with data assuming a level of imprecision over a set of observations, as well as structural elements to enhance the interpretability of a learned model [34]. It does provide a framework for formalizing AI methods, as well as an accessible translation of AI models into electronic circuits. Nevertheless, fuzzy logic does not provide learning abilities per se, so it is often combined with other aspects such a neural networks, evolutionary computing or statistical learning.

*Evolutionary computing* relies on the principle of natural selection, or natural patterns of collective behaviour [35]. The two most relevant subfields include genetic algorithms and swarm intelligence. Its main impact on AI is on multi-objective optimization, in which it can produce very robust results. The limitations of these models are like neural networks about interpretability and computing power.

*Statistical Learning* is aimed at AI employing a more classically statistical perspective, e.g., Bayesian modelling, adding the notion of prior knowledge to AI. These methods benefit from a wide set of well-proven techniques and operations inherited from the field of classical statistics, as well as a framework to create formal methods for AI. The main drawback is that, probabilistic approaches express their inference as a correspondence to a population [36] and the probability concept may not be always applicable, for instance, when vagueness or subjectivity need to be measured and addressed [37].

*Ensemble learning and meta-algorithms* is an area of AI that aims to create models that combine several weak base learners in order to increase accuracy, while reducing its bias and variance. For instance, ensembles can show a higher flexibility with respect to single model approaches on which some complex patterns can be modelled. Some well-known meta-algorithms for building ensembles are bagging and boosting. Ensembles can take advantage of significant computational resources to train many base classifiers therefore enhancing the ability to augment resolution of the pattern search - although this does not always assure the attainment of a higher accuracy.

*Logic-based artificial intelligence* is an area of AI commonly used for task knowledge representation and inference. It can represent predicate descriptions, facts and semantics of a domain by means of formal logic, in structures known as logic programs. By means of inductive logic programming hypotheses can be derived over the known background.

# 7. THE RISE OF DEEP LEARNING: RETHINKING THE MACHINE LEARNING PIPELINE

The idea of creating an artificial machine is as old as the invention of the computer. Alan Turing in the early 1950s proposed the Turing test, designed to assess whether a machine could be defined as intelligent. Two of the main pioneers in this field are Pitts and McCulloch [38] who, in 1943, developed a technique designed to mimic the way a neuron works. Inspired by this work, a few years later, Frank Rosenblatt [39] developed the first real precursor of the modern neural network, called Perceptron. This algorithm describes an automatic learning procedure that can discriminate linearly separable data. Rosenblatt was confident that the perceptron would lead to an AI system in the future. The introduction of perceptron, in 1958, signalled the beginning of the AI evolution. For almost 10 years afterwards, researchers used this approach to automatically learn how to discriminate data in many applications, until Papert and Minsky [3], demonstrated a few important limitations of Perceptron. This slowed down the fervour of AI progress and more specifically, they proved that the perceptron was not capable of learning simple functions, such as the exclusive-or XOR, no matter how long the network was trained.

Today, we know that the model implied by the perceptron is linear and the XOR function does not belong to this family, but at the time this was enough to stop the research behind neural nets and began the first AI winter. Much later in 1974, the idea of organizing the perceptron in layers and training them using the delta rule [40] shaped the creation of more complex neural nets. With the introduction of the Multilayer Neural Nets [41], researchers were confident that adding multiple hidden layers to the networks would produce deep architectures that further increase the complexity of the hypothesis that can be expressed. However, the hardware constraints that were present at that time, limited, for many years, the number of layers that could be used in practice. To overcome these hardware limitations different network configurations were proposed. For almost another decade, researchers focused on producing new efficient network architectures that are suitable for specific contexts. Notably, these developments included the Autoencoder [42] useful in extracting relevant features from data, the Belief nets used to model statistical variables, the Recurrent neural nets [43] and its variant Long Short Term Memory [44] used for processing sequence of data, and the Convolutional neural nets [45] used to process images. Despite these new AI solutions, the aforementioned hardware limitations were a big restriction during training.

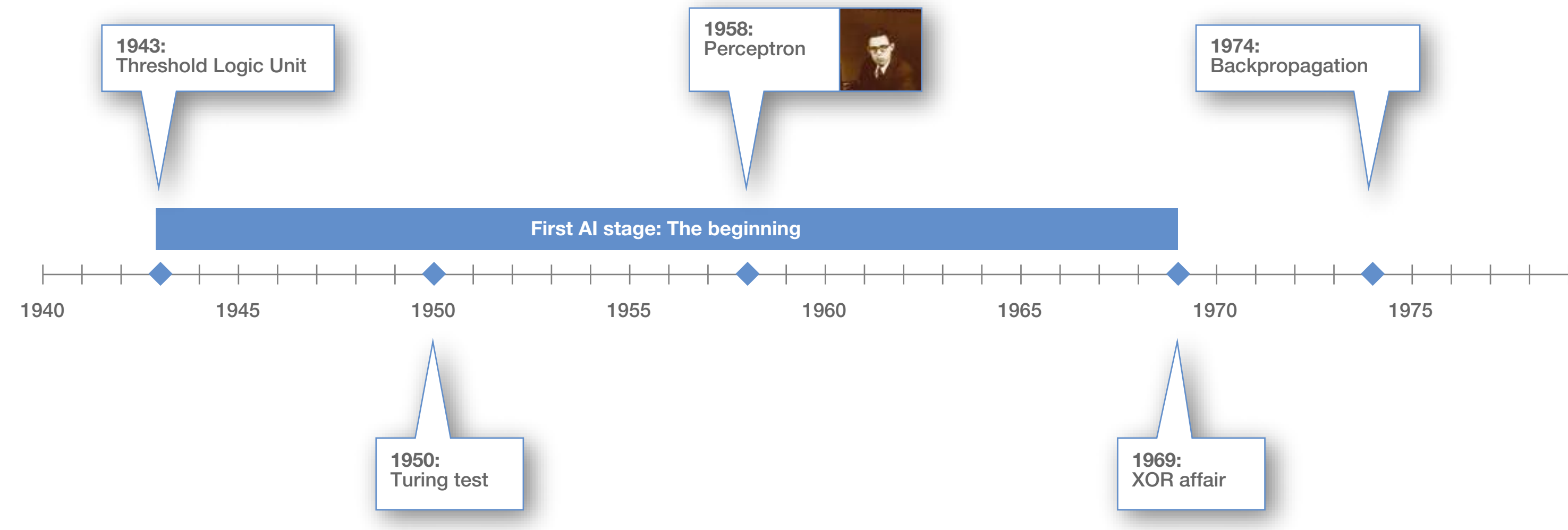

**Figure 6**
Timeline of the main discoveries and seasons in AI: from perceptron to deep learning

With recent hardware advances, such as the parallelization using GPU, the cloud computing and the multi-core processing finally led to the present stage of AI. In this stage, deep neural nets have made tremendous progress in terms of accuracy and they can now recognize complex images and perform voice translation in real time. However, researchers are still dealing with issues relating to the overfitting of the networks, since large datasets are often required and not always available. Furthermore, with the vanishing of the gradient, this leads to a widespread problem generated during the training of a network with many layers. For this reason, more sophisticated training procedures have recently been proposed. For example, in 2006, Hinton introduced the idea of unsupervised pretraining and Deep Belief Nets [46]. This approach has each pair of consecutive layers trained separately using an unsupervised model similar to the one used in the Restricted Boltzman Machine [47]; then the obtained parameters are frozen and a new pair of layers are trained and stacked on top of the previous ones. This procedure can be repeated many times leading to the development of a deeper architecture with respect to the traditional neural nets. Moreover, this unsupervised pre-training approach has led to increasing neural net papers when in 2014, for the first time, a neural model became state-of-the-art in the speech recognition.

In 2010, a large database, known as Imagenet containing millions of labelled images was created and this was coupled with an annual challenge called Large Scale Visual Recognition Challenge. This competition requires teams of researchers to build AI systems and they receive a score based on how accurate their model is. In the first two years of the contest, the top models had an error rate of 28% and 26%. In 2012, Krizhevsky, Sutskever and Hinton [48] submitted a solution that had an error rate of just 16% and in 2015 the latest submitted models [49] were capable of beating the human experts with an overall error of 5%. One of the main components of this significant improvement, in such a short time, was due to the extensive use of graphics processing units (GPUs) for speeding up the training procedure, thus allowing the use of larger models which also meant a lower error rate in classification.

In the last 3 years researchers have also been working on training deep neural nets that are capable of beating human experts in different fields, similar to the solution used for AlphaGo [50] or DeepStack [51]and in 2017, they overtook human experts with 44000 played hands of poker.

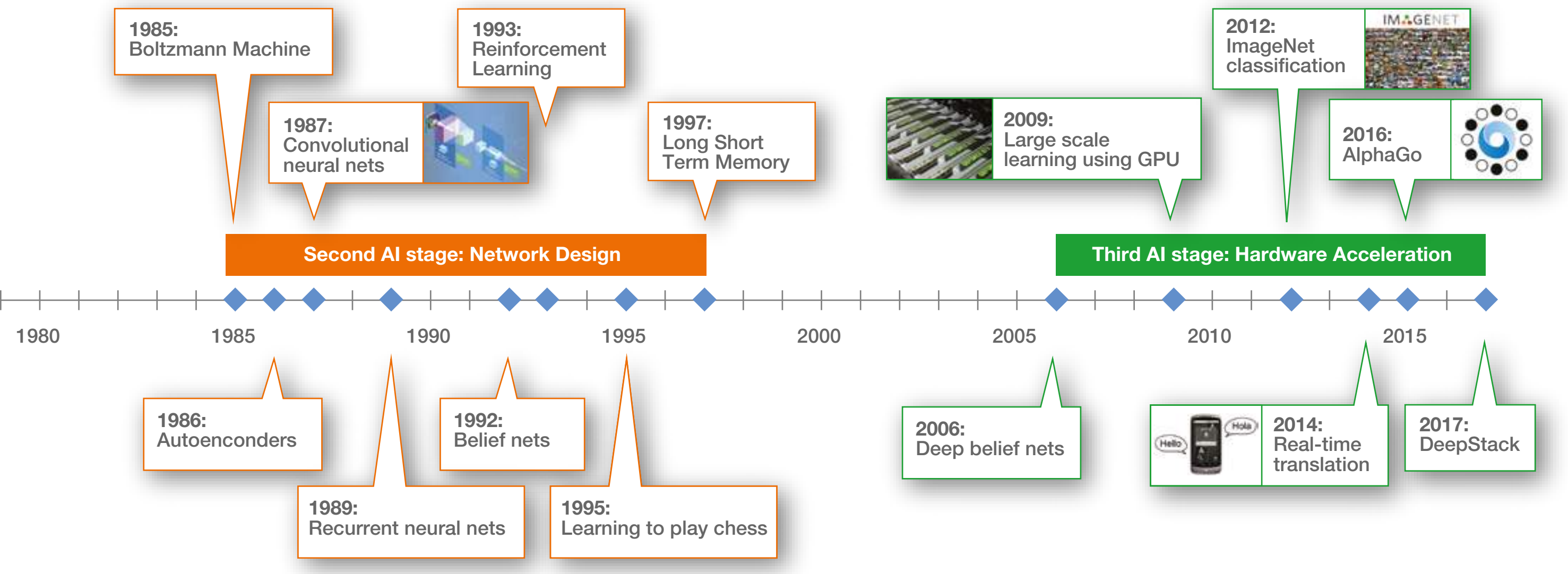

1985:
**Boltzmann Machines**

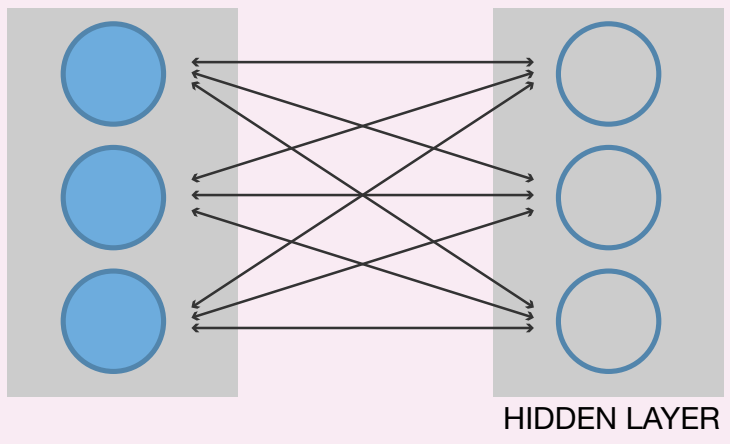

Boltzmann Machines represent a type of neural network modelled by using stochastic units with a specific distribution (for example Gaussian). Learning procedure involves several steps called Gibbs sampling, which gradually adjust the weights to minimize the reconstruction error. They are useful if it is required to model probabilistic relationships between variables. A variant of this machine is the Restricted Boltzmann Machines where the visible and hidden units are restricted to form a bipartite graph that allows implementation of more efficient training algorithms.

1986:
**AutoEncoder**

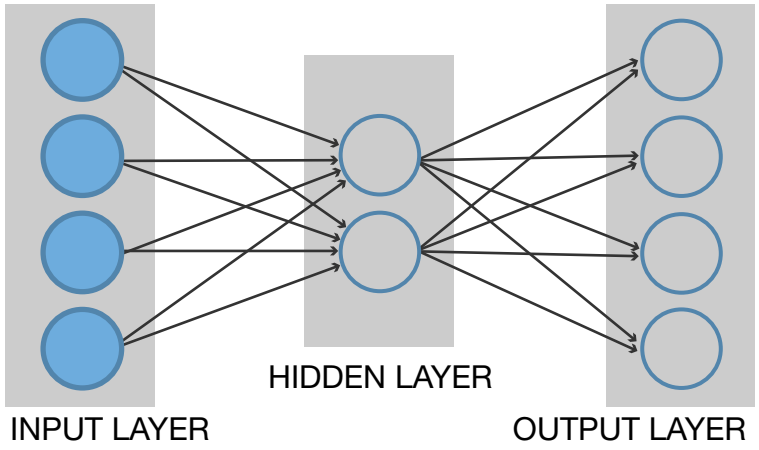

An Autoencoder is a neural network designed to extract features directly from the data. This network has the same number of input and output nodes and it is trained using an unsupervised approach to recreate the input vector rather than to assign a class label to it. Usually, the number of hidden units is smaller than the input/output layers, which achieve encoding of the data in a lower dimensional space and extract the most discriminative features.

1987:
**Convolutional neural network (CNN)**

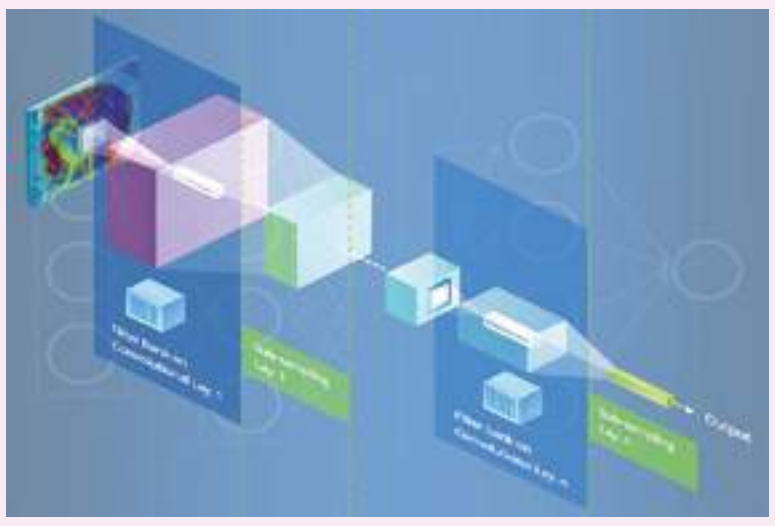

CNNs have been proposed to process efficiently imagery data. The name of these networks comes from the convolution operator that provides an easy way to perform complex operations using convolution filter. CNNs use locally connected neurons that represent data specific kernels. The main advantage of a CNN is that during back-propagation, the network has to adjust a number of parameters equal to a single instance of the kernel which drastically reduces the connections from the typical neural network. The concept of CNN is inspired by the neurobiological model of the visual cortex and can be briefly summarized as a sequence of convolution and subsampling of the image until high level features can be extracted.

1989:
**Recurrent neural nets (RNN)**

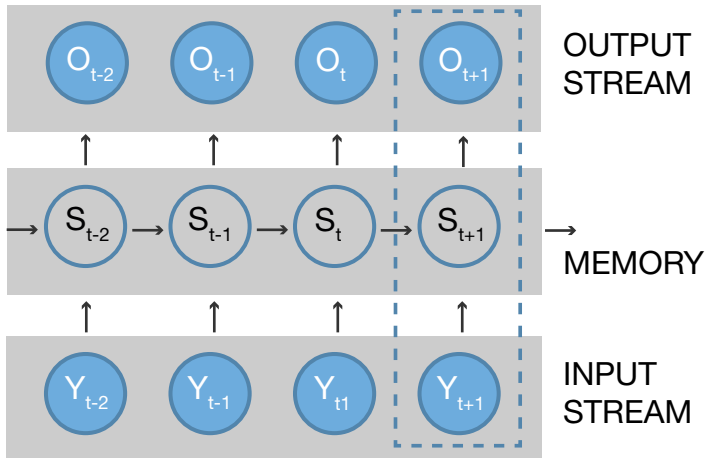

RNN is a neural network that contains hidden units capable of analysing streams of data. Since RNN suffers from the vanishing gradient and exploding gradient problems, a variation called Long Short-Term Memory units (LSTMs) was proposed in 1997 to solve this problem. Specifically, LSTM is particularly suitable for applications where there are very long time lags of unknown sizes between important events. RNN and LSTM share the same weights across all steps that greatly reduce the total number of parameters that the network needs to learn. RNNs have shown great successes in many Natural Language Processing tasks such as language modelling, bioinformatics, speech recognition and generating image description.

# 8. HARDWARE FOR AI

In 1965, Gordon Moore observed that the number of transistors, in a dense integrated circuit, doubles approximately every year. Ten years later, he revised his forecast, updating his prediction to the number doubling every two years. Moore's prediction has been accurate for several decades and has been used in the semiconductor industry to guide long-term planning. In 2015, Moore realised that the rate of progress in the hardware would reach saturation and the transistors would arrive at the limits of miniaturisation at the atomic level. Experts estimate that Moore's law could end in 2025. Today, his prediction is still valid and the number of transistors is increasing even if, after 2005, the frequency and the power started to reduce, leading to a core scaling rather than a frequency improvement. Therefore, since 2005, we are no longer getting faster computers, but the hardware is designed in a multi-core manner. To take full advantage of this different hardware implementation, the software has to be written in a multi-threaded manner too. In future, experts believe that revolutionary technologies may help sustain Moore's law. One of the key challenges will be the design of gates in nanoscale transistors and the ability of controlling the current flow as, when the device dimension shrinks, the connection between transistors becomes more difficult.

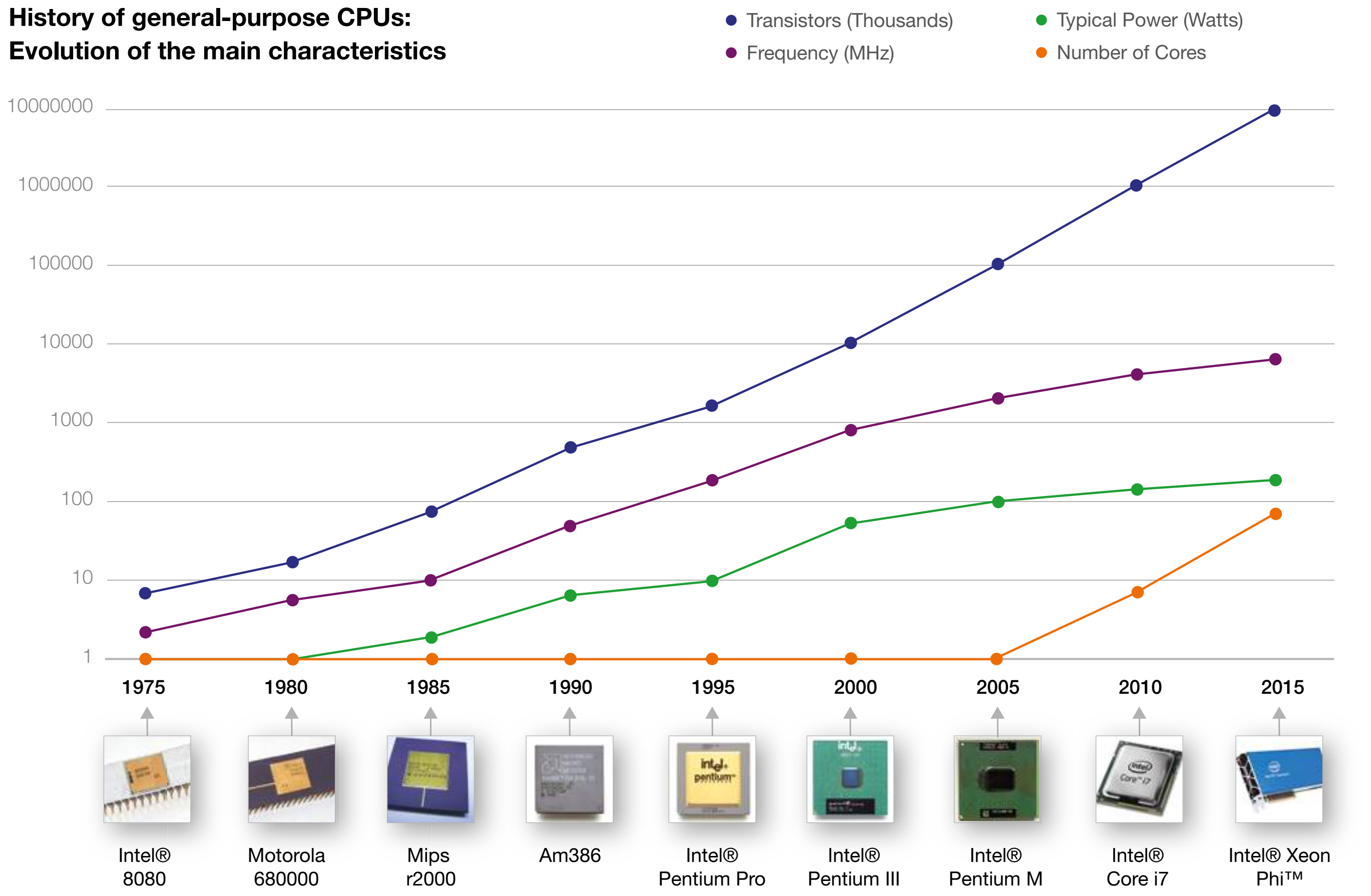

**Figure 7**
History of general-purpose CPUs: Evolution of the main characteristics

Modern machines combine powerful multicore CPUs with dedicated hardware designed to solve parallel processing. GPU and FPGA are the most popular dedicated hardware commonly available in workstations  developing AI systems.

A GPU (Graphics Processing Unit) is a chip designed to accelerate the processing of multidimensional data such as an image. A GPU consists of thousands of smaller cores, intended to work independently on a subspace of the input data, that needs heavy computation. Repetitive functions that can be applied to different parts of the input, such as texture mapping, image rotation, translation and filtering, are performed at a much faster rate and more efficiently, through the use of the GPU. A GPU has dedicated memory and the data must be moved in and out in order to be processed.

FPGA (Field Programmable Gate Array) is a reconfigurable digital logic containing an array of programmable logic blocks and a hierarchy of reconfigurable interconnections. An FPGA is not a processor and therefore it cannot run a program stored in the memory. An FPGA is configured using a hardware description language (HDL) and unlike the traditional CPU, it is truly parallel. This means, that each independent processing task is assigned to a dedicated section of the chip and many parts of the same program can be performed in simultaneously. A typical FPGA may also have dedicated memory blocks, digital clock manager, IO banks and several other features, which vary across different models. While a GPU is designed to perform efficiently, with similar threads on different subsets of the input, an FPGA is designed to parallelize sequential serial processing of the same program.

**1960 Central Processing Unit (CPU):** Integrated circuits that allow large number of transistors to be manufactured in a single chip.

**2001 Graphics Processing Unit (GPU):** Nvidia introduced the term GPU for integrated chips with image formation/ rendering engines.

**2010 Junction less transistor:** A control gate wrapped around a silicon nanowire that can control the passage of electrons without the use of junctions or doping.

**2014 TrueNorth –Neuromorhpic Adaptive Plastic Scalable Electronics:** The first neuromorphic integrated circuit to achieve one million individually programmable neurons with 256 individually programmable synapses.

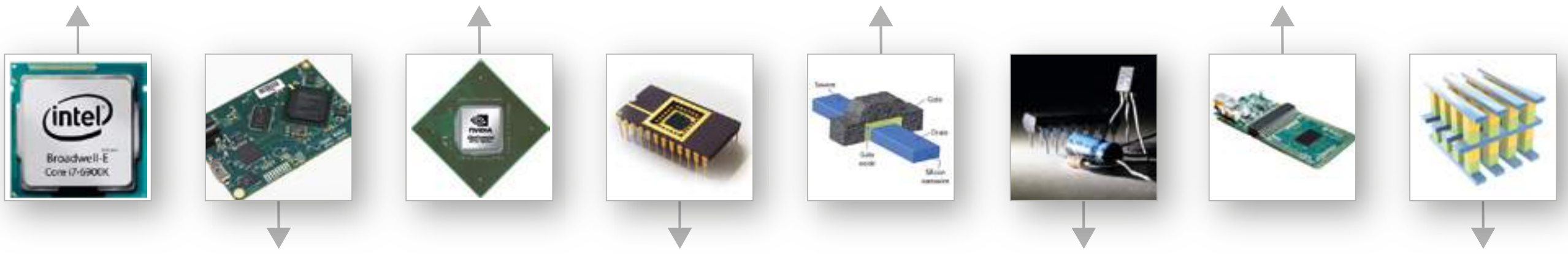

**1990: Field Programmable Gate Array (FPGA):** Integrated circuits designed to be configured after manufacturing.

**2008 Memristor:** A fourth basic passive circuit element. The memristor's properties permit the creation of smaller and better-performing electronic devices.

**2011 single-electron transistor:** 1.5 nanometres in diameter, made out of oxide based materials. Could spur the creation of microscopic computers.

**2015 3D Xpoint:** Non-volatile memory claimed to be significantly faster.

**Figure 8**
Timeline of the main hardware discoveries that have influenced the evolution of an AI system

# 9. ROBOTICS AND AI

Building on the advances made in mechatronics, electrical engineering and computing, robotics is developing increasingly sophisticated sensorimotor functions that give machines the ability to adapt to their ever-changing environment. Until now, the system of industrial production was organized around the machine; it is calibrated according to its environment and tolerated minimal variations. Today, it can be integrated more easily into an existing environment. The autonomy of a robot in an environment can be subdivided into perceiving, planning and execution (manipulating, navigating, collaborating). The main idea of converging AI and Robotics is to try to optimise its level of autonomy through learning. This level of intelligence can be measured as the capacity of predicting the future, either in planning a task, or in interacting (either by manipulating or navigating) with the world. Robots with intelligence have been attempted many times. Although creating a system exhibiting human-like intelligence remains elusive, robots that can perform specialized autonomous tasks, such as driving a vehicle [52], flying in natural and man-made environments [53], swimming [54], carrying boxes and material in different terrains [55], pick up objects [56] and put them down [57] do exist today.

Another important application of AI in robotics is for the task of perception. Robots can sense the environment by means of integrated sensors or computer vision. In the last decade, computer systems have improved the quality of both sensing and vision. Perception is not only important for planning but also for creating an artificial sense of self-awareness in the robot. This permits supporting interactions with the robot with other entities in the same environment. This discipline is known as social robotics. It covers two broad domains: human-robot interactions (HCI) and cognitive robotics. The vision of HCI it to improve the robotic perception of humans such as in understanding activities [58], emotions [59], non-verbal communications [60] and in being able to navigate an environment along with humans [61]. The field of cognitive robotics focuses on providing robots with the autonomous capacity of learning and acquiring knowledge from sophisticated levels of perception based on imitation and experience. It aims at mimicking the human cognitive system, which regulates the process of acquiring knowledge and understanding, through experience and sensorisation [62]. In cognitive robotics, there are also models that incorporate motivation and curiosity to improve the quality and speed of knowledge acquisition through learning [63, 64].

AI has continued beating all records and overcoming many challenges that were unthinkable less than a decade ago. The combination of these advances will continue to reshape our understanding about robotic intelligence in many new domains. Figure 9 provides a timeline of the milestone in robotics and AI.

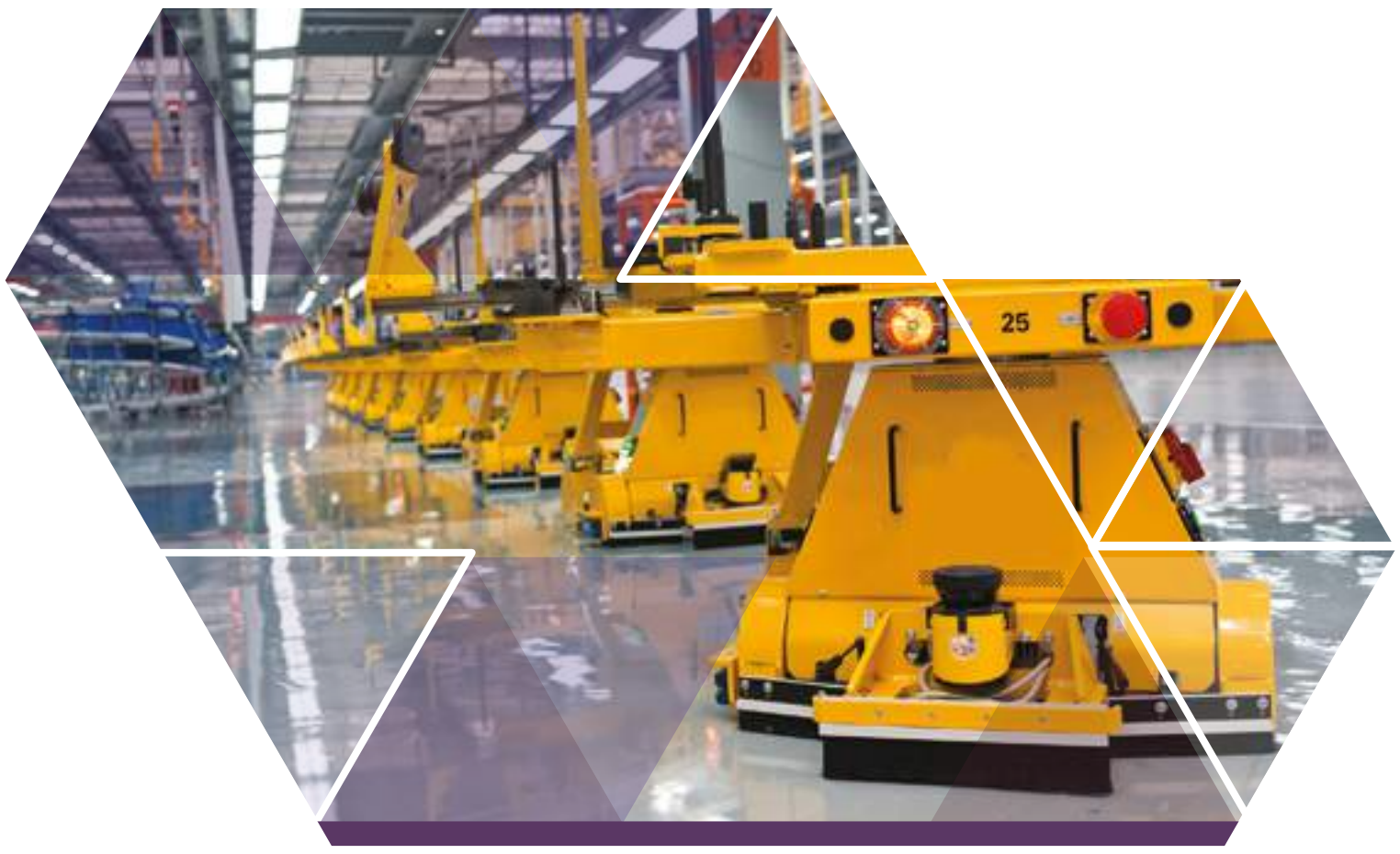

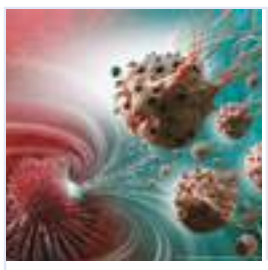

**2016 Nanorobots:** A team from Polytechnique of Montréal created a nanotransporter-bot that can administer drugs without damaging surrounding organs and tissues.

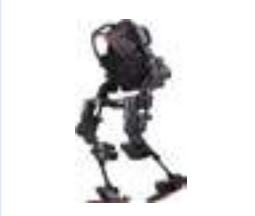

**2014 Robot exoskeleton:** A complete paralysed man was able to walk again using a robotic exoskeleton designed by Ekso Bionics.

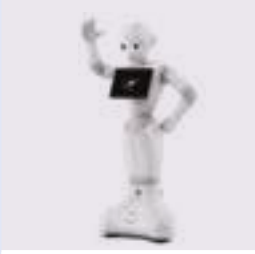

**2014 Pepper:** Japanese company Softbank presented the first robot, so-named Pepper, to be used for customer service. The robot has integrated an emotion engine to interact with people.

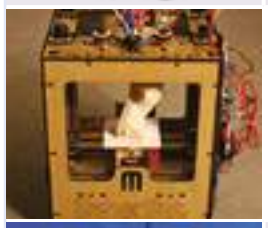

**2010 3D Printing:** First 3D printers were made commercially available.

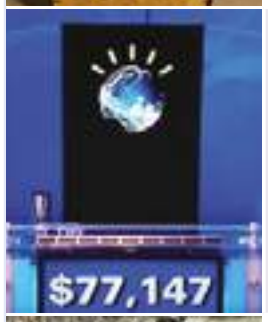

**2010 IBM Watson:** IBM's Watson computer beat human champions on the game show Jeopardy! by analysing natural language and finding answers to questions more rapidly and accurately than its human rivals.

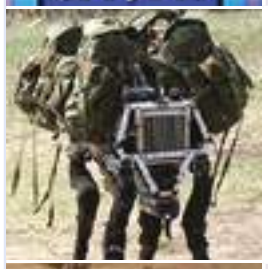

**2005 Robot BigDog:** Boston Dynamics created the first robot that could carry 150 Kg of equipment. The robot was able to traverse rough terrains using its four legs.

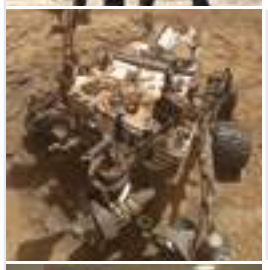

**2004 Mars Robot:** Robots landed on mars. Although they were only supposed to work for 90 days, they extended their lifetime for several years and remain operative until today.

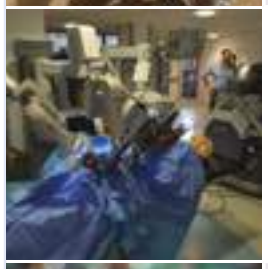

**2000 DaVinci Surgical System:** A surgical robot for minimally invasive (keyhole) surgery was approved by the FDA. The robot is controlled by a surgeon from a master console.

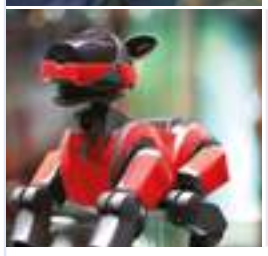

**1999 Aibo Robot:** First robotic pet dog. It could "learn", interact with its enviornment and responded more than 100 voice commands.

RoboCup

**1997 First Robocup competition:** An international competition for promoting AI and robotics where robots play a soccer tournament and other dexterity games.

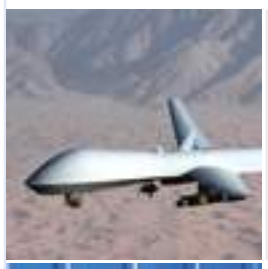

**1989 MQ-1 Predator drone:** The predator drone is introduced by the United State Air Force. It was equipped with reconnaissance cameras and could carry missiles.

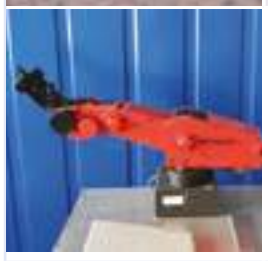

**1987 Mitsubishi Movemaster:** It was the first small robotic arm gripper which could perform tasks such as assembling small products or handling chemicals

AlphaGo

**2017 Go is solved:** A team from Google DeepMind created an algorithm named AlphaGo that beat top players of the ancient far-eastern board game Go.

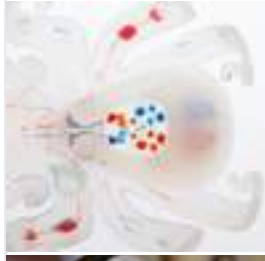

**2016 Microfluidic robot:** The first autonomous, entirely soft robot powered by a chemical reaction and a microfluidic logic was developed by a team at Harvard University.

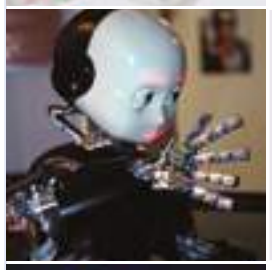

**2010 iCub:** A 1 meter high humanoid robot for research in human cognition at IIT, Italy. The robot can express emotions and is equipped with tactile sensors to interact with the environment.

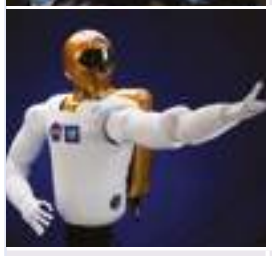

**2010 Robotnaut 2:** NASA revealed a humanoid robot with a wide range of sensors that can replace human astronauts.

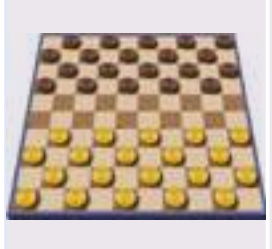

**2007 Checkers is solved:** A program from University of Alberta named Chinook was able to solve the problem of checkers and beat humans at several competitions.

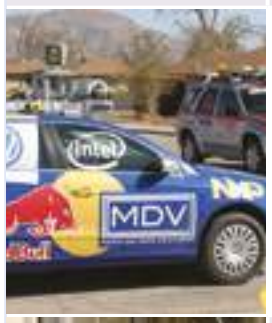

**2005 Autonomous vehicle challenge:** A team from Stanford University won the challenge organized by DARPA for driving autonomously off-road across a 175-mile long desert terrain without human intervention.

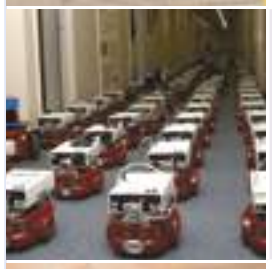

**2002 Darpa's Centibots:** First collaborative robot swarm of mobile robots that could survey an area and build a map in real time without human supervision.

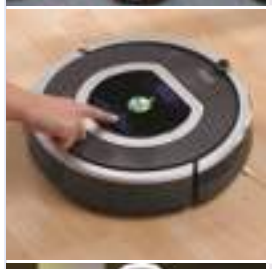

**2002 Roomba:** The first household robot for cleaning. It was able to detect and avoid obstacles as well as navigating within a house without using maps.

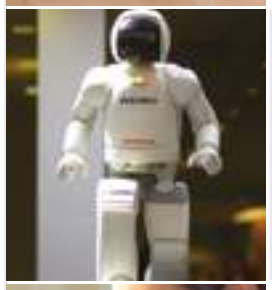

**2000 Asimo:** Robot Asimo from Honda presented the first humanoid robot that could walk like humans, climb stairs, change its direction and detect hazards using a video camera.

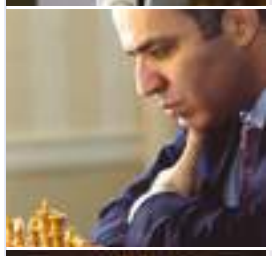

**1997 Deep blue beats Garry Kasparov:** After a rematch in 2016, deep blue defeated Garry Kasparov by 2 to 1.

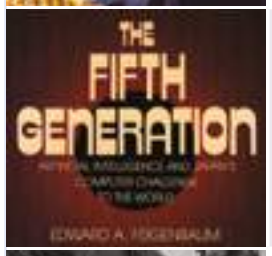

**1992 End of next AI project:** End of Japan's multimillion program for developing the fifth generation computer systems based on AI.

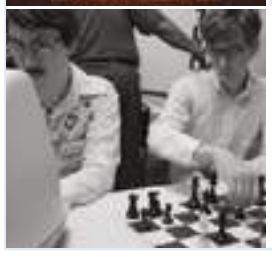

**1989 Computer beats Human at chess:** Computers beat humans at chess for the first time.

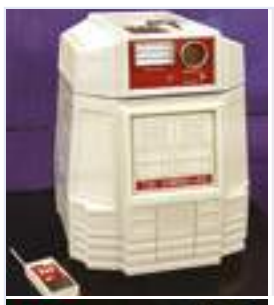

**1984 Hero Robot Kit:** First playful robot with human interaction abilities. It could navigate a room guided by sonar and tried to remain near humans by listening to their voices.

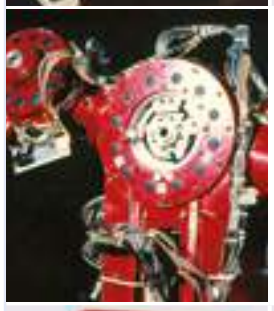

**1981 Direct Drive arm:** The technology of the direct drive arm was first presented by Takeo Kanade. This helped minimise a lot of the friction and backlash of earlier robots that used chains and tendons.

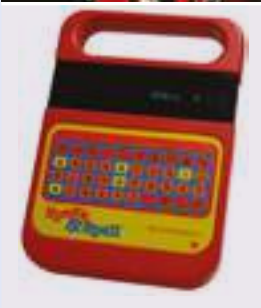

**1978 Speak and Spell:** The first commercially available human artificial speech system, by Texas Instruments Inc. The system was able to perform a mathematical model of the human track and elicit human sounds.

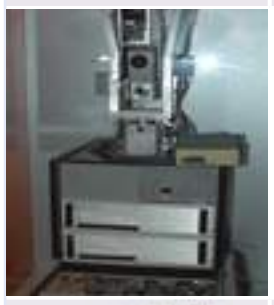

**1970 Shakey the robot:** The first mobile robot able to navigate within an environment using AI. It was equipped with a TV camera, laser range finders and bump sensors. It could also communicate wirelessly over a radio link.

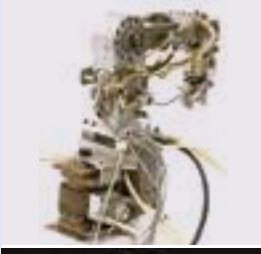

**1969 Stanford Robotic arm:** The first electrically powered computer-controlled robot arm. The robot used a set of optical and contact sensors.

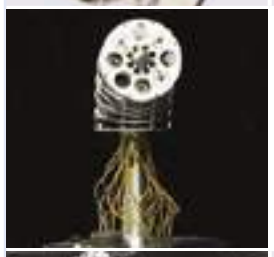

**1963 The Orm robot:** A team at Stanford university developed the first soft-robot inspired by snake.

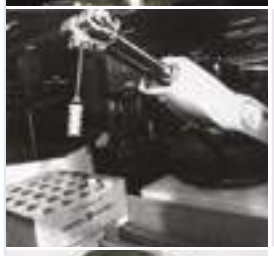

**1961 Unimate:** The first robot for industrial mass-production started working at General Motors.

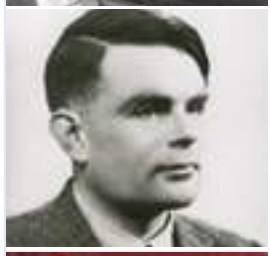

**1956 Alan Turing's Test:** Alan Turing proposed a test to be able to identify whether the machine is ingelligent.

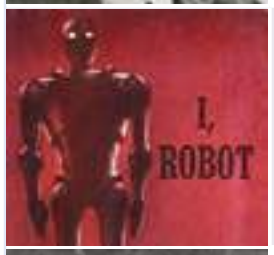

**1950 Isaac Assimov published I, Robot:** The first novel literature book dedicated to Robots.

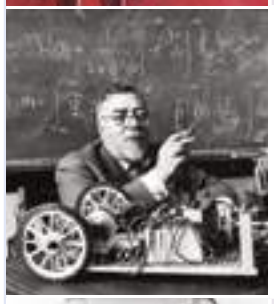

**1948 Norbert Wiener Cybernetics:** The term Cybernetics was coined for the first time and a potential convergence of artificial intelligence (first proposed by Walter Pitts as logical calculus) and control systems.

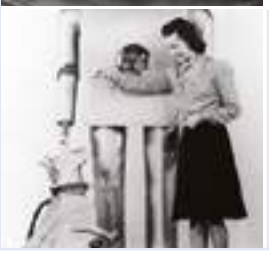

**1939 Elektro and Sparko:** The first robot (automaton) was presented at the world's fair.

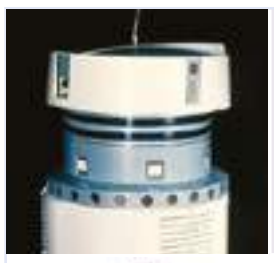

**1985 Denning Sentry Robot:** The first security sentry robot was developed, which could patrol for more than 14 hours and alert about anything unusual in a range of 150-foot radius.

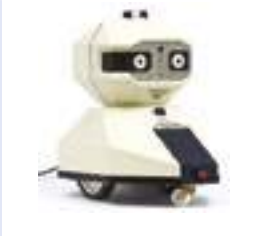

**1982 FRED robot:** The first playful robot was developed by Atari engineering. It never made it to the market.

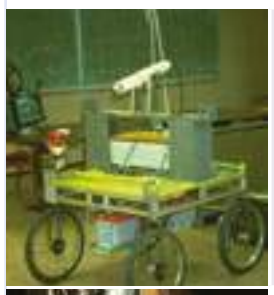

**1979 Stanford Cart:** A most advanced mobile robot that was able to move through a room avoiding obstacles. It was able to take pictures about the environment from different angles.

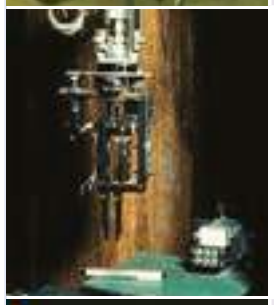

**1974 The silver arm:** MIT designed the first robot that is able to assemble small parts. The robot was equipped with pressure and touch sensors to mimic the soft touch of human fingers.

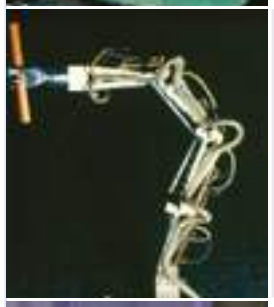

**1970 The Tentacle Arm:** Marvin Minsky designed a tentacle arm robot inspired by an octopus. It had more than twelve tentacles that allowed the robot to go around obstacles.

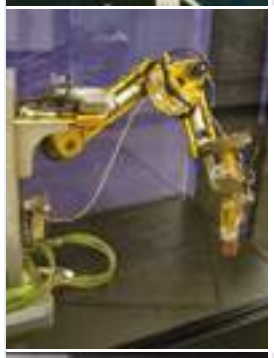

**1963 The Rancho Arm:** The first assistive robot was developed at a hospital in California and later controlled by a computer at Stanford University. The robot was composed of six joints used as a tool for the handicapped. This is the first robotic arm controlled by a computer.

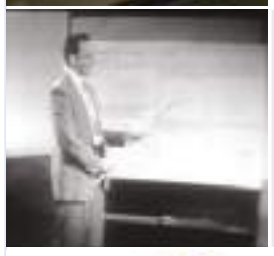

**1959 Automatically Programmed Tools:** The first computer assisted manufacturing (CAM) system developed at MIT.

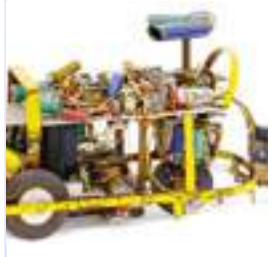

**1951 Robot Squee:** A robot inspired by a squirrel with the capacity of sensing the environment using light sensors.

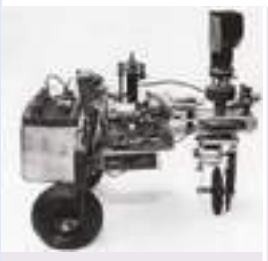

**1950 Robot Elsie:** Grey Walter developed the first mobile robot (automaton) with a goal-seeking behaviour.

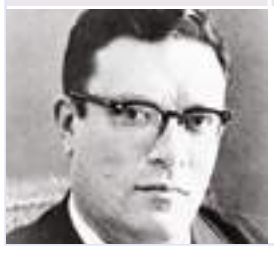

**1941:** Isaac Asimov's article published in Astounding Science Fiction: For the first time the term "robot" was used in an article and the three laws of robotics were presented.

**Figure 9**
A timeline of robotics and AI.

# 10. PROGRAMMING LANGUAGES FOR AI

| Logo | Language | Date | Type | Infuenced by | AI resources |
|---|---|---|---|---|---|
| | **Lisp** | 1958 | Multi-paradigm (functional, procedural) | IPL | • Homoiconic: easy to deal with large amount of data.<br>• Good mathematical alignment.<br>• Lots of resources for symbolic AI (Eurisko or CYC) |
| | **C++** | 1983 | Procedural | C, Algol 68 | • Fast execution times.<br>• Some compatible libraries for AI such as Alchemy for Markov logic and Mlpack for general ML |
| | **C#** | 2000 | Multi-paradigm (functional, procedural) | C++, Java, Haskell | • Easy prototyping and well elaborated environment.<br>• Most used language for AI in games as provides good compatibility with popular games engines such as Unity. |
| | **Clojure** | 2007 | Functional | Lisp, Erlang, Prolog | • Easy design and cloud infrastructure that works on top of the JVM.<br>• Rapid interactive development and libraries for development of behaviour trees (alter-ego) |
| | **Erlang** | 1986 | Functional Concurrent | Lisp, Prolog | • Good framework to deal with concurrency and elastic clouds (scalability).<br>• Libraries for logic programming such as erlog. |
| | **Go** | 2009 | Procedural Concurrent | Algo, CSP, Python | • Easy concurrency and asynchronous patterns with a decent runtime.<br>• A few libraries for machine learning such as Golearn. |
| | **Haskell** | 1990 | Functional | Lisp | • Easy parallelization and possibility of handling infinite computations.<br>• A few utilities to implement neural networks (LambdaNet) and general ML (HLearn). |
| | **Java** | 1995 | Procedural Concurrent | C++, Ada 83 | • VM provides efficient maintainability, portability and transparency.<br>• A myriad for libraries and tools for AI such as Tweety and ML (DeepLearning4, Weka, Mallet etc.) |
| | **Julia** | 2012 | Multi-paradigm | Lisp, Lua, Matlab, Python | • Easy integration with C and Fortran. Scientific oriented language.<br>• Several ML packages such as Mocha, or MLBase. |

| Logo | Language | Date | Type | Infuenced by | AI resources |
|---|---|---|---|---|---|
| | **Lua** | | (procedural, functional) | C++, Scheme | • Versatile and lightweight language.<br>• It is the de-facto language used for the machine and deep learning framework Torch. |
| | **Matlab** | 1993 | Multi-paradigm | APL | • Solid Integrated environment. Matrix, linear algebra oriented language.<br>• A selection of toolboxes and utilities for machine learning, statistics and signal processing. |
| SWI Prolog | **Prolog** | 1984 | (procedural, functional) | Planner | • Good set of utilities for expressing the relationships between objects and symbolic computation.<br>• Large set of internal functionalities to perform logic programming. |
| | **Python** | 1972 | Procedural | C++, java, haskell, perl | • Highly useful standard library that makes the language versatile and flexible. Focus on rapid development.<br>• Plethora of frameworks and utilities for AI, ML, deep learning, scientific computing, natural processing language etc. |
| | **R** | 1972 | Declarative | Lisp, Scheme | • Most comprehensive sets of statistical analysis functions and packages.<br>• Rich community of tools for AI or ML provided freely through the CRAN repository. |
| | **Scala** | 1993 | Multiparadigm (procedural, functional) | Erlan, Haskel, Java, Lisp, Lisp (Scheme) | • Fast run time (almost as C). It runs on top of the JVM. Very good support for distributed systems.<br>• Several libraries and frameworks for AI, ML and numerical computing (ScalaNLP). |

Programming languages played a major role in the evolution of AI since the late 1950s and several teams carried out important research projects in AI; e.g. automatic demonstration programs and game programs (Chess, Ladies) [65]. During these periods researchers found that one of the special requirements for AI is the ability to easily manipulate symbols and lists of symbols rather than processing numbers or strings of characters. Since the languages of the time did not offer such facilities, a researcher from MIT, John MacCarthy, developed, during 1956-58, the definition of an ad-hoc language for logic programming, called LISP (LISt Processing language). Since then, several hundred derivative languages, so-called "Lisp dialects", have emerged (Scheme, Common Lisp, Clojure); Indeed, writing a LISP interpreter is not a hard task for a Lisp programmer (it involves only a few thousand instructions) compared to the development of a compiler for a classical language (which requires several tens of thousands of instructions). Because of its expressiveness and flexibility, LISP was very successful in the artificial intelligence community until the 1990s.

Another important event at the beginning of AI was the creation of a language with the main purpose of expressing logic rules and axioms. Around 1972 a new language was created by Alain Colmerauer and Philippe Roussel named

Prolog (PROgramming in Logic). Their goal was to create a programming language where the expected logical rules of a solution can be defined and the compiler automatically transforms it into a sequence of instructions. Prolog is used in AI and in natural language processing. Its rules of syntax and its semantics are simple and considered accessible to non-programmers. One of the objectives was to provide a tool for linguistics that was compatible with computer science.

In the 1990s, the machine languages with C / C ++ and Fortran gained popularity and eclipsed the use of LISP and Prolog. Greater emphasis was placed on creating functions and libraries for scientific computation on these platforms and were used for intensive data analysis tasks or artificial intelligence with early robots. In the middle of the 1990s, the company Sun Microsystems, started a project to create a language that solved secutiry flaws, distributed programming and multi-threading of C++. In addition, they wanted a platform that could be ported to any type of device or platform. In 1995, they presented Java, which took the concept of object orientation much further than C++. Equally, one of the most important additions to Java was the Java VM (JVM) which enabled the capability of running the same code in any device regardless of their internal technology and without the need of pre-compiling for every platform. This added new advantages to the field of AI that were be introduced in devices such as cloud servers and embedded computers. Another important feature of Java was that it also offered one of the first frameworks, with specific tools for the internet, bringing the possibility of running applications in the form of java applets and javascripts (i.e. self-executing programs) without the need of installation. This had an enormous impact in the field of AI and a set the foundation in the fields of web 2.0/3.0 and the internet of things (IoT).

However, the development of AI using purely procedural languages was costly, time-consuming and error prone. Consequently, this turned the attention into other multi-paradigm languages that could combine features from functional and procedural object-oriented languages. Python, although first published in 1991, started to gain popularity as an alternative to C/C++ with Python 2.2 by 2001. The Python concept was to have a language that could be as powerful as C/C++ but also expressive and pragmatic for executing "scripts" like Shell Script. It was in 2008, with the publication of Python 3.0, which solved several initial flaws, when the language started to be considered a serious contender for C++, java and other scripting languages such as Perl.

Since 2008, the Python community has been trying to catch up with specific languages for scientific computing, such as Matlab and R. Due to its versatility, Python is now used frequently for research in AI. However, although python has some of the advantages of functional programming, run-time speeds are still far behind other functional languages, such as Lisp or Haskell, and even more so from C/C++. In addition, it lacks of efficiency when managing large amounts of memory and highly-concurrent systems.

From 2010 and mostly driven by the necessity of translating AI into commercial products, (that could be used by thousands and millions of users in real time), IT corporations looked for alternatives by creating hybrid languages, that combined the best from all paradigms without compromising speed, capacity and concurrency. In recent years, new languages such as Scala and Go, as well as Erlang or Clojure, have been used for applications with very high concurrency and parallelization, mostly on the server side. Well-known examples are Facebook with Erlang or Google with Go. New languages for scientific computation have also emerged such as Julia and Lua.

Although functional programming has been popular in academia, its use in industrial settings has been marginal and mainly during the times when “expert systems” were at their peak, predominantly during the 1980s. After the fall of expert systems, functional programing has, for many years, been considered a failing relic from that period. However, as multiprocessors and parallel computing are becoming more available, functional programming is proving to be a choice of many programmers to maximise functionality from their multicore processors. These highly expensive computations are usually needed for heavy mathematical operations or pattern matching, which constitute a fundamental part of running an AI system. In the future, we will see new languages that bring simplifications on existing functional languages such as Haskell and Erlang and make this programming paradigm more accessible. In addition, the advent of the internet-of-things (IoT) has drawn the attention to the programming of embedded systems. Thus, efficiency, safety and performance are again matters for discussion. New languages that can replace C/C++ incorporating tips from functional programming (e.g. Elixir) will become increasingly popular. Also, new languages that incorporate simplifications as well as a set of functions from modern imperative programming, while maintaining a performance like C/C++ (e.g. Rust), will be another future development.

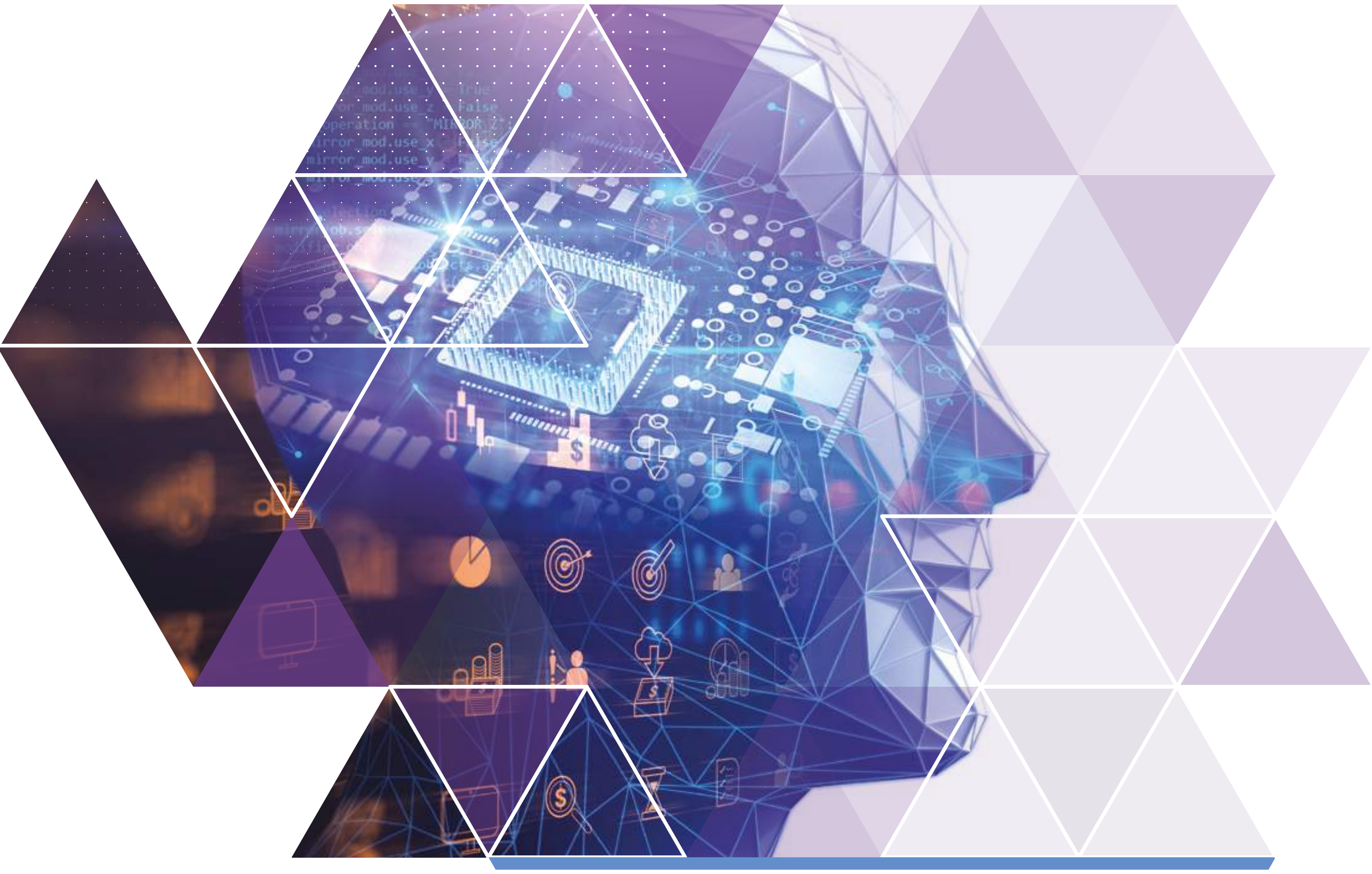

"Programming languages played a major role in the evolution of AI. Driven by the necessity of translating AI into commercial products, hybrid languages are emerging, which combine the best from all paradigms without compromising speed, capacity and concurrency."

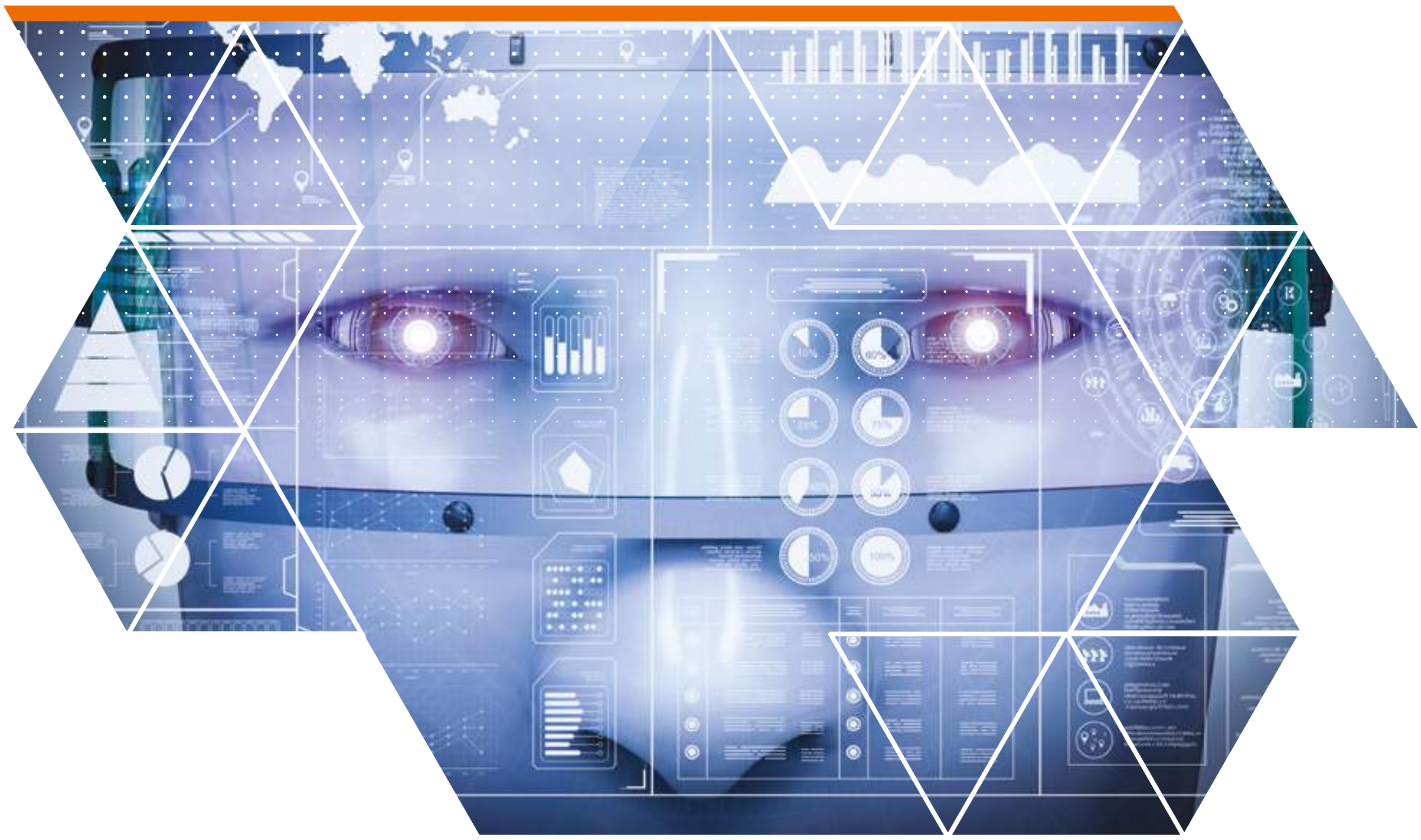

"Machine vision integrates image capture and analysis with machine learning to provide automatic inspection, scene recognition and robot navigation. Scene reconstruction along with object detection and recognition are the main sub-domains of machine vision."

# 11. IMPACT OF MACHINE VISION

Machine vision integrates image capture systems with computer vision algorithms to provide automatic inspection and robot guidance. Although it is inspired by the human vision system, based on the extraction of conceptual information from two-dimensional images, machine vision systems are not restricted to 2D visible light. Optical sensors include single beam lasers to 3D high definition Light Detection And Ranging (LiDAR) systems, also known as laser scanning 2D or 3D sonar sensors and one or multiple 2D camera systems. Nevertheless, most machine vision applications are based on 2D image-based capture systems and computer vision algorithms that mimic aspects of human visual perception. Humans perceive the surrounding world in 3D and their ability to navigate and accomplish certain tasks depends on reconstructing 3D information from 2D images that allows them to locate themselves in relation to the surrounding objects. Subsequently, this information is combined with prior knowledge in order to detect and identify objects around them and understand how they interact. Scene reconstruction along with object detection and recognition are the main sub-domains of computer vision.

Regardless of the imaging sensors used, the most common approaches of reconstructing 3D information are normally based on either time-of-flight techniques, multi-view geometry and/or on photometric stereo. The former is used in laser scanners to estimate the distance between the light source and the object based on the time required for the light to reach the object and return back. Time-of-flight approaches are used to measure distances in kilometres and they are accurate to a millimetre scale, since they are limited by the ability to measure time. On the other hand, multi-view geometry problems include 'structure' problems, 'stereo correspondence' problems and 'motion' problems. Recovery of the 3D 'structure' implies that given 2D projections of the same 3D point, in two or more images, the 3D coordinates of the point are estimated based on triangulation. 'Stereo correspondence' refers to the problem of finding the image point that corresponds to a point from another 2D view. Finally, 'motion' refers to the problem of recovering the camera coordinates given a set of corresponding points in two or more image views. 3D laser scanners based on triangulation can reach micrometre accuracy but their range is constrained to a few meters. Several sub-problems such as 'structure from motion' uses multi-view geometry principles to extract corresponding points between 2D views of the same object and reconstruct its shape.

Stereo-vision assumes the robust extraction of corresponding salient points/features across images, the so-called interest point detection. These features should be invariant to photometric transformation such as changes in the lighting conditions and covariant to geometric transformations. For over two decades researchers have proposed several approaches. The Scale-invariant feature transform (SIFT) extracts features that are invariant to scale, rotation and translation transformations and robust to illumination variations and moderate perspective transformations. Since its introduction in 1999-2004, it has been successful in several vision applications, including object recognition, robot localisation and mapping.

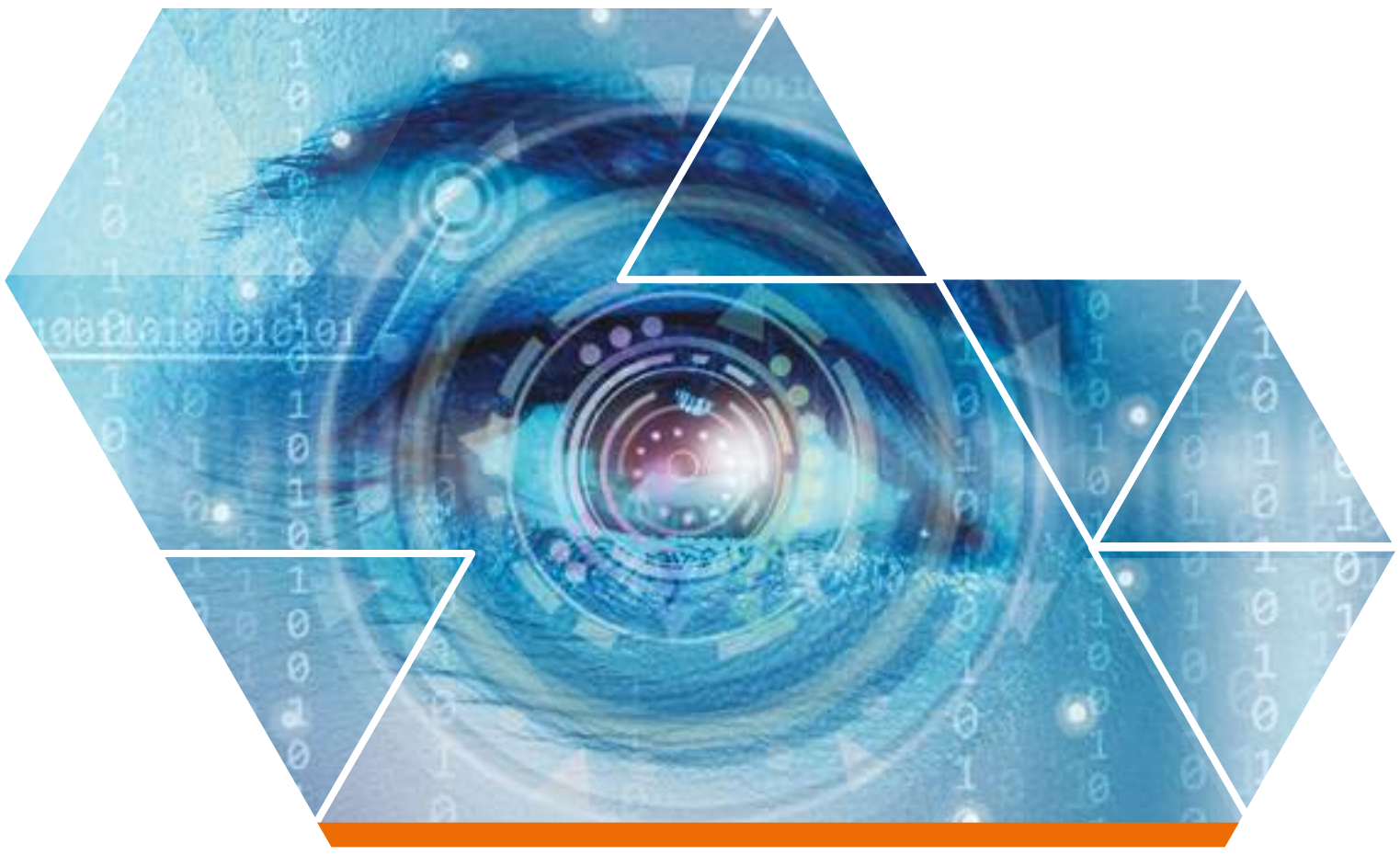

Representing and recognising object categories have proven much harder problems to generalise and solve, compared with 3D reconstruction, since there are thousands of objects that can belong to an arbitrary number of categories simultaneously. Several ideas about object detection are related to Gestalt psychology, which is a theory of mind with relation to visual perception. A major aspect of the theory is about grouping entities together based on their proximity, similarity, symmetry, common fate, continuity and so on. From the 1960s to early 1990s, research in object recognition was centred on geometric shapes. This was a bottom-up process, which uses a small number of primitive 3D dimensional objects that are assembled together in various configurations to form complex objects. In the 1990s, appearance-based models were explored, which were based on manifold learning of the object appearance parameterised by the pose and illumination [66]. These techniques are not robust to occlusion, clutter and deformation. By the mid-late 1990s, sliding window approaches were designed that classify whether an object is found for each instance of a sliding window across an image [67]. The main challenges were how to design features that represent appropriately the appearance of the object and how to efficiently search a large number of positions and scales. Local features approaches were also developed and they aimed towards those which were invariant to image scaling, geometric transformations and illumination changes [68]. In the early 2000s 'parts-and-shape' models along with 'bags of features' were suggested. Parts-and-shape models represent complex objects using combinations of multi-scaled deformable objects [69]. On the other hand, bags of features methods, represent visual features as words and relate object recognition and image classification to the expressive power of natural language processing approaches [70].

Machine learning in object recognition facilitated a shift, from solving a problem based on mathematical modelling alone, to learning algorithms based on real-data and statistical modelling. A major breakthrough in object recognition and classification came in 2012 with the emergence of deep neural networks and the availability of large labelled image databases, such as ImageNet. Compared to classical object recognition methods, which depend on feature extraction followed by feature matching methodologies, deep learning has the advantage of encoding both feature extraction and image classification via the structure of a neural network. The superb performance of deep neural networks resulted in an increase of image classification from 72% in 2010 to 96% in 2015, which outperforms human accuracy and has had a significant impact in real-life applications [71]. Both Google and Baidu updated their image search capabilities based on the Hinton's deep neural network architecture. Face detection has been introduced in several mobile devices and Apple even created an app to recognise pets. The accuracy of these models in object recognition and image classification exceed human-level accuracy and spread waves of technological changes across the industry.

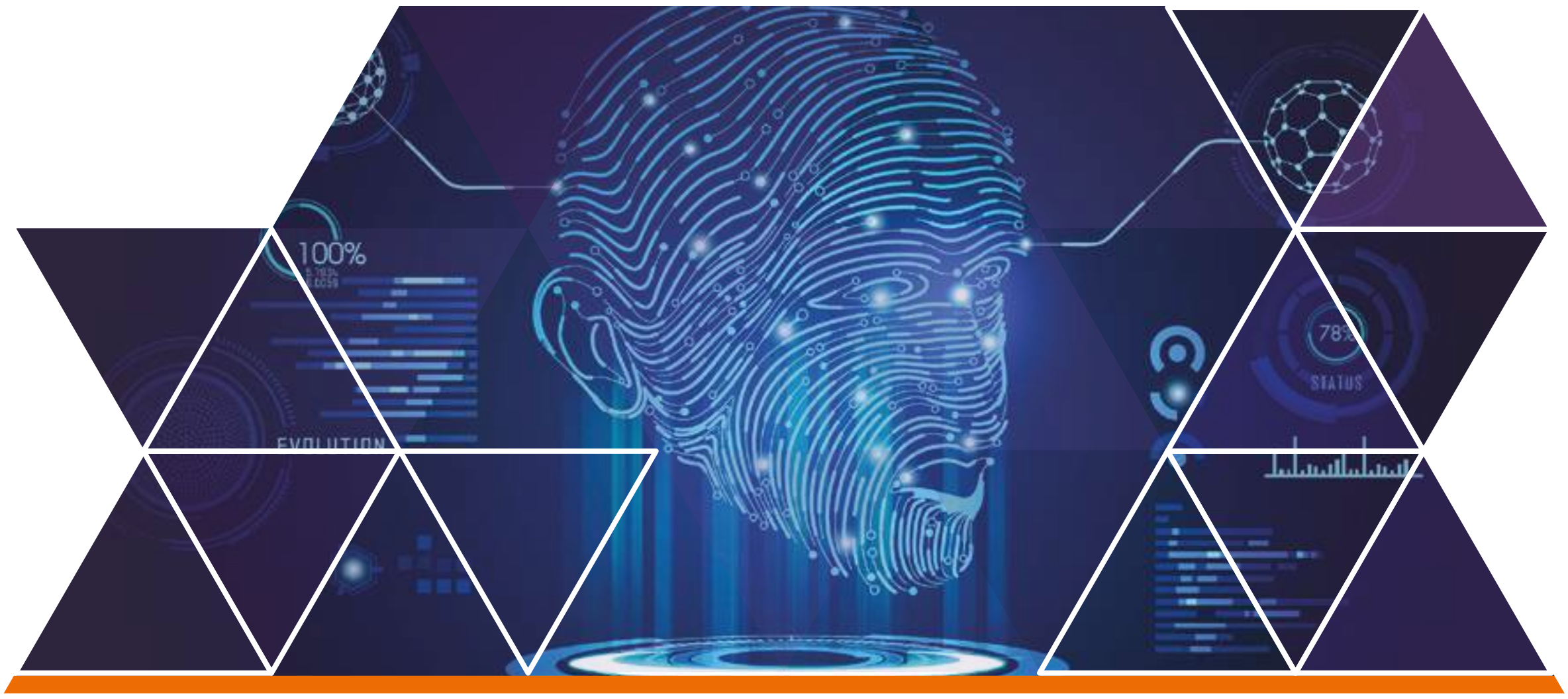

## IMAGING DEVICES/HARDWARE

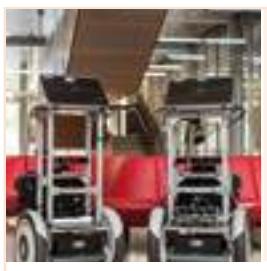

**2017 BWIBots:** Vision robots learn the human's preferences and how to cooperate by working side by side with humans (Khandelwal et al. 2017).

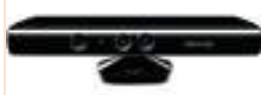

**2009 Kinect:** Microsoft announced a device that used structured-light computational stereo technology to track body's posture. Within 60 days, it sold 8 million units and claimed the Guinness World Record of the 'fastest selling consumer electronics device'.

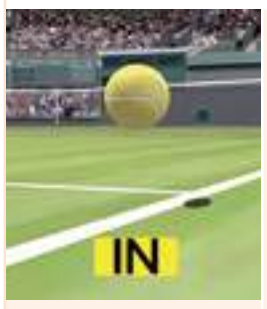

**2001 Hawk-Eye:** A real-time vision system with multiple high-performance cameras for providing a 3D representation of the trajectory of a ball using triangulation.

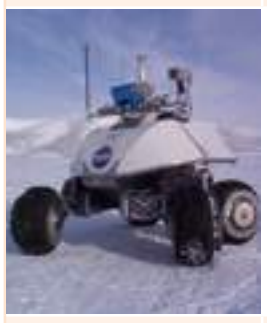

**1997 Nomad robot:** Autonomous used to search Antarctic meteorites based on advanced perception and navigation technologies developed at Carnegie Mellon University.

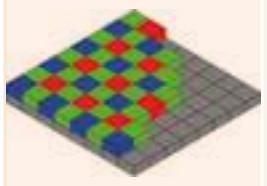

**1974 Bayer filter camera:** Bryce Bayer, an American scientist working for Kodak, captured vivid colour information onto a digital image.

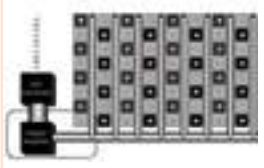

**1969 Charge-Couple Device (CCD):** CCD was invented at American Bell Laboratories by William Boyle and George E. Smith. CCD is the major technology for digital imaging as it converts incoming photons into electron charges.

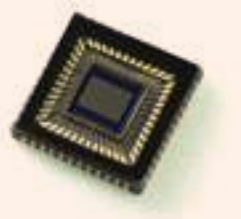

**1963 Complementary Metal-Oxide-Semiconductor (CMOS):** Frank Wanlass, an American electrical engineer patented CMOS used in digital logic circuits as well as analogue circuits and image sensors.

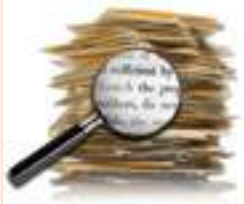

**1914 Optical Character Reader (OCR):** Goldberg invented a machine that could read characters and convert them into standard telegraph code.

## COMPUTER VISION/CONCEPTS

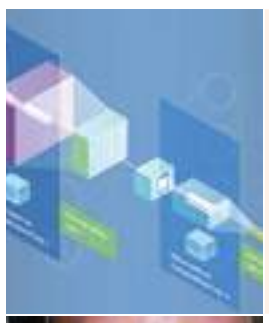

**2012-present Deep Neural Networks in Image Classification:** DNNs are trained with big image datasets such as ImageNet and currently have exceeded human abilities in object/face recognition (Krizhevsky et al. 2012).

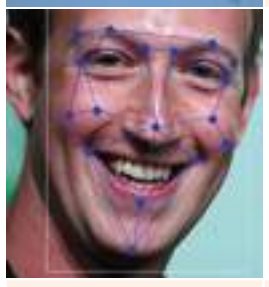

**2004 Real-time face detection:** Machine learning approach of sliding-window based object recognition for robust face detection has been introduced (Viola et al. 2004).

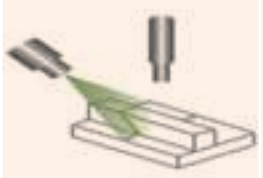

**2002 Active stereo with structured light:** Zhang et al. introduced the idea of using light patterns to estimate robust correspondence between a pair of images.

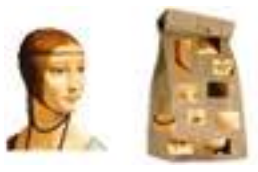

**2001 Bag of words in computer vision:** Representing visual features as words to allow natural language processing information retrieval to apply in object recognition and image classification (Leung et al. 2001).

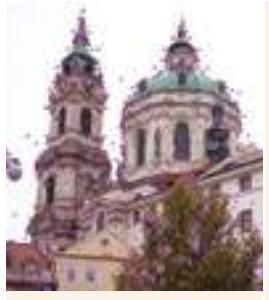

**1999 Scale-Invariant Feature Transform (SIFT):** David Lowe patented an algorithm to detect and describe local features in images. SIFT features are invariant to uniform scaling, orientation and illumination changes.

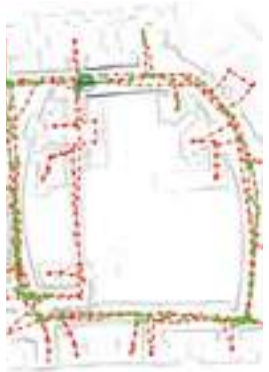

**Early 1990s Simultaneous Localisation and Mapping (SLAM):** Leonard and Durrant-whyte pioneered a probabilistic method for handling uncertainty of noisy sensor readings and allows autonomous vehicles to localise themselves (Leonard et al. 1991).

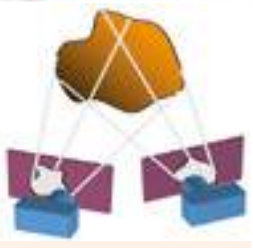

**1981 Computational Stereo:** Grimson presented the theory of computational stereo vision that is biologically plausible.

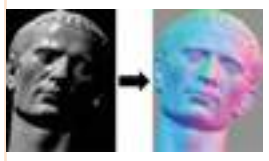

**1980 Photometric Stereo:** Woodham presented a method to extract surface normals from multiple images based on smoothness constrained posed by the illumination model.

Khandelwal et al. The International Journal of Robotics Research, 2017

Krizhevsky et al. Advances in neural networks, 2012

Leonard et al. Intelligent Robots and Systems, 1991

Leung et al. International Journal of Computer Vision 2001

Viola et al. Interantional Journal of Computer Vision 2004

**Figure 10**
A timeline of Computer Vision and AI.

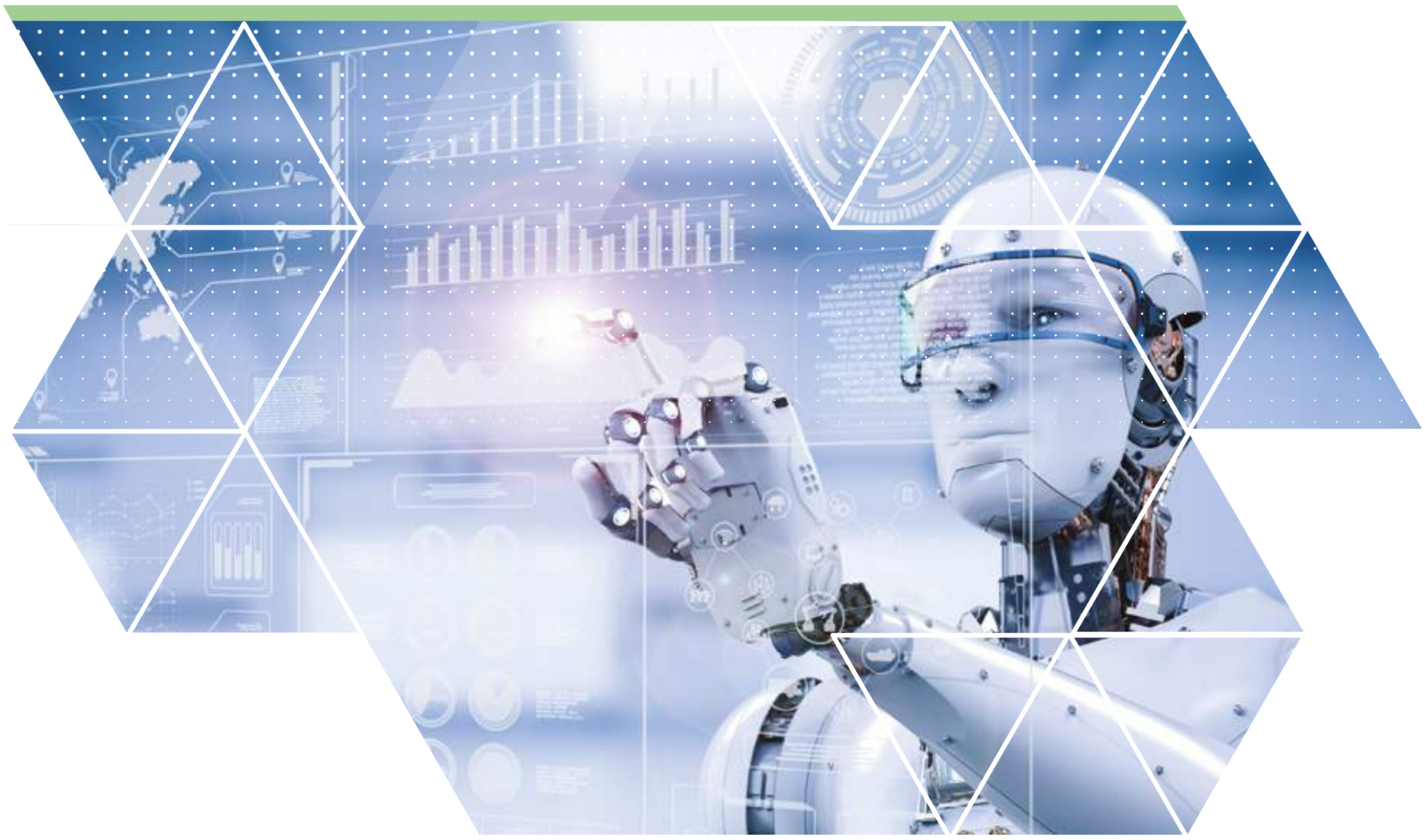

"

The development of AI is closely coupled with our pursuit of understanding the human brain. A long-term goal of computational neuroscience is to emulate the brain by mimicking the causal dynamics of its internal functions so that the models relate to the brain function.

"

# 12. ARTIFICIAL INTELLIGENCE AND THE BIG BRAIN

The development of AI is closely coupled with our pursuit of understanding the human brain. A long-term goal of computational neuroscience is to emulate the brain by mimicking the causal dynamics of its internal functions so that the models relate to the brain functions in human/animal behaviour characteristics [72]. Although there is a lot of excitement that this approach would help understand how intelligent behaviour emerges, there is also heated debate on whether this is a realistic goal as we do not fully understand how the brain works [73, 74]. The human brain is highly complex with more than 80 billion neurons and trillion of connections. Simulation scales can range from molecular and genetic expressions to compartment models of subcellular volumes and individual neurons to local networks and system models.

Deep Neural Network nodes are an oversimplification of how brain synapses work. Signal transmission in the brain is dominated by chemical synapses, which release chemical substances and neurotransmitters to convert electrical signals via voltage-gated ion channels at the presynaptic cleft into post-synaptic activity. The type of neurotransmitter characterises whether a synapse facilitates signal transmission (excitatory role) or prevents it (inhibitory role). Currently, there are tenths of known neurotransmitters, whereas new ones continuously emerge with varying functional roles. Furthermore, dynamic synaptic adaptations, which affect the strength of a synapse, occur in response to the frequency and magnitude of the presynaptic signal and reflect complex learning/memory functions, (Spike time dependent plasticity). Recently, evidence has found that surrounding cells, such as glia cells that are primarily involved in 'feeding' the neurons, can also affect their function via the release of neurotransmitters.

The earliest and most elaborate attempts for modelling neuronal complexity are based on conductance-based biophysical models of synaptic interaction (COBA) and spike generation, such as the Hodgkin-Huxley (HH) model. Several popular software platforms, such as GENESIS and NEURON, exploit COBA models to develop systematic modelling approaches of realistic brain networks. Simpler models, such as integrate-and-fire, are faster to simulate and they are also used in large-scale simulations when the dynamics of spike generation are less important.

It is evident that the computational complexity of the brain networks depends not only on the number of neurons and synapses but also on the topology of the network and the level of biological details it describes. The mammalian cerebral cortex is organised intuitively to minimise wiring. Group of neurons form cortical columns, which encode functionally similar features and they are considered the smallest functional brain units from where consciousness emerges. In fact the major objective of the Blue Brain project, a leading research project on computational neuroscience, started in 2005, was to emulate the rat neocortical columns of around 10,000 neurons and 108 synapses.

Currently, there is a trend toward developing neuromorphic hardware circuits to allow near real-time large-scale simulations of neural networks [75]. The neuromorphic term refers to mimicking the structure of neurons (dendrites, axon and synapse) to achieve functional equivalence. For example, the Neurogrid is a neuromorphic supercomputer, developed at Stanford, which combines analogue dendritic computation (presynaptic ion-channel function) with digital axonal communication. Each chip integrates one million neurons interconnected with 256 million synapses. Analog implementations are more directly related to synaptic activity and thus maximise energy efficiency and allow emulation of a greater number of neurons and synapses. On the other hand, digital implementations are more flexible to reconfigure and allow storage of connectional weights with greater precision. One of the most promising technologies in this area is the memristors, which has the ability to simultaneously perform logical and storage operations. Similar to neural synapses, the more current that flows through a memristor, the more it can flow. Therefore, it can model several biological functions including COBA synaptic interactions time varying delayed networks and spike-timing-dependent plasticity. Memristors have the potential to replace the existing digital/analog neuromorphic chips with much more efficient and versatile technology.

With the current exponential growth in the efficiency and flexibility of neuromorphic hardware, the ability to make a computer as fast and as complex as the human brain is becoming increasingly possible. Several large simulations of the brain are aimed to mainly replicate the cortical dynamics obtained in in-vivo and in-vitro key neuroscience studies and validate cognitive theories and hypothesis. The big question is if this would result in a conscious, intelligent creature and how we would be able to detect this in the very beginning without a thorough understanding of what consciousness or intelligence is.

## MODELS/SOFTWARE PLATFORMS

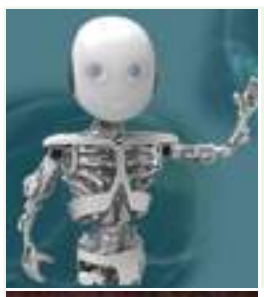

**2015-present NeuroRobotics, Human Brain Project:** Robotic systems interfaced with brain models for closed loop cognitive experiments in simulated environment.

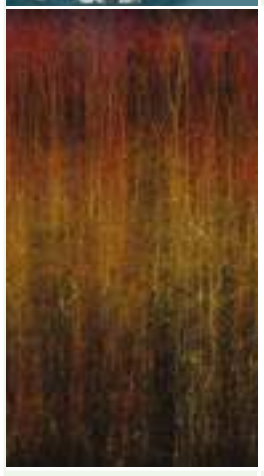

**2015 Somatosensory cortex of the juvenile rat - Blue Brain Project & Human Brain Project:** First digital reconstruction of the microcircutiry of the somatosensory cortex of juvenile rat. The study aimed to study detailed cortical dynamics and reproduced in vitro and in vivo experimental results (Markram, Muller et al. 2015).

**2013 Human Brain Project:** It is an EU-funded research project that aims to advance knowledge in the fields of neuroinformatics, brain simulation, neuromorphic computing and neurorobotics.

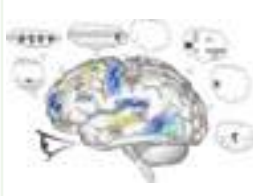

**2012-present Spaun – University of Waterloo, Canada:** First computer model to produce complex behaviour and achieve human performance in simple tasks. It ran on Nengo platform and modelled brain function in a biologically realistic scale (Eliasmith, Stewart et al. 2012).

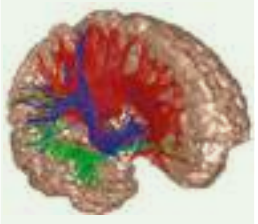

**2005-8 Thalamo-cortical Model – The Neurosciences institute, San Diego, USA:** Simulation of thalamo-cortical dynamics with of a neural network that represents 300-by-300 $mm^2$ mammalian thalamo-cortical surface.
One second of simulation took 50 days to run (Izhikevich and Edelman 2008).

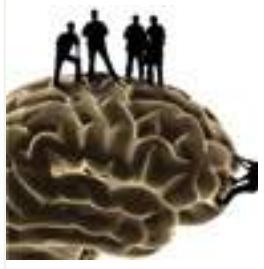

**2005-present Nengo – University of Waterloo, Canada:** Open source software that provides tools to specify the collective function of large groups of neurons along with low-level electrophysiological details.

## HARDWARE

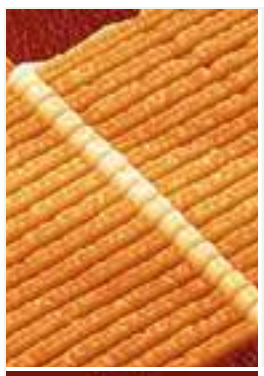

**2017 STDP via ferroelectric solid-state synapses, (memristors):** Thin ferroelectric layer within two electrodes of which one can adjust the resistance by means of the voltage pulses to emulate spike-timing-dependent plasticity (Boyn, Grollier et al. 2017).

BRAINS IN SILICON
STANFORD UNIVERSITY

**2014-present Neurogrid – Stanford:** A neuromorphic supercomputer combines analogue dendritic computation with digital axonal communication. Each chip integrates 1 million spiking neurons and 256 million synapses.

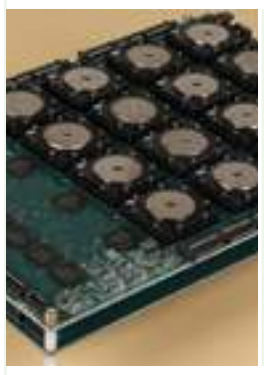

**2014 TrueNorth – IBM & SyNAPSE (Systems on Neuromorhpic Adaptive Plastic Scalable Electronics):** The first neuromorphic integrated circuit to achieve one million individually programmable neurons with 256 individually programmable synapses.

The BRAIN Initiative

**2013 The BRAIN Initiative (Brain Research through Advancing Innovative Neurotechnologies):** A US initiative that aimed to develop technologies to understand brain function.

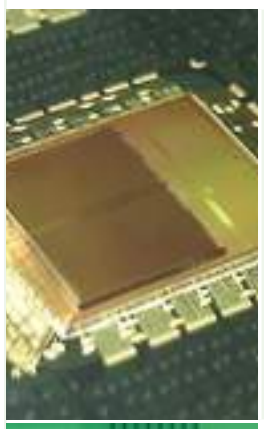

**2009-present SpiNNaker– University of Manchester:** Massive parallel computer architecture composed by billion computing elements communicating using stochastic spike events. It is based on a neuromorphic architecture of six layers thalamocortical model and 'unreliable'/stochastic communication.

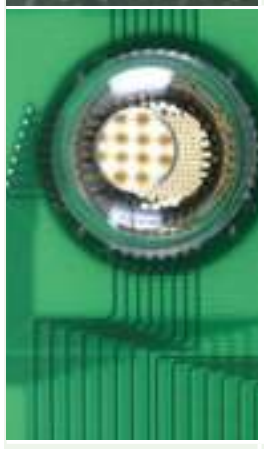

**2006 Silicon Retina – Stanford:** The first artificial retina constructed in silico to reproduce the responses of the four major ganglion cell types that drive visual cortex, producing 3600 spiking outputs. It paved the way of a neural prosthesis that matches the dimensions of biological retina (Zaghloul and Boahen 2006).

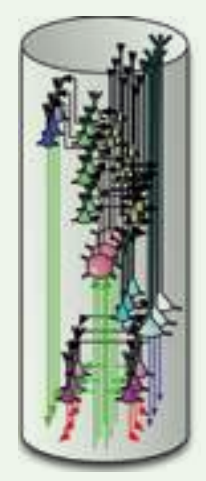

**2006 Cortical columns modelling - IBM & Blue Brain Project:** Run on an IBM Blue Gene/P architecture to model cortical minicolumns, consisting of 22 million 6-compartment neurons, with 11 billion synapses, with spatial delays corresponding to a 16 $cm^2$ cortex surface and a simulation length of one second real-time. (Lundqvist, Rehn et al. 2006).

**2005 The Blue Brain Project:** A Swiss brain initiative that aimed to realistically simulate the mammalian brain.

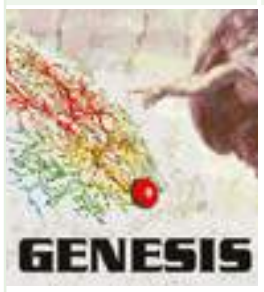

**1990-present NEURON – Yale & Duke**
**1985-present GENESIS (General NEural SImulation System) – Caltech** Common open-source software simulation platforms that provides COBA models of synaptic interaction and plasticity.

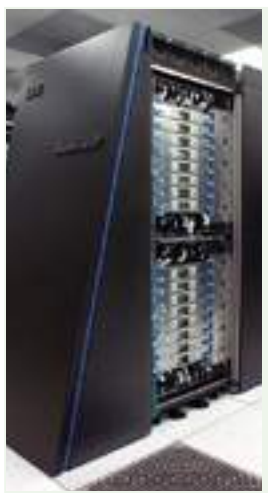

**2005-2015 IBM Blue Gene:** A series of supercomputers that aimed to reach the performance of one petaflop. In 2009, the project was awarded the National Medal on Technology and Innovation. The latest model, Blue Gene/Q, consists of 18 core chips with a 64-bit A2 processor cores and it has peak performance of 20 Petaflops.

Boyn et al. 2017 Nature Communications

Eliasmith et al. 2012 Science

Izhikevich et al. 2008 PNAS

Lundqvist et al. 2006 Network Computation in Neural Systems

Markram et al. 2015 Cell

Niebur et al. 1993 Mathematical Biosciences

Zaghloul et al. 2006 Journal of Neural Engineering

**Figure 11**
Timeline of AI and the Big Brain.

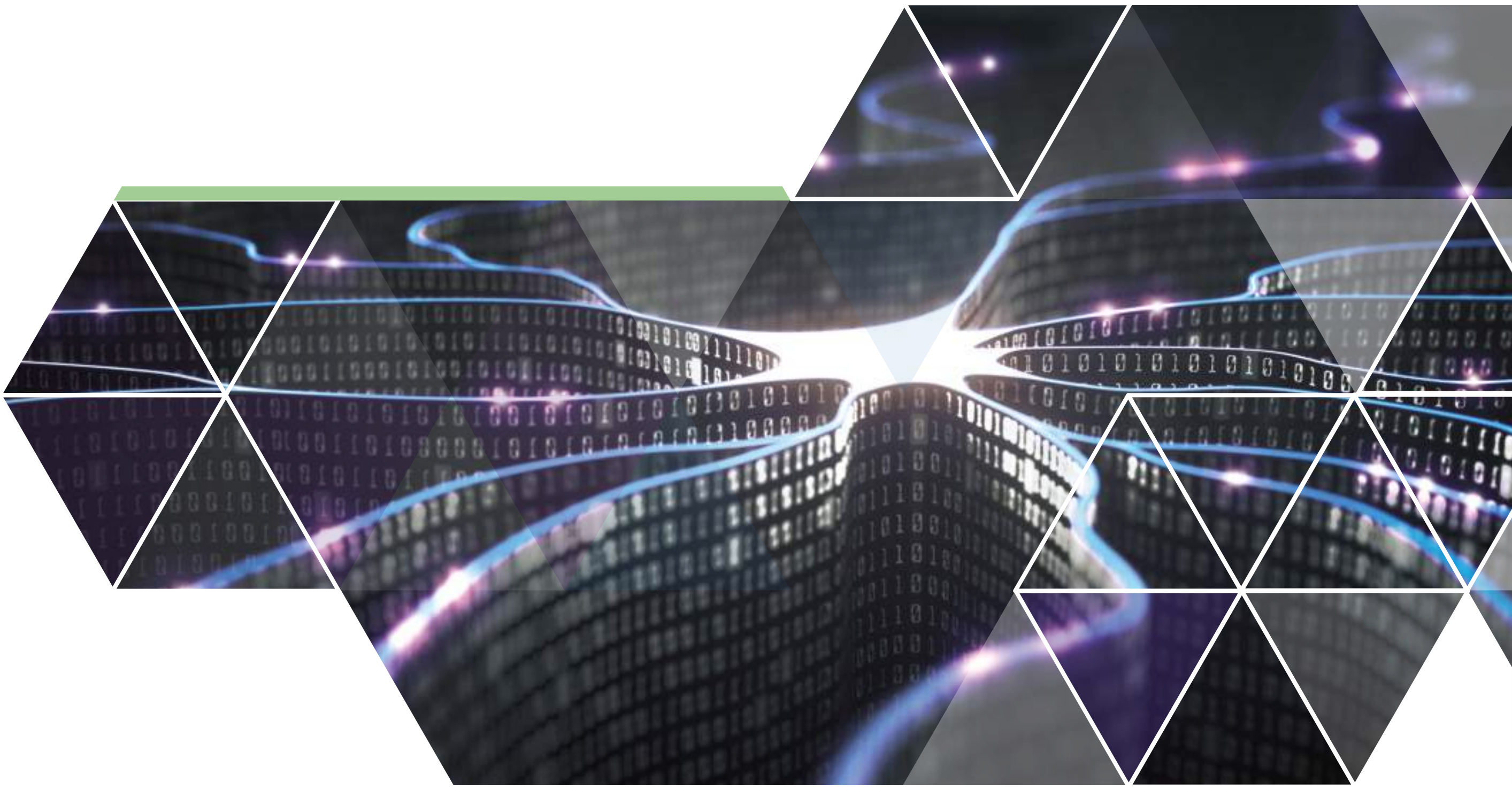

# 13. ETHICAL AND LEGAL QUESTIONS OF AI

## 13.1 ETHICAL ISSUES IN ARTIFICIAL INTELLIGENCE

**Threat to Privacy**
Data is the "fuel" of AI and special attention needs to be paid to the information source and if privacy is breached. Protective and preventive technologies need to be developed against such threats. Although solutions to this issue may be unconnected to the AI operation per se, it is the responsibility of AI operators to make sure that data privacy is protected. Additionally, applications of AI, which may compromise the rights to privacy, should be treated with special legislation that protects the individual.

**Threats to security and weaponisation of AI**

With the proliferation of security risks, such as terrorism and regional conflicts, we are immersed in a global arms race that has developed a demand for new weapons powered by AI, such as autonomous drones and missiles, as well as virtual bots and malicious software aimed at sophisticated espionage. This could lead to the possibility of an unprecedented war escalation. The danger of AI is that we could potentially lose control over it. This has led associations and non-governmental organizations (NGOs) to carry out awareness-raising actions to contain the use of these military robots and potentially ban the use of such weapons.

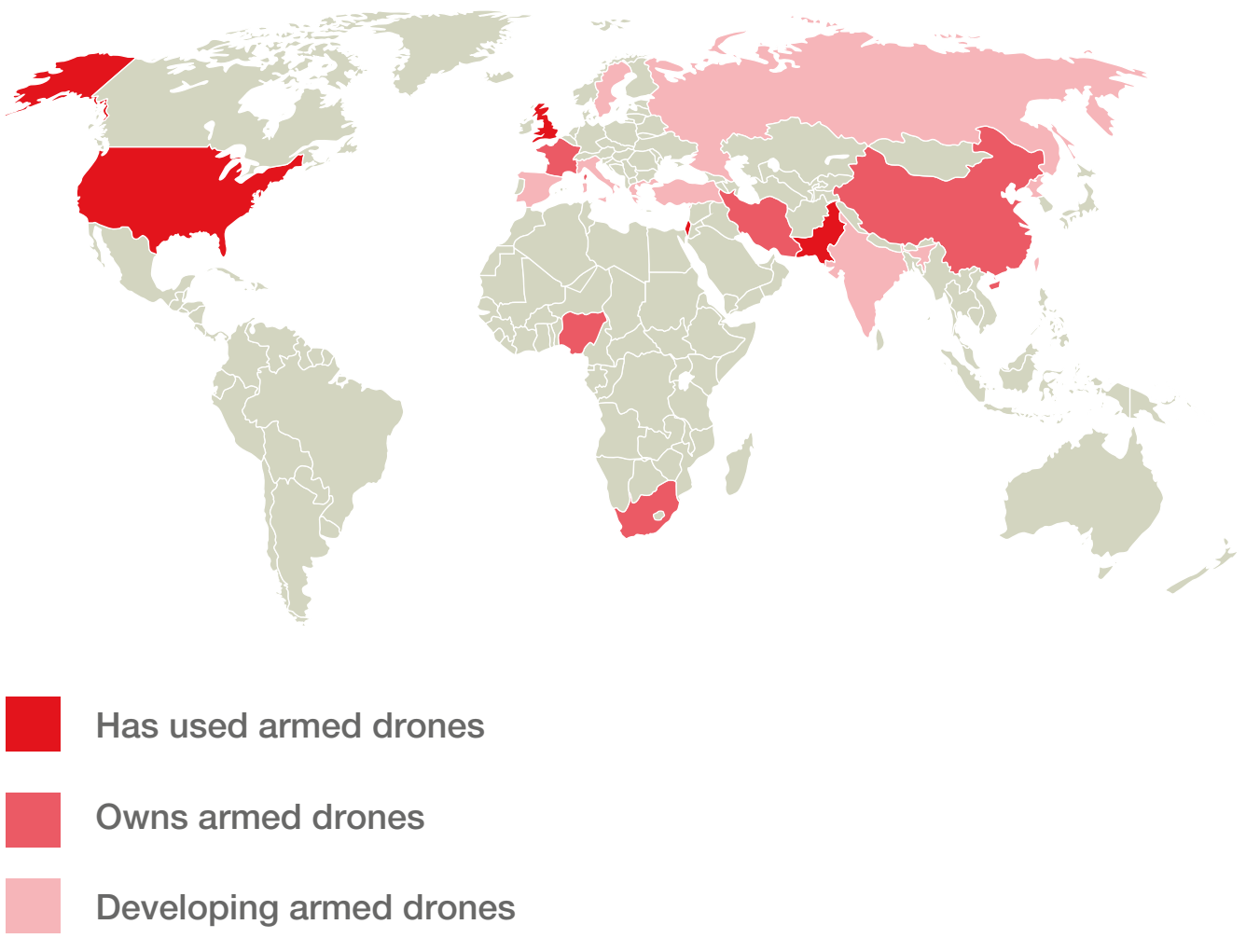

**Figure 12**
Countries using, owning or developing armed drones

The US military has just released a prospective document entitled 'Robotic and Autonomous Systems Strategy', which details its strategy for robotics and autonomous systems. The use of robotics and autonomous systems by the US military defines the following five objectives: 1) Increase knowledge abilities in operations' theatres, 2) Reduce the amount of charge carried by the soldier; 3) Improve logistics capacity; 3) Facilitate movement and manoeuvring; 4) Increase the protection of forces. Although offensive capacity is not mentioned, the four points introduced by the US military are ambiguous with respect to their scope and limitations.

**Economics and Employment Issues**
Currently, 8% of jobs are occupied by robots, but in 2020 this percentage will rise to 26%. Robots will become increasingly autonomous and be able to interact, execute and make more complex decisions. Thanks to 'big data', robots now have a formidable database that allows them to experiment and learn which algorithms work best.
The accelerated process of technological development now allows labor to be replaced by capital (machinery). However, there is a negative correlation between the probability of automation of a profession and its average annual salary, suggesting a possible increase in short-term inequality [76]. The problem is not the number of jobs lost through automation, but in creating enough to compensate for potential job losses. In previous industrial revolutions, new industries hired more people than those who lost their jobs in companies that closed, because they could not compete with the speed of development in new technologies [77]. An important note, about this revolution, is the fact that it is not only the manual trades likely to be automated, but also in jobs involving tasks of an intermediate nature, such secretarial, administration and other office work. To deal with this situation, it is necessary to put in place legal frameworks that make sure the benefits of automation do not solely go to the employer but are distributed equally, to guarantee the maintenance of education, health and pensions.

**Human Bias in Artificial Intelligence**
In an article published by Science magazine, researchers saw how machine learning technology reproduces human bias, for better or for worse. Words related to the lexical domain of flowers are associated with terms related to happiness and pleasure (freedom, love, peace, joy, paradise, etc.). The words relating to insects are, conversely, close to negative terms (death, hatred, ugly, illness, pain, etc.).

This reflects the links that humans have made themselves. AI biases have already been highlighted in other applications. One of the most notable was probably Tay, a Microsoft AI launched in 2016, which was supposed to embody a teenager on Twitter, able to chat with internet users and improve through conversations. However, in just a few hours, the program, learning from its exchanges with humans, began to make racist and anti-Semitic remarks, before being suspended by Microsoft (see Sidebar – Failures of AI). The problem is not only at language level. When an AI program became a juror in a beauty contest in September 2016, it eliminated most black candidates as the data on which it had been trained to identify "beauty" did not contain enough black skinned people.

## 13.2 LEGAL ISSUES AND QUESTIONS OF AI

**Legal Responsibility**

Initially, the legal framework that would apply to robots and AI would have the purpose of limiting the risks derived from the operation of these systems and limit the damage that could occur from unintended consequences with their operation. AI could not have constitutional rights, as these are the property of individuals, but can have some property rights, in order to guarantee their possible liability for any damage caused. For this reason they could have some form of judicial protection. In this case, robots and AI would be subject to two types of responsibility: 1) Predictability in their actions; and 2) civil liability in any harmful consequences of their actions (although there is also a fiscal responsibility as a consequence of non-compliance with the obligations of this type).

The European Parliament is already working on a Recommendations Report (2015/2103, dated 31 May 2016) on civil law on robotics, which provides guidelines for regulating civil liability arising from the use of robots [78]. It refers to the contractual and non-contractual responsibilities that may derive from its action and recommends that this liability be defined as an objective as well as establishing the need to have compulsory insurance for civil liability, for any damages arising from the possession and use of such robots. That means that in the event of an accident, the report proposes a compulsory insurance scheme, a policy identical to that of automobiles. The manufacturers will have a contractual obligation to compensate possible victims and to consolidate a fund to protect against robot accidents. The report also considers the effect of "short-circuits", to protect humans from any accidents or aggression. This does not mean that the responsibility of man will be entirely negated but rather a sliding scale could be created: the more sophisticated a robot is, the more responsibility the designer would bear.

In view of the recent government policy, autonomous machines (cars without drivers, drones, medical devices) will soon have civil liability. Autonomous machines must have a name, a first name and a registration number. In the event of an accident, identification will be aided with some kind of civil status.

**Civil Rights for AI and Robots**

Robots, to the extent that they are autonomous, could be granted the status of electronic persons with specific rights and obligations. There are also calls to harmonize the cohabitation between robots and humans. For instance, domestic robots are sort of "intimate gadgets". They create the notion of empathy with the humans they interact with on a daily basis. A legal framework would make it possible to crystallize this particular type of relationship in law. In comparison with domestic animals, whose legal status was clarified in January 2015, can be drawn here. There is however a distinction in that, unlike animals, robots are not biologically alive and have no sensitivity. However, they are still endowed with an intelligence that can be superior to one of an animal. The idea of providing AI systems and robots a set of rights, such as ones provided to domestic animals, requires an understanding of how machines could process their own feelings and emotions, if machines are equipped in future with some sort of emotional intelligence.

From the point of view of the labour market, the use of robots will mean the disappearance of certain jobs that, until now, have been performed by humans. To reduce the social impact of unemployment caused by robots and autonomous systems, the EU parliament proposed that they should pay social security contributions and taxes as if they were human. By producing a surplus value from their work, they generate an economic benefit. It is one of the more controversial points in the proposals of the EU Robotics Report.

# 14. LIMITATIONS AND OPPORTUNITIES OF AI

AI has the potential to change the world but there are still many problems to overcome before its widespread applications. Furthermore, its practical use is not without failures (see Fig 13, Example Failures of AI). Recently, we have seen a surge of interest in deep learning with promising results that will reshape the future of AI. But deep learning is only one of the many tools that the AI community has developed over the years. It is important to put into perspective the current development of AI and its specific limitations.

Intelligence as a multi-component model: A machine to be called “intelligent” should satisfy several criteria that include the ability of reasoning, building models, understanding the real word and anticipate what might happen next. The concept of “intelligence” is made of the following high-level components: perception, common sense, planning, analogy, language and reasoning.

Large datasets and hard generalisation: After extensive training on big datasets, today machines can achieve impressive results in recognising images or translating speech. These abilities are obtained thanks to the derivation of statistical approximations on the available data. However, when the system has to deal with new situations when limited training data is available, the model often fails. We know that humans can perform recognition even with small data since we can abstract principles and rules to generalise to a diverse range of situations. The current AI systems are still missing this level of abstraction and generalisability.

Black box and a lack of interpretation: Another issue with the current AI system is the lack of interpretation. For example, deep neural networks have millions of parameters and to understand why the network provides good or bad results becomes impossible. Despite some recent work on visualising high-level features by using the weight filters in a convolution neural network, the obtained trained models are often not interpretable. Consequently, most researchers use current AI approaches as a black box.

Robustness of AI: Most current AI systems can be easily fooled, which is a problem that affects almost all machine learning techniques.

Despite these issues, it is certain that AI will play a major role in our future life. As the availability of information around us grows, humans will rely more and more on AI systems to live, to work and to entertain. Therefore, it is not surprising that large tech firms are investing heavily on AI related technologies. In many application areas, AI systems are needed to handle data with increasing complexities. Given increased accuracy and sophistication of AI systems, they will be used in more and more sectors including finance, pharmaceuticals, energy, manufacturing, education, transport and public services. In some of these areas they can replace costly human labour and create new potential applications and work along with/for humans to achieve better service standards. It has been predicted that the next stage of AI is the era of augmented intelligence. Ubiquitous sensing systems and wearable technologies are driving towards intelligent embedded systems that will form a natural extension of human beings and our physical abilities. Human sensing, information retrieval and physical abilities are limited in a way that AI systems are not. AI algorithms along with advanced sensing systems could monitor the world around us and understand our intention, thus facilitating seamless interaction with each other.

Advances in AI will also play a critical role in imitating the human brain function. Advances in sensing and computation hardware will allow to link brain function with human behaviour at a level that AI self-awareness and emotions could be simulated and observed in a more pragmatic way. Recently, quantum computing has also attracted a new wave of interest from both academic institutions and technological firms such as Google, IBM and Microsoft. Although the field is at its infancy and there are major barriers to overcome, the computational power it promises, potentially relevant to the field of AI, is well beyond our imagination.

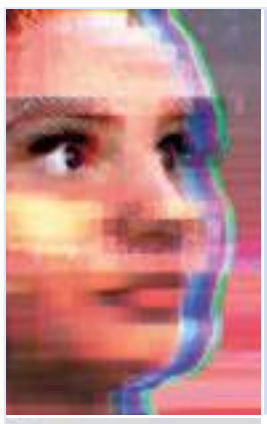

**2016:** Artificial intelligence fails to beat real stupidity. Microsoft lunched Tay, an AI Twitter bot that just after 24 hours turned into a racist, Hitler-lover, incestual sex-promoting robot. Tay was developed to learn by interacting with people, but it failed to cater for the darker side of human.

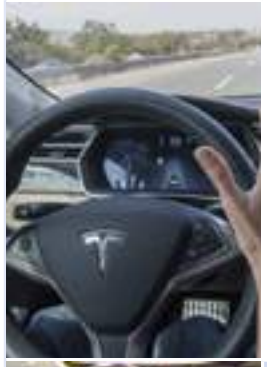

**2016:** Tesla driver dies while using autopilot mode. A driver died in the first fatal crash involving a self-driving car. The car's vision system failed to distinguish a large truck and trailer crossing the highway at full speed.

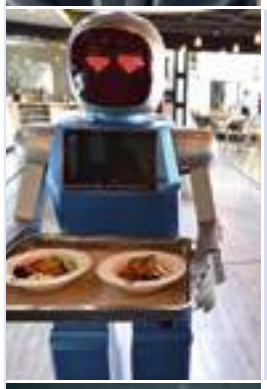

**2016:** Robots failed in their duties and lost their jobs. Three restaurants in China fired their robot waiters. The robots broke down continuously, two of the restaurants have closed and the third re-hired humans.

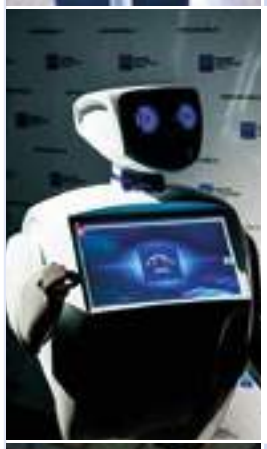

**2016:** Service robot caused traffic jams. Promobot, a humanoid robot, was designed for promoting products or conduct surveys. It has created many traffic jams after escaping from its lab and it was also arrested while it was collecting voter opinions for political candidates to gain unfair advantage.

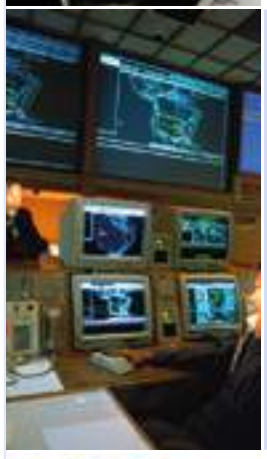

**1983:** Nuclear attack early warning system falsely claimed that an attack is taking place. A nuclear warning system of the Soviet Union reported the launch of a ballistic missile from the United States. This was a false alarm that it could have cause a nuclear attack and an immediate escalation of the cold-war to a full-scale nuclear war.

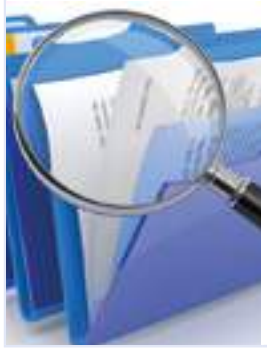

**1966** Mistranslations DARPA funded a project to translate Soviet documents into English. It was a clear failure and after spending $20 million, it was closed.

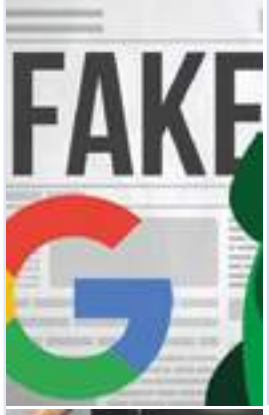

**2016:** Search engine AI highly ranks fake news. The highly ranked entry of the Google search engine on the query "Did the Holocaust happen" was a neo-Nazi and denial website. Since that, Google has tried to improve their AI search engine to provide authoritative results.

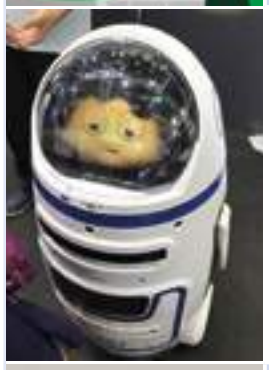

**2016:** Fatty the robot smashes glass and injures visitor. During a demonstration in China, a robot called Fatty piercing a bystander's ankle after broke the glass of a booth. There hasn't been much information about Fatty since the incident.

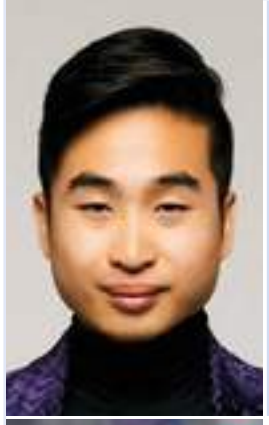

**2016:** Facial recognition rejected a legitimate citizens. An AI face recognition software used in the administration office of New Zealand, rejected an Asian citizen since it mistakenly registered his eyes as being closed. Currently, it is estimated that 20% of faces are still mistakenly classified.

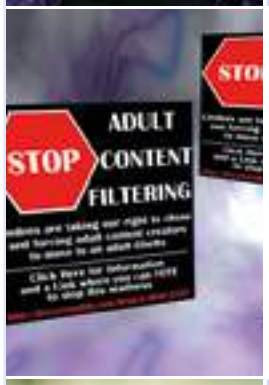

**2015:** Adult content filtering software failed to remove inappropriate content. Google's new YouTube Kids app failed to remove "inappropriate content," including explicit sexual language and jokes about paedophilia.

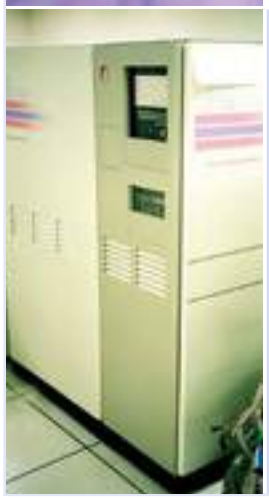

**1980:** The Fifth Generation Computer Systems. A government/industry research project in Japan aimed to create a computer using parallel computing. After 10 years of research and $400 million, the project was terminated without having met its goals to provide a platform for future developments in AI.

**Figure 13**
Example failures of AI

# 15. CONCLUSION AND RECOMMENDATIONS

There are many lessons that can be learnt from the past successes and failures of AI. To sustain the progress of AI, a rational and harmonic interaction is required between application specific projects and visionary research ideas. Along with the unprecedented enthusiasm of AI, there are also fears about the impact of the technology on our society. A clear strategy is required to consider the associated ethical and legal challenges to ensure that the society as a whole will benefit from the evolution of AI and its potential adverse effects are mitigated from early on. Such fears should not hinder the progress of AI but motivate the development of a systematic framework on which future AI will flourish. Most critical of all, it is important to understand science fiction from practical reality. With sustained funding and responsible investment, AI is set to transform the future of our society - our life, our living environment and our economy.

The following recommendations are relevant to the UK research community, industry, government agencies and policy makers:

- Robotics and AI are playing an increasingly important role in the UK's economy and its future growth. We need to be open and fully prepared for the changes that they bring to our society and their impact on the workforce structure and a shift in the skills base. Stronger national level engagement is essential to ensure the general public has a clear and factual view of the current and future development of robotics and AI.
- A strong research and development base for robotics and AI is fundamental to the UK, particularly in areas in which we already have a critical mass and international lead. Sustained investment in robotics and AI would ensure the future growth of the UK research base and funding needs to support key Clusters/Centres of Excellence that are internationally leading and weighted towards projects with greater social-economic benefit.
- It is important to address legal, regulatory and ethical issues for practical deployment and responsible innovation of robotics and AI; greater effort needs to be invested on assessing the economic impact and understanding how to maximise the benefits of these technologies while mitigating adverse effects.
- The government needs to tangibly support the workforce by adjusting their skills and business in creating opportunities based on new technologies. Training in digital skills and re-educating the existing workforce is essential to maintain the competitiveness of the UK.
- The UK has a strong track record in many areas of RAS and AI. Sustained investment in robotics and AI is critical to ensure the future growth of the UK research base and its international lead. It is also critical to invest in and develop the younger generation to be robotics and AI savvy with a strong STEM foundation by making effective use of new technical skills.

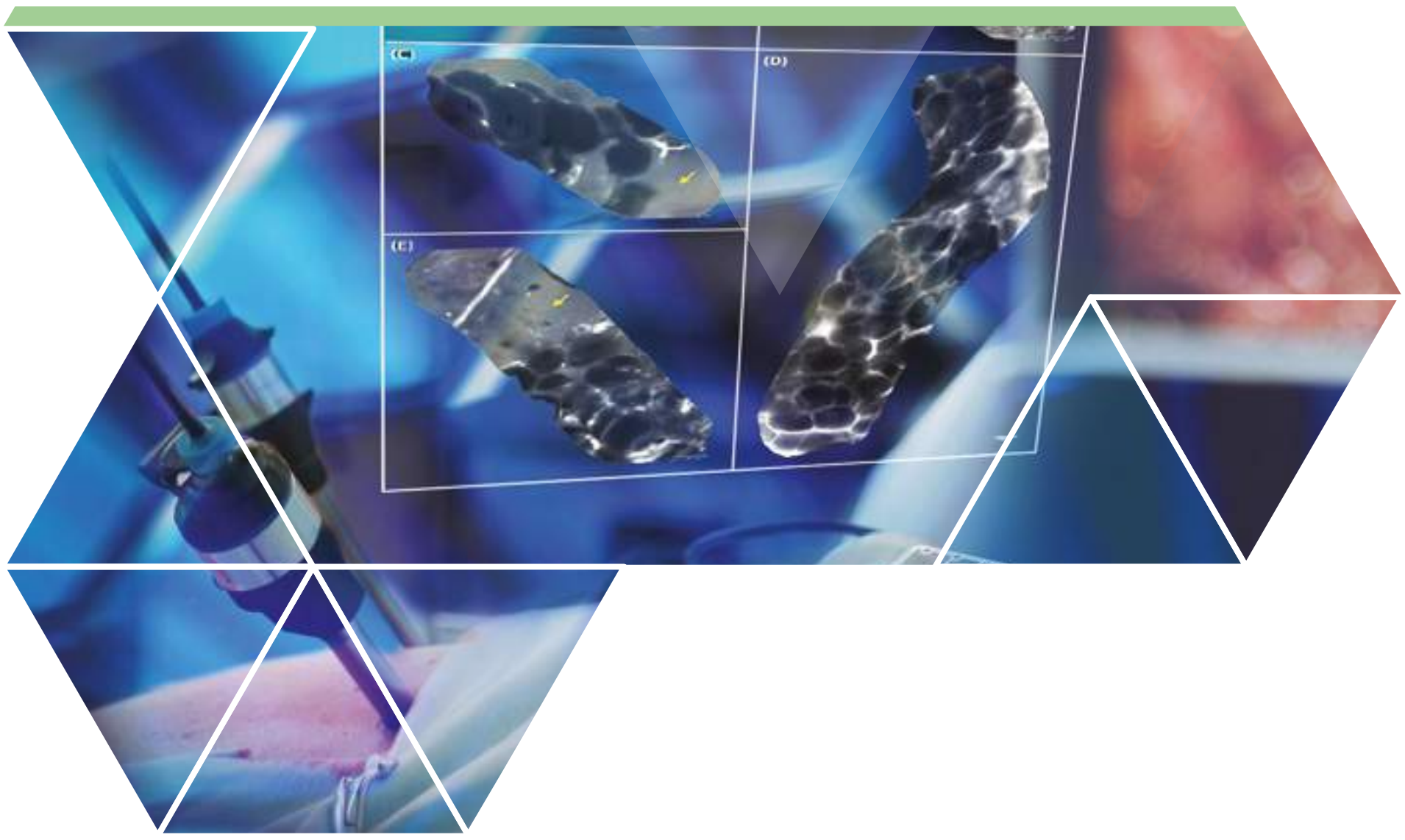

# REFERENCES

[1] S. J. Russell and P. Norvig, Artificial intelligence: a modern approach (3rd edition): Prentice Hall, 2009.

[2] I. Lighthill, "Artificial Intelligence: A General Survey," in Artificial Intelligence: A Paper Symposium. London: Science Research Council, 1973.

[3] M. Minsky and S. Papert, "Perceptrons," 1969.

[4] J. J. Hopfield, "Neural networks and physical systems with emergent collective computational abilities," Proceedings of the national academy of sciences, vol. 79, pp. 2554-2558, 1982.

[5] D. E. Rumelhart, G. E. Hinton, and R. J. Williams, "Learning internal representations by error propagation," DTIC Document1985.

[6] M. Wooldridge and N. R. Jennings, "Intelligent agents: Theory and practice," The knowledge engineering review, vol. 10, pp. 115-152, 1995.

[7] L. A. Zadeh, "Fuzzy logic—a personal perspective," Fuzzy Sets and Systems, vol. 281, pp. 4-20, 2015.

[8] N. Spinrad, "Mr Singularity," Nature, vol. 543, pp. 582-582, 2017.

[9] E. Horvitz, "One Hundred Year Study on Artificial Intelligence: Reflections and Framing," ed: Stanford University, 2014.

[10] S. Lohr, "The age of big data," New York Times, vol. 11, 2012.

[11] P. Alston, "Lethal robotic technologies: the implications for human rights and international humanitarian law," JL Inf. & Sci., vol. 21, p. 35, 2011.

[12] A. Young and M. Yung, "Deniable password snatching: On the possibility of evasive electronic espionage," in Security and Privacy, 1997. Proceedings., 1997 IEEE Symposium on, 1997, pp. 224-235.

[13] D. Kirat, G. Vigna, and C. Kruegel, "Barecloud: bare-metal analysis-based evasive malware detection," in 23rd USENIX Security Symposium (USENIX Security 14), 2014, pp. 287-301.

[14] C. Bryant and R. Waters, "Worker at Volkswagen plant killed in robot accident," in Finantial Times, ed, 2015.

[15] J. Andreu-Perez, D. R. Leff, K. Shetty, A. Darzi, and G.-Z. Yang, "Disparity in Frontal Lobe Connectivity on a Complex Bimanual Motor Task Aids in Classification of Operator Skill Level," Brain connectivity, vol. 6, pp. 375-388, 2016.

[16] J. Harrison, K. İzzetoğlu, H. Ayaz, B. Willems, S. Hah, U. Ahlstrom, et al., "Cognitive workload and learning assessment during the implementation of a next-generation air traffic control technology using functional near-infrared spectroscopy," IEEE Transactions on Human-Machine Systems, vol. 44, pp. 429-440, 2014.

[17] A. J. Gonzalez and V. Barr, "Validation and verification of intelligent systems-what are they and how are they different?," Journal of Experimental & Theoretical Artificial Intelligence, vol. 12, pp. 407-420, 2000.

[18] S. Ratschan and Z. She, "Safety verification of hybrid systems by constraint propagation based abstraction refinement," in International Workshop on Hybrid Systems: Computation and Control, 2005, pp. 573-589.

[19] E. Broadbent, R. Stafford, and B. MacDonald, "Acceptance of healthcare robots for the older population: review and future directions," International Journal of Social Robotics, vol. 1, pp. 319-330, 2009.

[20] C.-A. Smarr, A. Prakash, J. M. Beer, T. L. Mitzner, C. C. Kemp, and W. A. Rogers, "Older adults' preferences for and acceptance of robot assistance for everyday living tasks," in Proceedings of the Human Factors and Ergonomics Society Annual Meeting, 2012, pp. 153-157.

[21] R. C. O'Reilly and Y. Munakata, Computational explorations in cognitive neuroscience: Understanding the mind by simulating the brain: MIT press, 2000.

[22] G.-Z. Yang. (2017) Robotics and AI Driving the UK's Industrial Strategy. Ingenia.

[23] S. Inc, "Artificial Intelligence (AI)," 2016.

[24] S. Farquhar. (2017). Changes in funding in the AI safety field. Available: https://www.centreforeffectivealtruism.org/blog/changes-in-funding-in-the-ai-safety-field

[25] T. Reuters, "Web of Science," 2012.

[26] G. Scimago, "SJR—SCImago Journal & Country Rank," ed, 2007.

[27] B. Diallo and M. Lupu, "Future Patent Search," in Current Challenges in Patent Information Retrieval, ed: Springer Berlin Heidelberg, 2017, pp. 433-455.

[28] N. Chen, L. Christensen, K. Gallagher, R. Mate, and G. Rafert, "Global Economic Impacts Associated with Artificial Intelligence," Study, Analysis Group, Boston, MA, February, vol. 25, 2016.

[29] C. M. R. Christine Zhen-Wei Qiang, Kaoru Kimura, "Economic Impacts of Broadband," in The World Bank, ed, 2009.

[30] N. Czernich, O. Falck, T. Kretschmer, and L. Woessmann, "Broadband Infrastructure and Economic Growth," Economic Journal, vol. 121, pp. 505-532, May 2011.

[31] J. Andreu-Perez, C. C. Poon, R. D. Merrifield, S. T. Wong, and G.-Z. Yang, "Big data for health," IEEE journal of biomedical and health informatics, vol. 19, pp. 1193-1208, 2015.

# REFERENCES

[32] D. Ravi, C. Wong, F. Deligianni, M. Berthelot, J. Andreu-Perez, B. Lo, et al., "Deep Learning for Health Informatics," IEEE Journal of Biomedical and Health Informatics, vol. 21, pp. 4-21, Jan 2017.

[33] Y. LeCun, Y. Bengio, and G. Hinton, "Deep learning," Nature, vol. 521, pp. 436-444, 2015.

[34] L. A. Zadeh, "Fuzzy logic= computing with words," Fuzzy Systems, IEEE Transactions on, vol. 4, pp. 103-111, 1996.

[35] D. B. Fogel, Evolutionary computation: toward a new philosophy of machine intelligence vol. 1: John Wiley & Sons, 2006.

[36] L. Breiman, "Statistical modeling: The two cultures (with comments and a rejoinder by the author)," Statistical science, vol. 16, pp. 199-231, 2001.

[37] S. Senn, "Trying to be precise about vagueness," Statistics in medicine, vol. 26, p. 1417, 2007.

[38] W. S. McCulloch and W. Pitts, "A logical calculus of the ideas immanent in nervous activity," The bulletin of mathematical biophysics, vol. 5, pp. 115-133, 1943.

[39] F. Rosenblatt, "The perceptron: A probabilistic model for information storage and organization in the brain," Psychological review, vol. 65, p. 386, 1958.

[40] P. Werbos, "Beyond regression: new tools for prediction and analysis in the behavioral sciences [Ph. D. thesis] Cambridge," Mass, USA: Hardward University, 1974.

[41] K. Hornik, M. Stinchcombe, and H. White, "Multilayer feedforward networks are universal approximators," Neural networks, vol. 2, pp. 359-366, 1989.

[42] D. E. Rumelhart, G. E. Hinton, and R. J. Williams, "Learning representations by back-propagating errors," Cognitive modeling, vol. 5, p. 1, 1988.

[43] R. J. Williams and D. Zipser, "A learning algorithm for continually running fully recurrent neural networks," Neural computation, pp. 270-280 1989.

[44] S. Hochreiter and J. Schmidhuber, "Long short-term memory," Neural computation, vol. 9, pp. 1735-1780, 1997.

[45] Y. LeCun, L. Bottou, Y. Bengio, and P. Haffner, "Gradient-based learning applied to document recognition," Proceedings of the IEEE, vol. 86, pp. 2278-2324, 1998.

[46] G. E. Hinton, S. Osindero, and Y.-W. Teh, "A fast learning algorithm for deep belief nets," Neural computation, vol. 18, pp. 1527-1554, 2006.

[47] G. E. Hinton and T. J. Sejnowski, "Learning and releaming in Boltzmann machines," Parallel Distrilmted Processing, vol. 1, 1986.

[48] A. Krizhevsky, I. Sutskever, and G. E. Hinton, "Imagenet classification with deep convolutional neural networks," in Advances in neural information processing systems, 2012, pp. 1097-1105.

[49] S. Ioffe and C. Szegedy, "Batch normalization: Accelerating deep network training by reducing internal covariate shift," arXiv preprint arXiv:1502.03167, 2015.

[50] D. Silver, A. Huang, C. J. Maddison, A. Guez, L. Sifre, G. Van Den Driessche, et al., "Mastering the game of Go with deep neural networks and tree search," Nature, vol. 529, p. 484, 2016.

[51] M. Moravčík, M. Schmid, N. Burch, V. Lisý, D. Morrill, N. Bard, et al., "DeepStack: Expert-Level Artificial Intelligence in No-Limit Poker," arXiv preprint arXiv:1701.01724, 2017.

[52] C. Rogers, "Google Sees Self-Driving Cars on Road within Five Years," Wall Street Journal, 2015.

[53] D. Floreano and R. J. Wood, "Science, technology and the future of small autonomous drones," Nature, vol. 521, pp. 460-466, 2015.

[54] Z. Chen, X. Jia, A. Riedel, and M. Zhang, "A bio-inspired swimming robot," in Robotics and Automation (ICRA), 2014 IEEE International Conference on, 2014, pp. 2564-2564.

[55] Y. Ohmura and Y. Kuniyoshi, "Humanoid robot which can lift a 30kg box by whole body contact and tactile feedback," in Intelligent Robots and Systems, 2007. IROS 2007. IEEE/RSJ International Conference on, 2007, pp. 1136-1141.

[56] Z. Kappassov, J.-A. Corrales, and V. Perdereau, "Tactile sensing in dexterous robot hands—Review," Robotics and Autonomous Systems, vol. 74, pp. 195-220, 2015.

[57] H. Arisumi, S. Miossec, J.-R. Chardonnet, and K. Yokoi, "Dynamic lifting by whole body motion of humanoid robots," in Intelligent Robots and Systems, 2008. IROS 2008. IEEE/RSJ International Conference on, 2008, pp. 668-675.

[58] M. Asada, "Towards artificial empathy," International Journal of Social Robotics, vol. 7, pp. 19-33, 2015.

[59] L. Zhang, M. Jiang, D. Farid, and M. A. Hossain, "Intelligent facial emotion recognition and semantic-based topic detection for a humanoid robot," Expert Systems with Applications, vol. 40, pp. 5160-5168, 2013.

[60] N. Mavridis, "A review of verbal and non-verbal human–robot interactive communication," Robotics and Autonomous Systems, vol. 63, pp. 22-35, 2015.

[61] T. Kruse, A. K. Pandey, R. Alami, and A. Kirsch, "Human-aware robot navigation: A survey," Robotics and Autonomous Systems, vol. 61, pp. 1726-1743, 2013.

[62] K. Mochizuki, S. Nishide, H. G. Okuno, and T. Ogata, "Developmental human-robot imitation learning of drawing with a neuro dynamical system," in Systems, Man, and Cybernetics (SMC), 2013 IEEE International Conference on, 2013, pp. 2336-2341.

[63] P.-Y. Oudeyer, "Socially guided intrinsic motivation for robot learning of motor skills," Autonomous Robots, vol. 36, pp. 273-294, 2014.

[64] M. T. Chan, R. Gorbet, P. Beesley, and D. Kulič, "Curiosity-Based Learning Algorithm for Distributed Interactive Sculptural Systems," in Intelligent Robots and Systems (IROS), 2015 IEEE/RSJ International Conference on, 2015, pp. 3435-3441.

[65] J. McCarthy, Programs with common sense: RLE and MIT Computation Center, 1960.

[66] H. Murase and S. K. Nayar, "Visual Learning and Recognition of 3-D Objects from Appearance," International Journal of Computer Vision, vol. 14, pp. 5-24, Jan 1995.

[67] P. Viola and M. Jones, "Robust real-time face detection," Eighth IEEE International Conference on Computer Vision, pp. 747-747, 2001.

[68] A. Esteva, B. Kuprel, R. A. Novoa, J. Ko, S. M. Swetter, H. M. Blau, et al., "Dermatologist-level classification of skin cancer with deep neural networks," Nature, vol. 542, pp. 115-+, Feb 2 2017.

[69] R. Fergus, P. Perona, and A. Zisserman, "Object class recognition by unsupervised scale-invariant learning," 2003 IEEE Computer Society Conference on Computer Vision and Pattern Recognition, pp. 264-271, 2003.

[70] T. Leung and J. Malik, "Representing and recognizing the visual appearance of materials using three-dimensional textons," International Journal of Computer Vision, vol. 43, pp. 29-44, 2001.

[71] T. R. Society, "Machine learning:  the power and promise of computers that learn by example," ed. The Royal Society, 2017.

[72] N. B. Anders Sandberg, "Whole Brain Emulation: A Roadmap," Future of Humanity Institute, Oxford University2008.

[73] M. Colombo, "Why build a virtual brain? Large-scale neural simulations as jump start for cognitive computing," Journal of Experimental and Theoretical Artificial Intelligence, vol. 29, pp. 361-370, 2017.

[74] P. Hankins, "Trying to simulate the human brain is a waste of energy," E. Lake, Ed., ed, 2017.

[75] A. Prieto, B. Prieto, E. M. Ortigosa, E. Ros, F. Pelayo, J. Ortega, et al., "Neural networks: An overview of early research, current frameworks and new challenges," Neurocomputing, vol. 214, pp. 242-268, Nov 19 2016.

[76] D. Hémous and M. Olsen, "The Rise of the Machines: Automation, Horizontal Innovation and Income Inequality," 2016.

[77] M. E. Virgillito, "Rise of the robots: technology and the threat of a jobless future," Labor History, vol. 58, pp. 240-242, 2017.

[78] T. E. P. s. L. A. Committee, "European Civil Law Rules in Robotics," 2016.

"

The impact of robotics and AI will affect not only manufacturing, transport and healthcare, but also jobs in agrifood, logistics, security, retail, and construction. It is important to assess the economic impact and understand the social, legal and ethical issues of robotics and AI in order to maximise the benefits of these technologies while mitigating adverse effects. Establishing our lead in robotics and AI is an opportunity that the UK cannot afford to miss. The future lies in our coordinated effort to establish our niche and leverage the significant strengths we already have, and expand upon those areas that are strategic to the UK.

"

UK-RAS NETWORK
ROBOTICS & AUTONOMOUS SYSTEMS

www.ukras.org

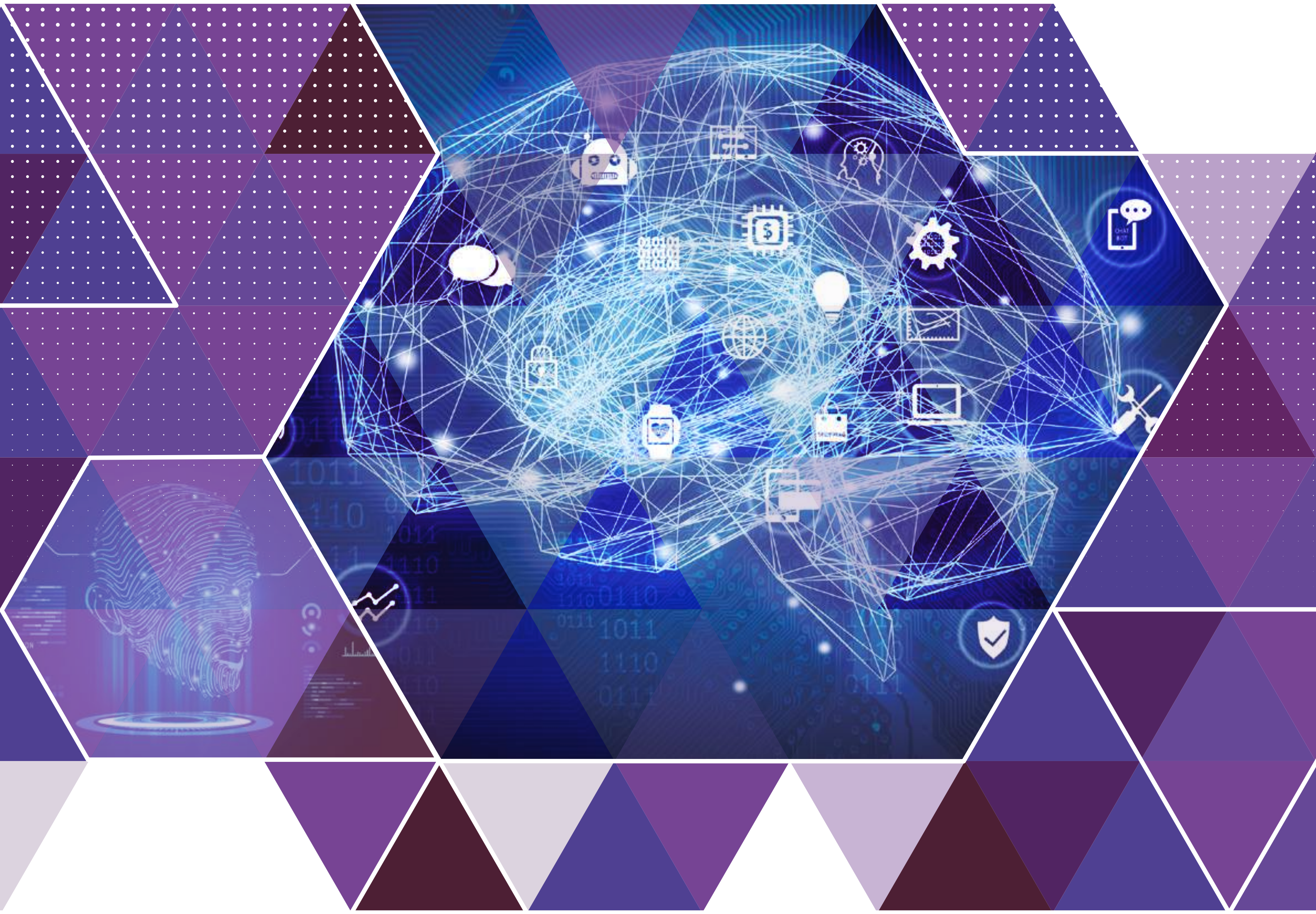